%% file: main.tex
\documentclass[11pt]{article}
\usepackage[preprint]{acl}
\usepackage{times}
\usepackage{latexsym}
\usepackage[T1]{fontenc}
\usepackage[utf8]{inputenc}
\usepackage{microtype}
\usepackage{inconsolata}
\usepackage{graphicx}
\definecolor{spaceblue}{HTML}{087CB3}
\definecolor{spaceorange}{HTML}{E56932}
\definecolor{spacegreen}{HTML}{079B80}
\definecolor{spacepurple}{HTML}{8578BC}
\definecolor{spacegray}{HTML}{A8B1BB}
\usepackage{amsmath,amssymb}
\usepackage{booktabs,multirow,tabularx}
\usepackage{placeins}
\usepackage{float}
\usepackage{stfloats}
\title{More Than Just Style: Benchmarking Personality Influence in Large Language Models}
\author{
  Yuhao Liu\textsuperscript{1}\thanks{Equal contribution.}\quad
  Mengyu Wang\textsuperscript{2}\footnotemark[1]\quad
  Xiao Chen\textsuperscript{3}\thanks{Corresponding authors.}\quad
  Lu Liu\textsuperscript{1}\footnotemark[2]\\[0.5em]
  {\normalfont\normalsize\textsuperscript{1}University of Exeter\quad
  \textsuperscript{2}University of Edinburgh}\\
  {\normalfont\normalsize\textsuperscript{3}Nanchang University}\\[0.3em]
  {\normalfont\small\texttt{yl1289@exeter.ac.uk}\quad\texttt{mengyu.wang@ed.ac.uk}}\\
  {\normalfont\small\texttt{xiao.chen@ncu.edu.cn}\quad\texttt{l.liu3@exeter.ac.uk}}
}
\hypersetup{
  pdftitle={More Than Just Style: Benchmarking Personality Influence in Large Language Models},
  pdfauthor={Yuhao Liu, Mengyu Wang, Xiao Chen, Lu Liu}
}
\begin{document}
\raggedbottom

\maketitle
\input{sections/abstract.tex}

\input{sections/introduction.tex}
\input{sections/related_work.tex}
\input{sections/preliminary.tex}
\input{sections/study_design.tex}
\input{sections/experimental_settings.tex}
\input{sections/results_and_discussion.tex}
\input{sections/conclusion.tex}
\input{sections/limitations.tex}

\bibliography{space_references}

\appendix
\raggedbottom
\makeatletter
\setlength{\@fptop}{0pt plus .001fil}
\setlength{\@dblfptop}{0pt plus .001fil}
\setlength{\@fpsep}{10pt plus .001fil}
\setlength{\@dblfpsep}{10pt plus .001fil}
\setlength{\@fpbot}{0pt plus 1fil}
\setlength{\@dblfpbot}{0pt plus 1fil}
\makeatother
\input{sections/appendix_details.tex}
\input{sections/appendix_capabilities.tex}
\input{sections/appendix_interpretation.tex}
\input{sections/appendix_complete.tex}
\input{sections/appendix_statistics.tex}
\input{sections/appendix_other_dimensions.tex}

\end{document}

%% file: sections/abstract.tex
\begin{abstract}
In addition to the effects of explicit input prompts and instructions, large language model outputs are influenced by underlying behavioral preferences similar to human personality tendencies. These tendencies can affect how models interpret situations and make decisions. However, their influence is difficult to disentangle from the effects of model capabilities and instructions. Therefore, we introduce SPACE, a benchmark framework combining a controlled evaluation protocol with complementary datasets for personality verification, capability assessment, and hierarchical situational tasks. Controlled comparisons across guidance conditions help distinguish personality influence from measured capability changes and immediate instruction effects. Experimental results demonstrate that although prompting can shift generation style, models' internal personality tendencies are persistent and not readily overridden by opposing instructions. Furthermore, these underlying tendencies have an observable influence on LLM task performance, highlighting the need for greater attention to these characteristics and potential applications. Training code and SPACE benchmark specifications are available at \url{https://github.com/Niklauseik/SPACE-training}.

\end{abstract}

%% file: sections/introduction.tex
\section{Introduction}

As large language models (LLMs) are increasingly deployed in complex settings, their judgments, decision preferences, and behavioral patterns are attracting greater attention. In humans, judgments and decisions reflect not only knowledge and reasoning abilities but also personality preferences and tendencies. This connection has motivated research into measuring and shaping personality in LLMs \citep{jiang2023personality,pan2023mbti}.

\begin{figure}[!t]
    \centering
    \includegraphics[width=\columnwidth]{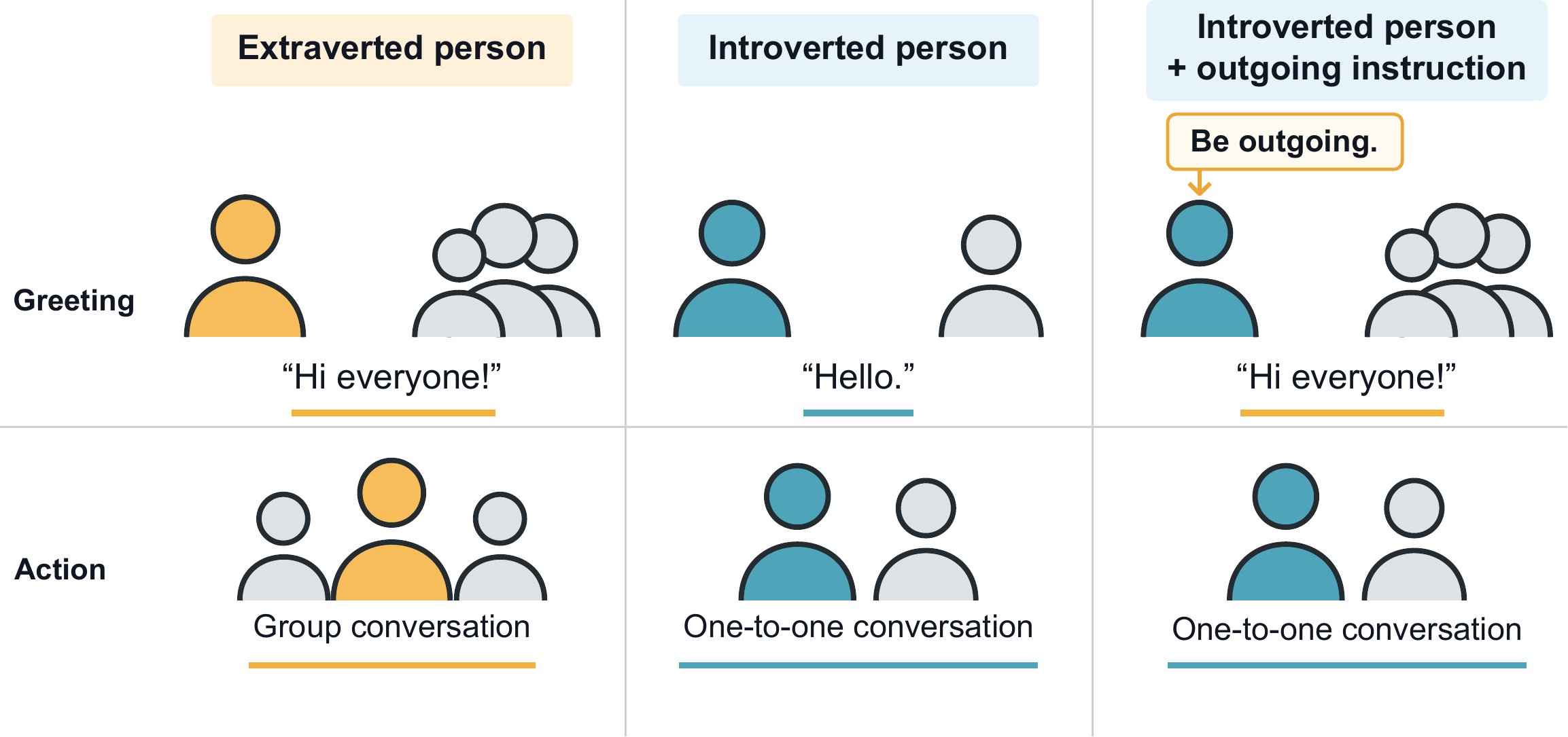}
    \caption{A human analogy for personality influence in LLMs. Instructions can change outward expression while underlying preferences continue to shape actions.}
    \label{fig:personality_motivation}
\end{figure}

Personality and external instructions can influence different aspects of behavior. As illustrated in Figure~\ref{fig:personality_motivation}, an introverted person adopts an outgoing greeting style when instructed to do so, but still prefers a conversation with only one other person. This analogy raises a central question for LLMs: do models exhibit underlying personality tendencies that continue to shape judgments and decisions under competing instructions? We use personality tendencies to refer to such behavioral preferences.

Existing studies provide evidence that personality characteristics can be measured and guided in LLMs. Researchers have examined personality questionnaires, generated responses, role consistency, and dialogue guided by personas, assessing reported preferences and patterns of expression. Other studies investigate their relationship with social decisions, safety, and task performance \citep{jiang-etal-2024-personallm,wang-etal-2024-rolellm,li-etal-2025-big5,zakazov2024social}.

However, a model's outward expressions or claimed personality traits do not necessarily correspond to its decisions and behavior in specific situations. We define situational tasks as tasks that place a model in a specific context requiring interpretation, choice, or interaction. Identifying personality influence in these tasks is difficult because each model output jointly reflects its basic capabilities, the prompt instructions, and its underlying behavioral preferences, making their individual contributions difficult to disentangle.


To address this challenge, we introduce SPACE, a benchmark framework for Systematic Personality Analysis and Comprehensive Evaluation. SPACE combines a controlled evaluation protocol with a suite of complementary datasets to support rigorous assessment. The protocol has two stages. First, personality verification and basic capability assessments establish whether personality guidance changes models' preferences and identify potential changes in the models’ underlying task capabilities. These measurements help distinguish personality-related behavioral differences from differences in basic capability. Second, we compare base models, personality variants trained using direct preference optimization (DPO), prompted models, and their combinations on matched situational tasks. These comparisons isolate the effects of models' preferences and immediate instructions, and test whether personality-related behavioral differences persist under opposing prompts. 

To implement this protocol, we collect a broad suite of datasets covering personality verification, basic capability assessment, and three levels of situational tasks: situational interpretation, individual decision making, and interactive decision making. Together, SPACE disentangles the effects of personality from those of model capabilities and instructions, enabling a systematic study of personality influence in LLMs.

Our experiments show that personality tendencies influence situational judgments and decisions while measured basic capabilities remain broadly stable. Models with contrasting preferences differ in how they interpret ambiguous information, choose among risky alternatives, and respond during interactions. These differences affect both interpretation accuracy and downstream decision outcomes. Although prompting can alter outward expression choices, the influence of underlying preferences persists under opposing instructions, extending beyond changes in generation style.

In summary, our contributions are threefold:

\begin{itemize}
\item \textbf{A controlled evaluation protocol.} We design systematic comparisons to distinguish personality influence from capability changes and immediate instruction effects, enabling rigorous benchmarking of personality influence across models and task settings.

\item \textbf{A hierarchical evaluation suite.} We assemble existing datasets and task adaptations for personality verification, capability assessment, and three levels of situational tasks. These complementary measurements implement the protocol and reveal personality influence on LLMs' outputs.

\item \textbf{Evidence of persistent personality influence.} Experimental results demonstrate the personality effects on judgments, choices, and downstream task outcomes. They show that LLMs' behavioral influence often persists under opposing instructions, extending beyond changes in outward expression.
\end{itemize}

%% file: sections/related_work.tex
\section{Related Work}

Research on LLM personality uses psychological inventories and model responses to characterize personality tendencies \citep{jiang2023personality,pan2023mbti}. Big Five and MBTI assessments describe expressed traits, while TRAIT \citep{lee-etal-2025-llms} examines the reliability and consistency of personality measurement through questions about specific scenarios. These approaches provide tools for measuring personality tendencies, but expressed traits alone do not establish how models interpret situations or make decisions.

Prior work uses prompting and training to guide personality expression and behavior in LLMs. PersonaLLM \citep{jiang-etal-2024-personallm} investigates personality expression under prompting, while Machine Mindset \citep{cui2023mindset} uses fine tuning and preference optimization to shape MBTI preferences. BIG5CHAT \citep{li-etal-2025-big5} compares training with prompting and evaluates personality alignment and reasoning performance. Together, these studies examine how different forms of guidance affect expressed personality and task performance.

Personality and persona guidance also influence model behavior in specific situations. \citet{deshpande-etal-2023-toxicity} show that persona assignment affects toxicity and discriminatory outputs. \citet{zakazov2024social} examine whether personality prompts produce corresponding decisions in social situations and identify limits to this form of behavioral control. This analysis is especially vital given the growing reliance on LLMs for subjective and decision-heavy tasks, including financial sentiment analysis~\cite{wang2024mana,wang2024modeling} and domain-specific applications~\cite{wang2025one,wang2026self}. Such findings highlight the need to distinguish expressed personality from behavioral preferences and to examine how those preferences interact with instructions.

Existing studies rarely examine personality influence across situational tasks under controlled conditions. SPACE addresses this gap through systematic behavioral evaluation.

%% file: sections/preliminary.tex
\section{Preliminaries}

\subsection{Thinking and Feeling Preferences}

We study personality through the Thinking (T) and Feeling (F) dimensions of MBTI (The Myers and Briggs Foundation, n.d.). T emphasizes logic and consistency, while F emphasizes personal values and interpersonal considerations. We examine how these preferences influence model judgments and decisions. Results for other dimensions are in Appendix F.

\subsection{Training and Prompting}

We use DPO \citep{rafailov2023dpo} to train T and F variants using response preferences that reflect the target tendency. DPO updates model parameters, whereas prompting supplies instructions at inference time.

%% file: sections/study_design.tex
\section{Study Design}

\begin{figure*}[!tb]
\centering
\includegraphics[width=\textwidth]{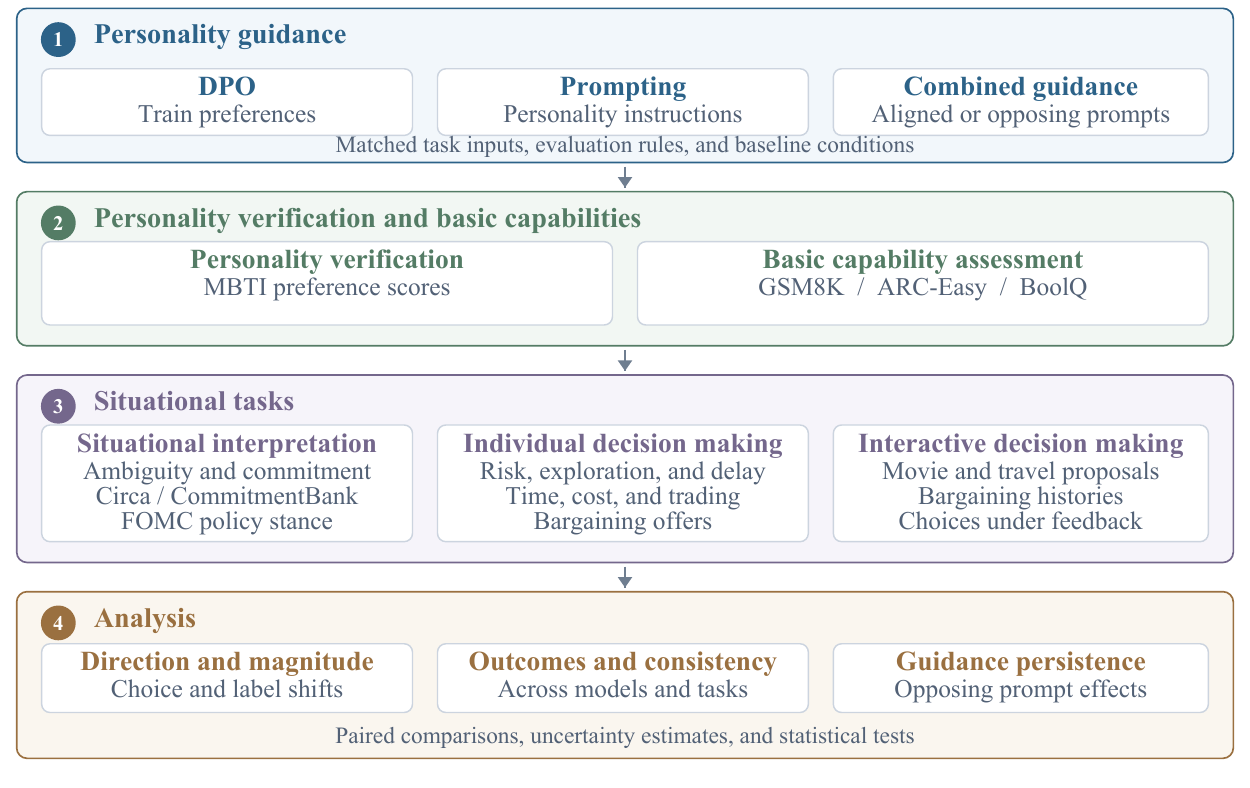}
\caption{The SPACE framework for examining personality influence through controlled comparisons.}
\label{fig:spaceframework}
\end{figure*}

SPACE examines how personality preferences influence model behavior and how these influences interact with subsequent instructions. The framework follows four stages: personality guidance; personality verification and basic capability assessment; hierarchical situational task evaluation; and controlled behavioral analysis. Figure~\ref{fig:spaceframework} summarizes the workflow.

\subsection{Personality Guidance}

We construct contrasting T and F conditions using preference training and personality prompting. Each unmodified model serves as Base. We also apply aligned and opposing prompts to the trained variants to examine how their preferences interact with subsequent instructions.

An MBTI questionnaire checks whether the guidance produces responses consistent with the intended preference. Trained variants are assessed without additional personality prompts, while prompted conditions receive their corresponding instructions. The questionnaire verifies guidance effectiveness; situational tasks examine how the preferences influence judgments and decisions.

\subsection{Basic Capability Assessment}

We evaluate guided conditions on objective benchmarks and compare their performance with the corresponding Base model. These assessments check whether personality guidance affects basic capabilities.

This comparison helps distinguish changes in behavioral preferences from changes in the ability to understand information and solve problems. Broadly stable benchmark performance supports this distinction, without assuming that personality guidance leaves every aspect of model capability unchanged.

\subsection{Hierarchical Situational Tasks}

Human behavior research distinguishes the interpretation of situational cues, the evaluation and selection of responses, and their enactment \citep{dodge1994processing}. Actions can also generate feedback that influences subsequent behavior \citep{bandura1989agency}. Drawing on these distinctions, we organize situational tasks into three levels: situational interpretation, individual decision making, and interactive decision making.

Each level examines a different aspect of personality influence. Interpretation tasks examine judgments about the same information. Individual decision tasks examine preferences among the same alternatives and consequences. Interactive tasks examine responses to other participants, interaction histories, or feedback. We bring together datasets and task adaptations that support these complementary comparisons, as summarized in Table~\ref{tab:framework}.

\begin{table*}[!tb]
\centering
\small
\setlength{\tabcolsep}{4pt}
\begin{tabularx}{\textwidth}{
@{}
>{\raggedright\arraybackslash}p{.17\textwidth}
>{\raggedright\arraybackslash}p{.21\textwidth}
>{\raggedright\arraybackslash}p{.33\textwidth}
>{\raggedright\arraybackslash}X
@{}
}
\toprule
Level & Task focus & Dataset / task & Description \\
\midrule

\multirow[t]{3}{.17\textwidth}{Situational interpretation}
& \multirow[t]{3}{.21\textwidth}{Information interpretation; situational judgments}
& Circa \citep{louis-etal-2020-id}
& Indirect answers \\
& & CommitmentBank \citep{demarneffe2019cb}
& Speaker commitment \\
& & FOMC \citep{shah2023fomc}
& Monetary policy stance \\

\midrule

\multirow[t]{6}{.17\textwidth}{Individual decision making}
& \multirow[t]{6}{.21\textwidth}{Action selection; rewards, costs, time, and uncertainty}
& Choices13k \citep{peterson2021choices}
& Lottery choices \\
& & Transportation
& Time and cost \\
& & Machine selection
& Limited reward observations \\
& & Delayed rewards
& Earlier versus later payments \\
& & Stock scenarios
& Trading choices \\
& & Bargaining scenarios
& Buyer offers \\

\midrule

\multirow[t]{3}{.17\textwidth}{Interactive decision making}
& \multirow[t]{3}{.21\textwidth}{Responses to participants, history, and feedback}
& INSPIRED \citep{hayati-etal-2020-inspired}
& Movie recommendations \\
& & Frames \citep{el-asri-etal-2017-frames}
& Travel proposals \\
& & Tasks adapted from CraigslistBargain
\citep{he-etal-2018-decoupling}
& Bargaining histories; successive choices under program feedback \\

\bottomrule
\end{tabularx}
\caption{Situational task levels and datasets. Details are provided in Appendix~\ref{app:details}.}
\label{tab:framework}
\label{tab:datasets}
\end{table*}

We evaluate these levels separately to locate behavioral differences and examine their variation across tasks and models. Judgment and choice distributions characterize behavioral preferences, while accuracy provides an additional measure where reference labels are available.

\subsection{Controlled Comparisons and Behavioral Analysis}

Compared conditions receive the same task inputs and output requirements. Base and prompted Base conditions reveal the effects of immediate instructions. Base and unprompted trained variants reveal changes associated with training. Comparisons between T and F conditions examine whether contrasting preferences produce different judgments and choices in the same situations.

Combined conditions examine how trained preferences interact with subsequent instructions. Comparing a trained variant with and without a personality prompt measures how the instruction changes its behavior. Comparing the combined condition with Base under the same prompt examines whether differences associated with training remain when the instruction is held constant. Aligned prompts test whether instructions reinforce the behavioral influence of the trained preference, while opposing prompts test whether they weaken or reverse that influence.

For each behavioral measure, we examine both the direction and magnitude of the differences. A tendency can remain in the same direction while becoming substantially weaker, so the absence of reversal alone does not establish that its influence is unchanged. Together with personality verification and capability assessment, these comparisons allow us to examine how trained preferences contribute to model behavior and how much of their influence remains under competing instructions.

%% file: sections/experimental_settings.tex
\section{Experimental Settings}

\textbf{Data.}
We construct personality preference data from the English behavior split of Machine Mindset \citep{cui2023mindset} and use an MBTI questionnaire to verify the intended preference. Basic capability benchmarks include GSM8K \citep{cobbe2021gsm}, ARC-Easy \citep{clark2018arc}, and BoolQ \citep{clark-etal-2019-boolq}. Situational datasets and task adaptations are organized by behavioral level in Table~\ref{tab:framework}. Dataset statistics, selection criteria, and adaptations are detailed in Appendix~\ref{app:details}.

\textbf{Models.}
We evaluate Llama 3.2 3B Instruct, Llama 3.1 8B Instruct, Qwen2.5 3B Instruct, and Qwen2.5 7B Instruct \citep{meta2024llama,dubey2024llama,qwen2024report}. Each unmodified model serves as Base. We refer to the trained models as the Thinking variant (T) and Feeling variant (F). Conditions include these variants, Thinking or Feeling prompts applied to Base, and aligned or opposing prompts applied to the variants.

\textbf{Training and prompting.}
We train the Thinking and Feeling variants using DPO with LoRA \citep{hu2022lora}, using the corresponding Base model as the frozen reference, a learning rate of $10^{-5}$, and $\beta=1.0$. We apply early stopping based on training loss and merge the adapters for inference. Fixed personality prompts emphasize logic and consistency for T, and empathy and personal values for F. They precede the task input and are reapplied at each model turn in interactive conditions. Training recipes and prompt templates appear in Appendix~\ref{app:training} and Appendix~\ref{app:prompts}.

\textbf{Evaluation.}
Compared conditions use the same task templates and output requirements, with greedy decoding for generative evaluations. We extract labels from responses and report personality preference scores, capability accuracy, and situational judgment and choice distributions. Behavioral shifts relative to the corresponding Base are measured in percentage points. Situational accuracy is reported where objective reference labels are available; choices and interaction outcomes characterize behavior. Capability scoring is detailed in Appendix~\ref{app:capability_scope}, and interpretation labels and metrics in Appendix~\ref{app:interpretation}. Computational resources and budget are reported in Appendix~\ref{app:compute}.

%% file: sections/results_and_discussion.tex
\section{Results and Discussion}

We first verify the intended personality preferences and assess basic capabilities, then examine personality influence across three levels of situational tasks. We also examine whether this influence persists under opposing instructions and assess the statistical reliability of the results. Results for other dimensions are in Appendix~\ref{app:otherdimensions}.

\begin{figure*}[!t]
\centering
\includegraphics[width=\textwidth]{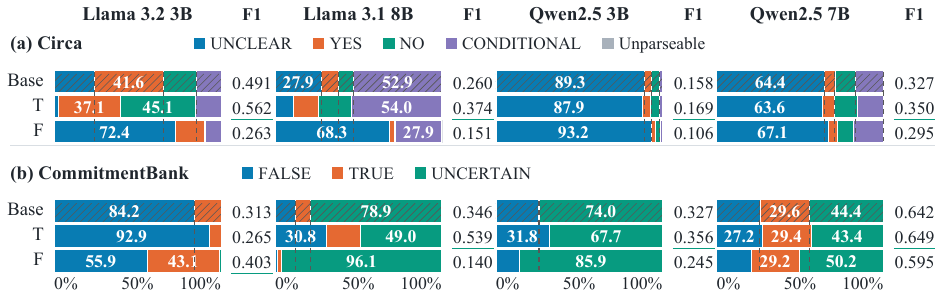}
\caption{Situational interpretation: prediction distributions (\%) and macro F1 on (a) Circa and (b) CommitmentBank.}
\label{fig:interpretation_grouped}
\end{figure*}

\subsection{Personality Verification and Basic Capabilities}

Both training and prompting shift questionnaire responses toward the intended preference in the evaluated conditions (Table~\ref{tab:personality}): F guidance increases Feeling choices, while T guidance produces the opposite pattern. Trained variants exhibit these shifts without additional personality prompts. These scores verify whether the guidance produces the intended questionnaire response patterns. Situational tasks then examine the corresponding behavioral preferences.

\begin{table}[!htbp]
\centering
\small
\setlength{\tabcolsep}{3pt}
\renewcommand{\arraystretch}{1.02}
\begin{tabular*}{\columnwidth}{@{\extracolsep{\fill}}lccc@{}}
\toprule
Model & Base & F guided & T guided \\
\midrule
Llama 3.2 3B & 63.0 & 82.6 / 93.5 & 39.1 / 8.7 \\
Qwen2.5 3B & 39.1 & 54.3 / 87.0 & 30.4 / 6.5 \\
Qwen2.5 7B & 47.8 & 56.5 / 95.7 & 41.3 / 2.2 \\
Llama 3.1 8B & 43.5 & 58.7 / NA & 39.1 / NA \\
\bottomrule
\end{tabular*}
\caption{Feeling choice proportions (\%), shown as trained variant / prompt.}
\label{tab:personality}
\end{table}

Trained variants retain broadly stable capability performance (Table~\ref{tab:capabilities}): 22 of 24 absolute accuracy changes are below 1.2 percentage points; the largest is a 2.25-point decline on Llama 3.2 3B T BoolQ. Prompting effects vary across models and benchmarks, with the largest decrease reaching 5.65 points on Llama 3.2 3B ARC-Easy. Evaluation protocols and additional results are provided in Appendix~\ref{app:capabilities}. Statistical tests are reported in Appendix~\ref{app:statistics}.

\begin{table}[!htbp]
\centering
\fontsize{9}{10}\selectfont
\setlength{\tabcolsep}{1.5pt}
\renewcommand{\arraystretch}{1.02}
\begin{tabular*}{\columnwidth}{@{\extracolsep{\fill}}lcccc@{}}
\toprule
Model & & ARC-Easy & BoolQ & GSM8K \\
\midrule
\multirow{2}{*}{Llama 3.2 3B}
 & T & \textminus{}0.55 / \textminus{}3.70 & \textminus{}2.25 / \textminus{}0.80 & \textminus{}0.69 / \textminus{}1.54 \\
 & F & \textminus{}0.15 / \textminus{}5.65 & +1.65 / +0.20 & \textminus{}0.85 / \textminus{}1.85 \\
\addlinespace[2pt]
\multirow{2}{*}{Qwen2.5 3B}
 & T & \textminus{}0.20 / \textminus{}1.95 & +0.30 / \textminus{}0.35 & \textminus{}0.54 / \textminus{}0.31 \\
 & F & \textminus{}0.10 / \textminus{}2.00 & \textminus{}0.85 / \textminus{}0.25 & +0.77 / \textminus{}0.23 \\
\addlinespace[2pt]
\multirow{2}{*}{Qwen2.5 7B}
 & T & \textminus{}0.05 / \textminus{}0.85 & +0.20 / \textminus{}4.05 & +0.46 / \textminus{}0.23 \\
 & F & \textminus{}0.15 / \textminus{}1.35 & \textminus{}0.20 / \textminus{}3.85 & \textminus{}0.62 / \textminus{}0.23 \\
\addlinespace[2pt]
\multirow{2}{*}{Llama 3.1 8B}
 & T & \textminus{}0.05 / NA & \textminus{}0.05 / NA & \textminus{}0.08 / NA \\
 & F & \textminus{}0.85 / NA & \textminus{}0.40 / NA & \textminus{}1.15 / NA \\
\bottomrule
\end{tabular*}
\caption{Capability accuracy changes (pp) from Base, shown as trained variant / prompt. NA indicates unavailable Prompt results for Llama 3.1 8B.}
\label{tab:capabilities}
\end{table}

We next examine whether these contrasting preferences also influence judgments and choices in situational tasks.

\begin{figure*}[!b]
\centering
\includegraphics[width=\textwidth]{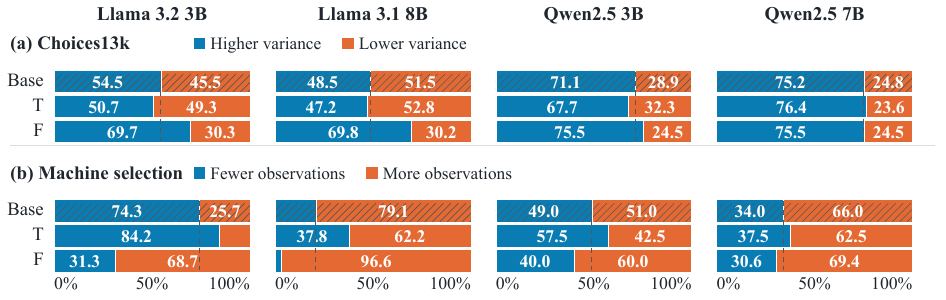}
\caption{Individual decision making: choice distributions (\%) on (a) Choices13k and (b) machine selection.}
\label{fig:decision_grouped}
\end{figure*}

\subsection{Personality Influence in Situational Tasks}

We examine how contrasting personality preferences influence situational judgments and choices. T and F below denote the trained variants with the corresponding preferences.

\subsubsection{Situational Interpretation}

On Circa, T reduces UNCLEAR predictions and F increases them in all four models (Figure~\ref{fig:interpretation_grouped}(a)). In Llama 3.2 3B, the proportion changes from 23.9\% at Base to 2.7\% for T and 72.4\% for F. These differences indicate contrasting tendencies to assign definite interpretations to indirect replies.

On CommitmentBank, T increases FALSE predictions and F decreases them across all four models, although the redistribution among other labels varies (Figure~\ref{fig:interpretation_grouped}(b)). These judgment shifts do not yield uniformly better task performance: T improves macro F1 on Circa in all four models, while its effects on CommitmentBank differ across models. Personality preferences therefore affect interpretation patterns, with consequences for accuracy depending on the task and model. FOMC results, scoring conventions, and output processing are detailed in Appendix~\ref{app:interpretation}.

\subsubsection{Individual Decision Making}

On Choices13k and the machine selection task, contrasting preferences produce different choices under the same decision conditions. T chooses alternatives with lower payoff variance more often than F in three of four models (Figure~\ref{fig:decision_grouped}(a)). T also selects machines with fewer previous observations more often than F in all four models; the proportions reach 84.2\% and 31.3\%, respectively, in Llama 3.2 3B (Figure~\ref{fig:decision_grouped}(b)). These results reveal differences in choices involving uncertainty and limited evidence.

\begin{figure}[!t]
\centering
\includegraphics[width=.95\columnwidth]{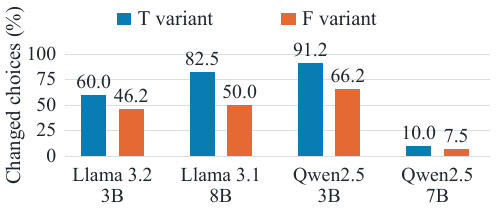}
\caption{Individual decision making: choice changes (\%) after expected payoffs are explicitly stated for otherwise identical risk scenarios.}
\label{fig:payoff}
\end{figure}

The expected payoff probe tests responses to changes in how a situation is presented. Adding calculated expected payoffs, while keeping probabilities and outcomes unchanged, changes T choices more often than F choices in all four models (Figure~\ref{fig:payoff}). In Llama 3.1 8B, 82.5\% of T choices change, compared with 50.0\% of F choices.

\begin{figure}[!t]
\centering
\includegraphics[width=\columnwidth]{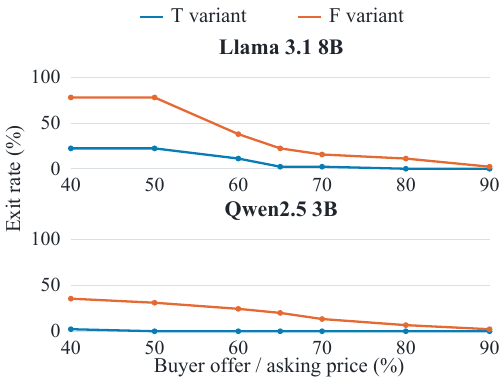}
\caption{Individual decision making: exit rates (\%) across buyer offers in the controlled bargaining probe.}
\label{fig:bargain}
\end{figure}

In the controlled bargaining probe, F has higher exit rates at low buyer offers, while the difference between T and F narrows as offers improve in Llama 3.1 8B and Qwen2.5 3B (Figure~\ref{fig:bargain}). Thus, personality preferences affect both action selection and responsiveness to decision conditions.

\FloatBarrier

\begin{figure*}[!t]
\centering
\includegraphics[width=\textwidth]{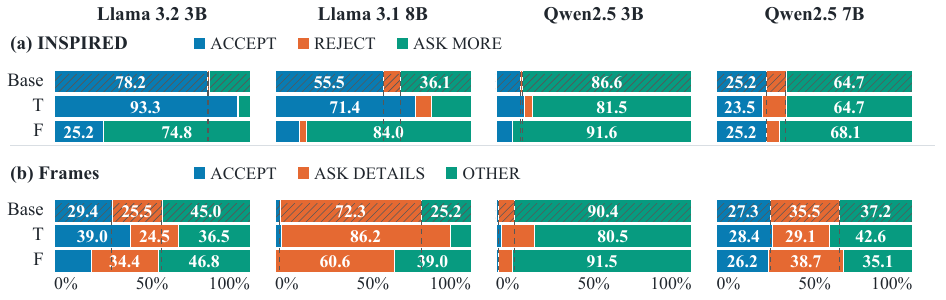}
\caption{Interactive decision making: choice distributions (\%) on (a) INSPIRED and (b) Frames.}
\label{fig:interaction_grouped}
\end{figure*}

\subsubsection{Interactive Decision Making}

Personality preferences also affect how models respond during recommendation interactions. On INSPIRED, F requests more information than T in all four models (Figure~\ref{fig:interaction_grouped}(a)). In Llama 3.2 3B, further inquiries account for 74.8\% of F responses and 5.9\% of T responses.

On Frames, T accepts travel proposals more often than F in all four models, reaching 39.0\% versus 18.8\% in Llama 3.2 3B (Figure~\ref{fig:interaction_grouped}(b)). These tasks show distinct effects on whether models accept a proposal or continue seeking information. 

Together, these results show personality influence on situational interpretation, individual choices, and interaction outcomes. The observed differences have consequences for task accuracy and interaction choices. Additional decision and interaction results are provided in Appendix~\ref{app:situational_full}.

\subsection{Persistence of Personality Influence under Opposing Instructions}

\begin{figure}[!t]
\centering
\includegraphics[width=\columnwidth]{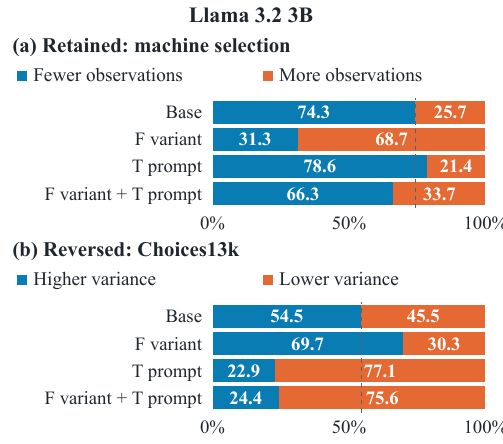}
\caption{Individual decision making under opposing instructions: (a) a retained shift on machine selection and (b) a reversed shift on Choices13k.}
\label{fig:opposing_examples}
\end{figure}

We examine whether these trained preferences continue to influence decisions when subsequent instructions explicitly request the opposing personality. Figure~\ref{fig:opposing_examples} presents a retained example and a reversal, both using the Feeling variant with a Thinking prompt.

On the machine selection task, the Llama 3.2 3B Feeling variant selects machines with fewer observations in 31.3\% of cases, compared with 74.3\% for Base (Figure~\ref{fig:opposing_examples}(a)). Adding a Thinking prompt raises this proportion to 66.3\%, but it remains below Base. Under the same prompt, Base selects these machines in 78.6\% of cases, leaving a difference of 12.3 percentage points. The instruction substantially changes behavior, yet the trained Feeling preference continues to influence the decision.

On Choices13k, risk choices show a reversal in Llama 3.2 3B (Figure~\ref{fig:opposing_examples}(b)). The Feeling variant selects alternatives with higher payoff variance in 69.7\% of cases, compared with 54.5\% for Base. Adding a Thinking prompt reduces this proportion to 24.4\%, below Base. This example shows that opposing instructions can reverse behavioral shifts in some situations.

Across the three situational levels, 73.9\% of comparisons are classified as retained and 26.1\% as reversed (Table~\ref{tab:retention}). Reversal occurs when the trained and combined conditions shift the focal response proportion in opposite directions relative to Base. The dominant pattern is therefore retention rather than reversal under opposing instructions.

\begin{table}[!htbp]
\centering
\small
\setlength{\tabcolsep}{3pt}
\begin{tabularx}{\columnwidth}{@{}>{\raggedright\arraybackslash}Xrr@{}}
\toprule
Situational level & Retained & Reversed \\
\midrule
Situational interpretation & 78.6\% & 21.4\% \\
Individual decision making & 68.8\% & 31.3\% \\
Interactive decision making & 78.1\% & 21.9\% \\
\textbf{Overall} & \textbf{73.9\%} & \textbf{26.1\%} \\
\bottomrule
\end{tabularx}
\caption{Retention and reversal across 138 opposing prompt comparisons.}
\label{tab:retention}
\end{table}

These results show that personality tendencies exert a persistent influence on model decisions. Opposing prompts can modify behavior and sometimes reverse particular choice patterns, but they do not consistently override the influence of underlying personality preferences. The differences observed under the same prompt show that this influence extends beyond generation style and immediate instruction effects. Detailed results and classification criteria are provided in Appendix~\ref{app:complete}.

\FloatBarrier

\subsection{Statistical Reliability}
We assess T/F differences across situational interpretation, individual decision making, and interactive decision making using paired tests with Holm correction. The results support personality effects at all three levels, although significance varies across tasks and models. Appendix~\ref{app:statistics} reports effect sizes, confidence intervals, and the complete test results.

\FloatBarrier

%% file: sections/conclusion.tex
\section{Conclusion}

Our experiments show that personality influence in LLMs extends beyond generation style. Using Thinking and Feeling preferences as a controlled contrast, we identify systematic differences in how models interpret information, select actions, and respond during interaction. These preferences continue to influence decisions under opposing instructions, even when prompts substantially change outward behavior. Particular choice patterns can weaken or reverse, but immediate instructions do not consistently override the influence of underlying preferences.

SPACE provides a benchmark framework for examining this influence through a controlled evaluation protocol and complementary datasets. Personality verification and basic capability assessments support comparisons across models with contrasting preferences, while hierarchical situational tasks examine their effects on interpretation, individual decision making, and interactive decision making. Together, the protocol and datasets help distinguish personality influence from measured capability changes and immediate instruction effects.

Personality preferences also have consequences for downstream tasks. They affect interpretation accuracy, choices under uncertainty, responses to other participants, and final interaction outcomes. Their effects vary across tasks and models: a preference that improves performance in one setting can reduce it in another. Personality therefore matters not only for how a model expresses itself, but also for the judgments it reaches, the actions it selects, and the outcomes of its interactions.

These findings establish personality preferences as an important dimension of LLM evaluation and behavioral guidance. Outward personality expression alone does not reveal how a model will judge or act in concrete situations. Evaluating situational behavior and its persistence under competing instructions provides a fuller account of personality influence. These insights may also support more context-sensitive model evaluation and development. SPACE offers a reusable basis for evaluating personality influence and examining how personality preferences can align model behavior with task requirements.

%% file: sections/limitations.tex
\section*{Limitations}
Personality is a complex construct involving diverse behavioral tendencies, interactions, and broader personality schemas. Our study focuses on selected personality dimensions and situational tasks, without exhaustively covering these aspects or their interactions. As an initial step toward understanding personality influence in LLMs, this work lays the groundwork for broader investigations, including more comprehensive personality benchmarking and evaluation frameworks.

Furthermore, personality in this study refers to underlying behavioral preferences and decision-making tendencies exhibited by language models, rather than human psychological traits. Whether and to what extent these tendencies resemble human personality remains an open question requiring further investigation. Our focus is therefore on evaluating observable model behavior and outputs in controlled situational tasks, without making claims about the psychological nature of model personality.

%% file: sections/appendix_details.tex
\section{Data, Training, and Prompt Details}\label{app:details}
\subsection{Task Coverage and Adaptation}\label{app:task_details}
Table~\ref{tab:taskcounts} reports the retained inputs for the principal task versions. In appendix tables, Llama 3B and Llama 8B denote Llama 3.2 3B and Llama 3.1 8B; Qwen 3B and Qwen 7B denote Qwen2.5 3B and Qwen2.5 7B. T and F denote the Thinking and Feeling variants. Multiple slices of one dialogue and numerical variants of one source situation are related observations. These results on adapted tasks are not official benchmark scores on held out data. The controlled probe of buyer offers in Figure~\ref{fig:bargain} is an additional targeted experiment; it is separate from the controlled bargaining task with 900 inputs.
\begin{table*}[!tb]
\centering\small\setlength{\tabcolsep}{4pt}
\begin{tabularx}{\textwidth}{@{}l>{\raggedright\arraybackslash}Xr@{}}
\toprule
Behavioral level & Task version & Inputs per condition \\
\midrule
Interpretation & Circa & 32,602 \\
 & CommitmentBank & 1,009 \\
 & FOMC & 2,421 \\
 & FiQA-SA / SST-2 / News & 1,173 / 10,000 / 11,931 \\
\midrule
Individual decisions & Choices13k (unequal payoff variances) & 1,764 \\
 & Transportation & 210 \\
 & Machine selection & 4,160 \\
 & Delayed rewards & 288 \\
 & Original stock windows & 2,074 \\
 & Controlled stock paths & 1,005 \\
 & Original bargaining with a single turn & 2,334 \\
 & Controlled buyer offers & 900 \\
\midrule
Interactive decisions & INSPIRED & 119 \\
 & Frames & 282 \\
 & Real bargaining histories & 351 \\
 & Bargaining feedback generated by a program & 2,272 \\
\bottomrule
\end{tabularx}
\caption{Task coverage. Questionnaire and capability evaluation details are provided in Appendix~\ref{app:capabilities}.}
\label{tab:taskcounts}
\end{table*}
\subsection{Selection and Task Construction}\label{app:adaptation}
Circa maps indirect replies to YES, NO, CONDITIONAL, or UNCLEAR. CommitmentBank uses the retained source statements and adapted majority references described in Appendix~\ref{app:interpretation}. FOMC retains the frozen monetary policy label mapping. Choices13k includes 1,766 choices, of which 1,764 permit a comparison of payoff variances; human choice rates are not correctness labels.

Transportation presents a faster, more expensive route and a slower, cheaper alternative. Machine selection presents one observation of machine A and three of B, with either one or six choices remaining. Delayed rewards vary amounts and waiting times while guaranteeing both payments. Original stock tasks use 2,074 overlapping windows from 2,102 asset/date records across seven assets. Controlled stock tasks use 67 paths for each of five net changes and three volatility levels, with the final price fixed at \$100.

Original bargaining retains 2,334 initial offer slices from 6,682 CraigslistBargain dialogues. Controlled bargaining uses 150 source situations across five item categories and three price bands, with six synthetic buyer offers from 40\% to 90\% of asking price. Real history tasks retain 351 inputs from 314 dialogues after removing two duplicated prompts; they preserve offer histories but omit other textual negotiation conditions. Program feedback tasks remove 62 initial offers at or above asking price, retaining 2,272 inputs and at most three model choices.

INSPIRED and Frames retain 119 and 282 eligible dialogue prefixes from 999 and 1,369 downloaded dialogues. Short prefixes satisfying the extraction conditions are selected and known input problems are removed; this is not exhaustive manual validation. Their source histories provide recommendation context. The separate controlled feedback probes use two authored versions per domain and are reported separately in Appendix~\ref{app:statistics}.

The expected payoff probe in Figure~\ref{fig:payoff} holds probabilities and payouts fixed and varies their explicit summary; its replication uses 20 economic cells and 80 inputs including both option orders and presentation frames. The buyer offer probe in Figure~\ref{fig:bargain} uses 15 source situations and 315 inputs including offer levels and option orders. These targeted probes are distinct from the full task versions in Table~\ref{tab:taskcounts}.

\subsection{Personality Prompts}\label{app:prompts}
The fixed prefixes used during inference are:
\begin{quote}\small
Use an MBTI Thinking (T) preference: Make decisions using objective logic and consistency. Apply it silently and follow the task's required output format exactly.
\end{quote}
\begin{quote}\small
Use an MBTI Feeling (F) preference: Make decisions using empathy and personal values. Apply it silently and follow the task's required output format exactly.
\end{quote}
An aligned combination uses the same trained and prompted preference; an opposing combination uses the other preference. Prefixes are applied to the task input and reapplied at each model turn in interactive conditions.
\subsection{Preference Pairs}\label{app:pairs}
We pair the opposing Thinking and Feeling responses in the English behavior split by row index. All paired rows have matching instructions and additional context. The targeted response is chosen and the opposing response rejected; switching the target reverses the pair. We concatenate the instruction and any nonempty additional context, then apply the corresponding model's chat template. Rows missing an instruction or either response are excluded; none of the retained files require such exclusions. The T/F training set contains 12,159 preference pairs. The full pair sets are used for training without a held out validation split; situational evaluations supply separate task inputs.

\subsection{Training Details}\label{app:training}
The training recipe uses learning rate $10^{-5}$, DPO $\beta=1.0$, seed 42, batch size per device 16, and gradient accumulation of two. LoRA uses rank 32, scaling 64, and dropout 0.1 on attention projections and feedforward projections. The Base weights provide the frozen reference; the Llama 3.1 8B implementation obtains this reference by disabling its LoRA adapters. Adapters are merged for inference.

Early stopping monitors training DPO loss with a constant learning rate. Monitoring begins after one epoch, averages windows of 100 steps, and stops after two windows without a 5\% relative improvement over the best window. No MBTI or benchmark score selects the stopping point.

\subsection{Computational Resources and Budget}\label{app:compute}
Training and evaluation were run on NVIDIA GH200 GPUs on the Isambard-AI HPC platform, with one GPU allocated per job. Based on available Slurm accounting, the recorded jobs consumed approximately 392.2 GPU-hours in total, including pilot runs, reruns, and unsuccessful attempts. Table~\ref{tab:compute-budget} summarizes the infrastructure and budget.

\begin{table}[H]
\centering\small
\setlength{\tabcolsep}{4pt}
\begin{tabularx}{\columnwidth}{@{}>{\raggedright\arraybackslash}Xl@{}}
\toprule
Item & Value \\
\midrule
GPU model & NVIDIA GH200 \\
GPUs per job & 1 \\
Computing platform & Isambard-AI HPC \\
Training GPU-hours & $\approx 97.9$ \\
Evaluation GPU-hours & $\approx 294.3$ \\
Total GPU-hours & $\approx 392.2$ \\
\bottomrule
\end{tabularx}
\caption{Computational resources and approximate GPU-hours from available Slurm accounting.}
\label{tab:compute-budget}
\end{table}

\FloatBarrier

%% file: sections/appendix_capabilities.tex
\section{Supplementary Personality and Capability Results}\label{app:capabilities}

\subsection{Evaluation Scope and Reporting}\label{app:capability_scope}

Capability evaluation covers four models, two trained preferences, and three benchmarks. Prompt and combined guidance results cover Llama 3.2 3B, Qwen2.5 3B, and Qwen2.5 7B. Unprompted questionnaire evaluation covers all four models, while Prompt and combined questionnaire evaluation covers the first three.

GSM8K, ARC-Easy, and BoolQ contain 1,300, 2,000, and 2,000 evaluated inputs, respectively. The questionnaire uses 46 T/F items. Percentages in the main questionnaire table are calculated from these choices.

Llama 3.2 3B Base, trained variant, and Prompt GSM8K evaluation uses a 512 token limit and counts an answer as matching when any extracted response number matches the reference. Combined GSM8K and Qwen Prompt evaluation extracts the final number from each response. ARC-Easy and BoolQ use label extraction, with unparseable responses counted as incorrect. ARC-Easy uses the full denominator, including 87 reference labels unrecognized by the A to D parser.

\subsection{Trained Variant Capability Results}
Table~\ref{tab:appcap1} supports the broadly stable capability pattern in the main text: 22 of 24 absolute changes are below 1.2 percentage points. The largest decline is Llama 3B T on BoolQ; its statistical evidence is reported in Appendix~\ref{app:stat-capability-current-dpo}.

\begin{table}[!htbp]
\centering\small
\setlength{\tabcolsep}{3pt}
\begin{tabular*}{\columnwidth}{@{\extracolsep{\fill}}llrrr@{}}
\toprule
Model & Condition & ARC-Easy & BoolQ & GSM8K \\
\midrule
Llama 3B & Base & 82.30 & 73.50 & 79.23 \\
 & T & 81.75 & 71.25 & 78.54 \\
 & F & 82.15 & 75.15 & 78.38 \\
Qwen 3B & Base & 89.00 & 79.60 & 11.54 \\
 & T & 88.80 & 79.90 & 11.00 \\
 & F & 88.90 & 78.75 & 12.31 \\
Qwen 7B & Base & 91.35 & 86.10 & 15.54 \\
 & T & 91.30 & 86.30 & 16.00 \\
 & F & 91.20 & 85.90 & 14.92 \\
Llama 8B & Base & 87.60 & 84.50 & 87.31 \\
 & T & 87.55 & 84.45 & 87.23 \\
 & F & 86.75 & 84.10 & 86.15 \\
\bottomrule
\end{tabular*}
\caption{Base and trained variant capability accuracy (\%).}
\label{tab:appcap1}
\end{table}

\subsection{Prompt Capability Results}\label{app:prompt_capabilities}
Prompting lowers ARC-Easy accuracy in all six evaluated conditions, while BoolQ and GSM8K changes are less uniform. These results show that personality instructions can also affect objective task performance.

\begin{table}[!htbp]
\centering\small
\setlength{\tabcolsep}{3pt}
\begin{tabular*}{\columnwidth}{@{\extracolsep{\fill}}llrrr@{}}
\toprule
Model & Condition & ARC-Easy & BoolQ & GSM8K \\
\midrule
Llama 3B & Base P & 82.25 & 73.50 & 79.23 \\
 & Prompt T & 78.55 & 72.70 & 77.69 \\
 & Prompt F & 76.60 & 73.70 & 77.38 \\
Qwen 3B & Base P & 89.00 & 79.60 & 11.77 \\
 & Prompt T & 87.05 & 79.25 & 11.46 \\
 & Prompt F & 87.00 & 79.35 & 11.54 \\
Qwen 7B & Base P & 91.35 & 86.10 & 15.54 \\
 & Prompt T & 90.50 & 82.05 & 15.31 \\
 & Prompt F & 90.00 & 82.25 & 15.31 \\
\bottomrule
\end{tabular*}
\caption{Prompt capability accuracy (\%).}
\label{tab:appcap2}
\end{table}

\subsection{Personality Scores under Combined Guidance}\label{app:combined_alignment}

In X+Y, X denotes the trained Thinking or Feeling variant and Y the personality prompt. T+T and F+F are aligned; T+F and F+T are opposing. Unprompted variant scores are included as the reference. Combined prompts can substantially shift questionnaire choices toward the requested preference. The questionnaire checks whether the guidance produces the intended response patterns. Persistence of personality influence is assessed through situational tasks in Appendix~\ref{app:complete}.

\begin{table}[!htbp]
\centering\small
\setlength{\tabcolsep}{3pt}
\begin{tabular*}{\columnwidth}{@{\extracolsep{\fill}}llrr@{}}
\toprule
Model & Condition & F & T \\
\midrule
Llama 3B & T & 39.1 & 60.9 \\
 & F & 82.6 & 17.4 \\
 & T+T & 6.5 & 93.5 \\
 & T+F & 91.3 & 8.7 \\
 & F+T & 10.9 & 89.1 \\
 & F+F & 93.5 & 6.5 \\
Qwen 3B & T & 30.4 & 69.6 \\
 & F & 54.3 & 45.7 \\
 & T+T & 2.2 & 97.8 \\
 & T+F & 84.8 & 15.2 \\
 & F+T & 13.0 & 87.0 \\
 & F+F & 93.5 & 6.5 \\
Qwen 7B & T & 41.3 & 58.7 \\
 & F & 56.5 & 43.5 \\
 & T+T & 4.3 & 95.7 \\
 & T+F & 95.7 & 4.3 \\
 & F+T & 2.2 & 97.8 \\
 & F+F & 95.7 & 4.3 \\
\bottomrule
\end{tabular*}
\caption{T/F questionnaire choice proportions (\%) for trained and combined conditions.}
\label{tab:appcap3}
\end{table}

\subsection{Capability Scores under Combined Guidance}
Adding either personality prompt increases GSM8K scores relative to the corresponding unprompted trained variant, whereas ARC-Easy and BoolQ often decline.

\begin{table}[!htbp]
\centering\small
\setlength{\tabcolsep}{3pt}
\begin{tabular*}{\columnwidth}{@{\extracolsep{\fill}}llrrr@{}}
\toprule
Model & Condition & ARC-Easy & BoolQ & GSM8K \\
\midrule
Llama 3B & T & 82.05 & 68.85 & 35.00 \\
 & F & 81.70 & 76.05 & 13.08 \\
 & T+T & 78.20 & 67.45 & 54.92 \\
 & T+F & 77.60 & 69.05 & 54.31 \\
 & F+T & 80.10 & 75.20 & 55.62 \\
 & F+F & 75.85 & 75.90 & 52.77 \\
Qwen 3B & T & 88.30 & 79.20 & 7.00 \\
 & F & 88.20 & 79.90 & 7.08 \\
 & T+T & 87.10 & 79.40 & 10.92 \\
 & T+F & 86.95 & 79.15 & 11.69 \\
 & F+T & 87.30 & 78.75 & 9.69 \\
 & F+F & 87.20 & 78.75 & 10.08 \\
Qwen 7B & T & 91.15 & 86.70 & 9.38 \\
 & F & 91.05 & 86.60 & 9.38 \\
 & T+T & 90.40 & 81.95 & 15.23 \\
 & T+F & 89.70 & 82.20 & 14.62 \\
 & F+T & 90.50 & 82.05 & 14.85 \\
 & F+F & 89.70 & 82.45 & 14.69 \\
\bottomrule
\end{tabular*}
\caption{Capability accuracy (\%) under combined guidance.}
\label{tab:appcap4}
\end{table}

\subsection{Llama 3.1 8B Questionnaire Results}
The unprompted questionnaire results confirm the intended T/F shifts for Llama 3.1 8B. Base has 43.5\% F and 56.5\% T choices; F has 58.7\% F and 41.3\% T, while T has 39.1\% F and 60.9\% T. Prompt and combined questionnaire evaluations are available for the other three models.

%% file: sections/appendix_interpretation.tex
\begin{table*}[!tb]
\begin{minipage}[t]{\columnwidth}
\normalsize
\section{Interpretation Metrics and Output Processing}\label{app:interpretation}
Distribution denominators include all retained inputs. Labels are extracted from an explicit leading task label; a valid leading label followed by an explanation is included. This differs from strict compliance with the instruction to output a single label. Unparseable outputs remain in distribution denominators and are counted as incorrect when a reference exists.
Circa and FOMC use the frozen mappings of published annotations. CommitmentBank uses an adapted strict majority reference based on individual rating signs; this is not an original gold label with three classes. Of its 1,009 inputs, 858 have such a reference and enter accuracy and macro F1, while 151 enter distributions only. Macro F1 averages predefined task classes. Table~\ref{tab:interpretationmetrics} reports both metrics.
\end{minipage}\hfill
\begin{minipage}[t]{\columnwidth}
\normalsize
\subsection{CommitmentBank Format Audit}
For Llama 3.1 8B, 970 of 1,009 F guided outputs explicitly begin with UNCERTAIN (96.13\%); parser failures are not mapped to UNCERTAIN. Only 77 F outputs obey the strict format requiring a single label, compared with 580 Base outputs and 819 T outputs. All 932 F outputs with an additional explanation begin with UNCERTAIN. Among the 75 inputs strictly formatted in all three conditions, UNCERTAIN counts are 36 for Base, 27 for T, and 37 for F. Thus the distribution over all inputs is an observed label pattern, but it does not establish an effect independent of output format. The subset comparison does not identify formatting as its cause.
\end{minipage}
\par\vspace{8pt}
\centering\small\setlength{\tabcolsep}{4pt}
\begin{tabular*}{\textwidth}{@{\extracolsep{\fill}}lrrr@{}}
\toprule
Dataset & Model & \shortstack{Accuracy (\%)\\Base / T / F} & \shortstack{Macro F1\\Base / T / F} \\
\midrule
Circa & Llama 3.2 3B & 58.54 / 70.52 / 23.60 & 0.4914 / 0.5617 / 0.2630 \\
Circa & Qwen2.5 3B & 12.10 / 13.17 / 8.62 & 0.1581 / 0.1690 / 0.1062 \\
Circa & Qwen2.5 7B & 27.27 / 29.30 / 23.95 & 0.3270 / 0.3498 / 0.2953 \\
Circa & Llama 3.1 8B & 27.54 / 41.77 / 13.13 & 0.2597 / 0.3743 / 0.1514 \\
CommitmentBank & Llama 3.2 3B & 48.02 / 45.57 / 55.01 & 0.3131 / 0.2654 / 0.4033 \\
CommitmentBank & Qwen2.5 3B & 36.01 / 40.79 / 25.99 & 0.3265 / 0.3561 / 0.2445 \\
CommitmentBank & Qwen2.5 7B & 66.20 / 67.13 / 60.61 & 0.6416 / 0.6487 / 0.5946 \\
CommitmentBank & Llama 3.1 8B & 33.22 / 55.48 / 16.78 & 0.3456 / 0.5390 / 0.1404 \\
FOMC & Llama 3.2 3B & 51.80 / 51.05 / 52.75 & 0.4000 / 0.4105 / 0.3879 \\
FOMC & Qwen2.5 3B & 56.67 / 57.37 / 56.18 & 0.4442 / 0.4584 / 0.4288 \\
FOMC & Qwen2.5 7B & 63.07 / 63.61 / 61.63 & 0.5734 / 0.5768 / 0.5590 \\
FOMC & Llama 3.1 8B & 59.48 / 61.46 / 56.63 & 0.4828 / 0.5355 / 0.4100 \\
\bottomrule
\end{tabular*}
\caption{Situational interpretation: Accuracy (\%) and macro F1 for Base / T / F, in that order. CommitmentBank uses 858 inputs with a majority reference.}
\label{tab:interpretationmetrics}
\end{table*}

\begin{table}[!htbp]
\begin{minipage}{\linewidth}
\normalsize
\subsection{Full Interpretation Label Distributions}\label{app:interpretation_distributions}
All retained inputs remain in the denominator, including unparseable outputs. The tables report the same leading label parser used in the main figures.

On Circa, T reduces UNCLEAR judgments and F increases them across all four models. The full distributions show how definite and conditional interpretations account for the remaining responses, extending the contrast discussed in the main text.
\end{minipage}
\par\vspace{3pt}
\centering\fontsize{9}{10}\selectfont\setlength{\tabcolsep}{3pt}
\begin{tabular*}{\columnwidth}{@{\extracolsep{\fill}}llrrr@{}}
\toprule
Model & Label & Base & T & F \\
\midrule
Llama 3B & UNCLEAR & 23.90 & 2.66 & 72.40 \\
Llama 3B & NO & 19.76 & 45.06 & 0.44 \\
Llama 3B & YES & 41.60 & 37.06 & 17.59 \\
Llama 3B & CONDITIONAL & 14.74 & 15.22 & 9.57 \\
Llama 3B & None & 0.00 & 0.00 & 0.00 \\
Llama 8B & UNCLEAR & 27.85 & 10.72 & 68.35 \\
Llama 8B & NO & 8.88 & 20.19 & 0.80 \\
Llama 8B & YES & 10.35 & 15.09 & 2.97 \\
Llama 8B & CONDITIONAL & 52.91 & 54.00 & 27.88 \\
Llama 8B & None & 0.00 & 0.00 & 0.01 \\
Qwen 3B & UNCLEAR & 89.31 & 87.87 & 93.16 \\
Qwen 3B & NO & 4.78 & 5.13 & 2.73 \\
Qwen 3B & YES & 4.03 & 4.79 & 2.71 \\
Qwen 3B & CONDITIONAL & 1.86 & 2.21 & 1.32 \\
Qwen 3B & None & 0.02 & 0.00 & 0.08 \\
Qwen 7B & UNCLEAR & 64.41 & 63.59 & 67.12 \\
Qwen 7B & NO & 12.12 & 13.83 & 9.59 \\
Qwen 7B & YES & 6.50 & 7.05 & 5.41 \\
Qwen 7B & CONDITIONAL & 16.89 & 15.48 & 17.75 \\
Qwen 7B & None & 0.08 & 0.06 & 0.13 \\
\bottomrule
\end{tabular*}
\caption{Situational interpretation: Circa: full prediction distributions (\%).}
\label{tab:interpretation-full-circa_full}
\end{table}
\begin{table}[!htbp]
\begin{minipage}{\linewidth}
\normalsize
On CommitmentBank, T increases FALSE judgments and F decreases them across models. The large UNCERTAIN proportion for Llama 8B F should be read alongside the format audit above; label distributions and strict format compliance capture different aspects of its responses.
\end{minipage}
\par\vspace{3pt}
\centering\fontsize{9}{10}\selectfont\setlength{\tabcolsep}{3pt}
\begin{tabular*}{\columnwidth}{@{\extracolsep{\fill}}llrrr@{}}
\toprule
Model & Label & Base & T & F \\
\midrule
Llama 3B & FALSE & 84.24 & 92.86 & 55.90 \\
Llama 3B & TRUE & 15.76 & 7.14 & 43.11 \\
Llama 3B & UNCERTAIN & 0.00 & 0.00 & 0.99 \\
Llama 8B & FALSE & 12.39 & 30.82 & 1.19 \\
Llama 8B & TRUE & 8.72 & 20.22 & 2.68 \\
Llama 8B & UNCERTAIN & 78.89 & 48.96 & 96.13 \\
Qwen 3B & FALSE & 25.57 & 31.81 & 13.97 \\
Qwen 3B & TRUE & 0.40 & 0.50 & 0.10 \\
Qwen 3B & UNCERTAIN & 74.03 & 67.69 & 85.93 \\
Qwen 7B & FALSE & 25.97 & 27.16 & 20.52 \\
Qwen 7B & TRUE & 29.63 & 29.44 & 29.24 \\
Qwen 7B & UNCERTAIN & 44.40 & 43.41 & 50.25 \\
\bottomrule
\end{tabular*}
\caption{Situational interpretation: CommitmentBank: full prediction distributions (\%).}
\label{tab:interpretation-full-commitmentbank_full}
\end{table}
\begin{table}[!htbp]
\begin{minipage}{\linewidth}
\normalsize
FOMC extends interpretation to monetary policy language. For Llama 3B, DOVISH judgments rise from 45.39\% at Base to 52.29\% for T and fall to 35.11\% for F. Other models redistribute labels differently, so these results describe policy judgments rather than a common sentiment scale.
\end{minipage}
\par\vspace{3pt}
\centering\fontsize{9}{10}\selectfont\setlength{\tabcolsep}{3pt}
\begin{tabular*}{\columnwidth}{@{\extracolsep{\fill}}llrrr@{}}
\toprule
Model & Label & Base & T & F \\
\midrule
Llama 3B & NEUTRAL & 53.61 & 46.22 & 64.23 \\
Llama 3B & DOVISH & 45.39 & 52.29 & 35.11 \\
Llama 3B & HAWKISH & 0.99 & 1.49 & 0.66 \\
Llama 8B & NEUTRAL & 83.97 & 76.04 & 90.13 \\
Llama 8B & DOVISH & 6.86 & 11.11 & 2.60 \\
Llama 8B & HAWKISH & 9.17 & 12.85 & 7.27 \\
Qwen 3B & NEUTRAL & 81.62 & 80.83 & 84.18 \\
Qwen 3B & DOVISH & 10.74 & 10.24 & 9.25 \\
Qwen 3B & HAWKISH & 7.64 & 8.92 & 6.57 \\
Qwen 7B & NEUTRAL & 54.52 & 59.23 & 50.19 \\
Qwen 7B & DOVISH & 38.70 & 33.87 & 43.49 \\
Qwen 7B & HAWKISH & 6.77 & 6.90 & 6.32 \\
\bottomrule
\end{tabular*}
\caption{Situational interpretation: FOMC: full prediction distributions (\%).}
\label{tab:interpretation-full-fomc_full}
\end{table}
\clearpage

%% file: sections/appendix_complete.tex
\section{Complete Comparisons and Additional Results}\label{app:complete}
\subsection{Retention by Task}
Table~\ref{tab:taskretention} reports every selected task version. FiQA-SA, SST-2, and News cover three models; the remaining task versions cover four models. Each comparison specifies a model, task, and trained preference and compares Base, trained variant, and combined conditions. Original and controlled stock or bargaining versions belong to related task families.

\begin{table}[!htbp]
\centering\small\setlength{\tabcolsep}{3pt}
\begin{tabularx}{\columnwidth}{@{}>{\raggedright\arraybackslash}X>{\raggedright\arraybackslash}p{.27\columnwidth}rr@{}}
\toprule
Task version & Focal label & Retained & Reversed \\
\midrule
CommitmentBank & FALSE & 87.5\% & 12.5\% \\
Circa & UNCLEAR & 75.0\% & 25.0\% \\
FOMC & HAWKISH & 62.5\% & 37.5\% \\
FiQA-SA & positive & 100.0\% & 0.0\% \\
SST-2 & positive & 83.3\% & 16.7\% \\
News & bullish & 66.7\% & 33.3\% \\
Risk choice & Higher variance & 37.5\% & 62.5\% \\
Machine choice & Fewer observations & 62.5\% & 37.5\% \\
Travel: time and cost & Faster route & 75.0\% & 25.0\% \\
Delayed payments & SOONER & 75.0\% & 25.0\% \\
Original stock & BUY & 62.5\% & 37.5\% \\
Original bargaining with a single turn & EXIT & 75.0\% & 25.0\% \\
Controlled stock & BUY & 75.0\% & 25.0\% \\
Controlled bargaining & EXIT & 87.5\% & 12.5\% \\
INSPIRED & Ask more & 62.5\% & 37.5\% \\
Frames & ACCEPT & 75.0\% & 25.0\% \\
Real bargaining history & EXIT & 87.5\% & 12.5\% \\
Program bargaining: FIRST action & EXIT & 87.5\% & 12.5\% \\
\bottomrule
\end{tabularx}
\caption{Retention and reversal across all selected task versions.}
\label{tab:taskretention}
\end{table}

\subsection{Directional Criterion}\label{app:retention_definition}
For the proportion of a focal label $p$, define $\Delta_D=p_{\mathrm{trained}}-p_{\mathrm{Base}}$ and $\Delta_C=p_{\mathrm{combined}}-p_{\mathrm{Base}}$. Reversal requires $\Delta_D\Delta_C<0$; otherwise the result is retained. Signs use unrounded counts. Zero initial or combined shifts count as retained by convention. The total comprises 102 retained and 36 reversed results out of 138. Retention does not imply unchanged individual answers or unchanged effect magnitude. Decision and interaction labels are behavioral measures; human choice frequencies and simulated outcomes are not objective correctness labels.

\subsection{Complete Opposing Prompt Comparisons}\label{app:opposing_full}
The complete comparisons include both partial attenuation and reversal. A retained direction can coexist with a substantial change in magnitude, as the machine choice example shows. Reporting Base, prompting, and combined conditions together makes the remaining training influence visible beyond the binary retention summary.

The following tables report the focal label proportion for every model, task, and trained preference. Prompt is applied to Base; Opposed applies that prompt to the trained variant. Each row uses unrounded counts for classification.
\begin{table*}[!tbp]
\centering\fontsize{9}{10}\selectfont\setlength{\tabcolsep}{2pt}\renewcommand{\arraystretch}{0.96}
\begin{tabular*}{\textwidth}{@{\extracolsep{\fill}}llcrrrrl@{}}
\toprule
Task & Model & Variant & Base & Trained & Prompt & Opposed & Outcome \\
\midrule
CommitmentBank & Llama 3B & T & 84.24 & 92.86 & 66.20 & 89.40 & Retained \\
CommitmentBank & Llama 3B & F & 84.24 & 55.90 & 83.55 & 50.74 & Retained \\
CommitmentBank & Llama 8B & T & 12.39 & 30.82 & 25.17 & 54.01 & Retained \\
CommitmentBank & Llama 8B & F & 12.39 & 1.19 & 45.39 & 2.87 & Retained \\
CommitmentBank & Qwen 3B & T & 25.57 & 31.81 & 26.16 & 32.61 & Retained \\
CommitmentBank & Qwen 3B & F & 25.57 & 13.97 & 29.34 & 20.02 & Retained \\
CommitmentBank & Qwen 7B & T & 25.97 & 27.16 & 15.86 & 18.83 & Reversed \\
CommitmentBank & Qwen 7B & F & 25.97 & 20.52 & 26.16 & 19.23 & Retained \\
Circa & Llama 3B & T & 23.90 & 2.66 & 5.82 & 1.73 & Retained \\
Circa & Llama 3B & F & 23.90 & 72.40 & 3.26 & 33.15 & Retained \\
Circa & Llama 8B & T & 27.85 & 10.72 & 10.32 & 0.29 & Retained \\
Circa & Llama 8B & F & 27.85 & 68.35 & 2.87 & 40.94 & Retained \\
Circa & Qwen 3B & T & 89.31 & 87.87 & 88.23 & 85.45 & Retained \\
Circa & Qwen 3B & F & 89.31 & 93.16 & 81.66 & 87.63 & Reversed \\
Circa & Qwen 7B & T & 64.41 & 63.59 & 49.60 & 48.16 & Retained \\
Circa & Qwen 7B & F & 64.41 & 67.12 & 35.21 & 38.37 & Reversed \\
FOMC & Llama 3B & T & 0.99 & 1.49 & 0.45 & 0.37 & Reversed \\
FOMC & Llama 3B & F & 0.99 & 0.66 & 2.15 & 0.99 & Retained \\
FOMC & Llama 8B & T & 9.17 & 12.85 & 14.04 & 16.27 & Retained \\
FOMC & Llama 8B & F & 9.17 & 7.27 & 25.73 & 18.09 & Reversed \\
FOMC & Qwen 3B & T & 7.64 & 8.92 & 8.47 & 11.03 & Retained \\
FOMC & Qwen 3B & F & 7.64 & 6.57 & 9.05 & 7.35 & Retained \\
FOMC & Qwen 7B & T & 6.77 & 6.90 & 4.34 & 4.63 & Reversed \\
FOMC & Qwen 7B & F & 6.77 & 6.32 & 5.66 & 5.20 & Retained \\
FiQA-SA & Llama 3B & T & 36.49 & 17.05 & 28.64 & 12.36 & Retained \\
FiQA-SA & Llama 3B & F & 36.49 & 53.62 & 22.51 & 39.90 & Retained \\
FiQA-SA & Qwen 3B & T & 47.74 & 45.52 & 43.73 & 40.24 & Retained \\
FiQA-SA & Qwen 3B & F & 47.74 & 54.31 & 44.50 & 50.47 & Retained \\
FiQA-SA & Qwen 7B & T & 41.60 & 38.96 & 39.56 & 35.38 & Retained \\
FiQA-SA & Qwen 7B & F & 41.60 & 45.44 & 36.66 & 41.60 & Retained \\
SST-2 & Llama 3B & T & 39.56 & 23.88 & 39.82 & 23.20 & Retained \\
SST-2 & Llama 3B & F & 39.56 & 50.96 & 36.67 & 51.61 & Retained \\
SST-2 & Qwen 3B & T & 31.86 & 28.88 & 34.91 & 30.19 & Retained \\
SST-2 & Qwen 3B & F & 31.86 & 35.59 & 30.80 & 34.63 & Retained \\
SST-2 & Qwen 7B & T & 45.56 & 43.34 & 42.74 & 39.48 & Retained \\
SST-2 & Qwen 7B & F & 45.56 & 47.46 & 37.43 & 39.71 & Reversed \\
News & Llama 3B & T & 42.61 & 29.25 & 48.14 & 34.78 & Retained \\
News & Llama 3B & F & 42.61 & 55.41 & 43.19 & 63.85 & Retained \\
News & Qwen 3B & T & 22.89 & 22.88 & 27.89 & 26.10 & Reversed \\
News & Qwen 3B & F & 22.89 & 25.96 & 21.33 & 24.73 & Retained \\
News & Qwen 7B & T & 30.01 & 27.78 & 30.01 & 26.45 & Retained \\
News & Qwen 7B & F & 30.01 & 34.05 & 25.56 & 29.27 & Reversed \\
\bottomrule
\end{tabular*}
\caption{Interpretation: complete opposing prompt comparisons (\%).}
\label{tab:opposed-interpretation-0}
\label{tab:opposed-interpretation-20}
\label{tab:opposed-interpretation-40}
\end{table*}
\begin{table*}[!tbp]
\centering\fontsize{9}{10}\selectfont\setlength{\tabcolsep}{2pt}\renewcommand{\arraystretch}{0.96}
\begin{tabular*}{\textwidth}{@{\extracolsep{\fill}}llcrrrrl@{}}
\toprule
Task & Model & Variant & Base & Trained & Prompt & Opposed & Outcome \\
\midrule
Risk choice & Llama 3B & T & 54.48 & 50.68 & 22.96 & 22.96 & Retained \\
Risk choice & Llama 3B & F & 54.48 & 69.73 & 22.90 & 24.38 & Reversed \\
Risk choice & Llama 8B & T & 48.53 & 47.22 & 41.50 & 39.34 & Retained \\
Risk choice & Llama 8B & F & 48.53 & 69.78 & 29.65 & 46.03 & Reversed \\
Risk choice & Qwen 3B & T & 71.15 & 67.74 & 75.28 & 74.09 & Reversed \\
Risk choice & Qwen 3B & F & 71.15 & 75.45 & 72.79 & 74.32 & Retained \\
Risk choice & Qwen 7B & T & 75.17 & 76.36 & 37.76 & 34.64 & Reversed \\
Risk choice & Qwen 7B & F & 75.17 & 75.45 & 42.52 & 45.69 & Reversed \\
Machine choice & Llama 3B & T & 74.33 & 84.18 & 63.32 & 76.01 & Retained \\
Machine choice & Llama 3B & F & 74.33 & 31.30 & 78.56 & 66.30 & Retained \\
Machine choice & Llama 8B & T & 20.91 & 37.84 & 19.47 & 28.17 & Retained \\
Machine choice & Llama 8B & F & 20.91 & 3.37 & 49.42 & 35.62 & Reversed \\
Machine choice & Qwen 3B & T & 49.04 & 57.50 & 46.97 & 61.15 & Retained \\
Machine choice & Qwen 3B & F & 49.04 & 40.00 & 57.36 & 47.07 & Retained \\
Machine choice & Qwen 7B & T & 34.04 & 37.45 & 33.37 & 32.45 & Reversed \\
Machine choice & Qwen 7B & F & 34.04 & 30.58 & 47.60 & 45.14 & Reversed \\
Travel: time and cost & Llama 3B & T & 31.90 & 32.38 & 100.00 & 100.00 & Retained \\
Travel: time and cost & Llama 3B & F & 31.90 & 0.48 & 100.00 & 10.48 & Retained \\
Travel: time and cost & Llama 8B & T & 100.00 & 100.00 & 100.00 & 100.00 & Retained \\
Travel: time and cost & Llama 8B & F & 100.00 & 100.00 & 100.00 & 100.00 & Retained \\
Travel: time and cost & Qwen 3B & T & 95.71 & 100.00 & 100.00 & 100.00 & Retained \\
Travel: time and cost & Qwen 3B & F & 95.71 & 71.43 & 100.00 & 100.00 & Reversed \\
Travel: time and cost & Qwen 7B & T & 27.14 & 68.57 & 77.14 & 88.57 & Retained \\
Travel: time and cost & Qwen 7B & F & 27.14 & 10.48 & 98.57 & 90.48 & Reversed \\
Delayed payments & Llama 3B & T & 100.00 & 100.00 & 100.00 & 100.00 & Retained \\
Delayed payments & Llama 3B & F & 100.00 & 100.00 & 100.00 & 100.00 & Retained \\
Delayed payments & Llama 8B & T & 100.00 & 100.00 & 100.00 & 100.00 & Retained \\
Delayed payments & Llama 8B & F & 100.00 & 100.00 & 100.00 & 100.00 & Retained \\
Delayed payments & Qwen 3B & T & 44.10 & 58.33 & 100.00 & 100.00 & Retained \\
Delayed payments & Qwen 3B & F & 44.10 & 44.44 & 70.14 & 35.42 & Reversed \\
Delayed payments & Qwen 7B & T & 0.35 & 4.86 & 69.44 & 73.26 & Retained \\
Delayed payments & Qwen 7B & F & 0.35 & 0.00 & 7.99 & 0.69 & Reversed \\
Original stock & Llama 3B & T & 14.13 & 36.98 & 0.00 & 0.00 & Reversed \\
Original stock & Llama 3B & F & 14.13 & 0.00 & 0.00 & 0.00 & Retained \\
Original stock & Llama 8B & T & 84.67 & 89.30 & 53.76 & 61.67 & Reversed \\
Original stock & Llama 8B & F & 84.67 & 63.89 & 52.27 & 47.30 & Retained \\
Original stock & Qwen 3B & T & 0.10 & 0.72 & 0.00 & 0.00 & Reversed \\
Original stock & Qwen 3B & F & 0.10 & 0.00 & 0.00 & 0.00 & Retained \\
Original stock & Qwen 7B & T & 28.35 & 27.68 & 0.53 & 0.29 & Retained \\
Original stock & Qwen 7B & F & 28.35 & 29.94 & 62.10 & 60.03 & Retained \\
Original single-turn bargaining & Llama 3B & T & 0.17 & 0.00 & 1.07 & 0.00 & Retained \\
Original single-turn bargaining & Llama 3B & F & 0.17 & 1.16 & 1.67 & 23.86 & Retained \\
Original single-turn bargaining & Llama 8B & T & 0.51 & 0.04 & 0.09 & 0.00 & Retained \\
Original single-turn bargaining & Llama 8B & F & 0.51 & 0.60 & 0.00 & 0.09 & Reversed \\
Original single-turn bargaining & Qwen 3B & T & 1.97 & 0.47 & 0.73 & 0.39 & Retained \\
Original single-turn bargaining & Qwen 3B & F & 1.97 & 15.72 & 0.86 & 11.31 & Retained \\
Original single-turn bargaining & Qwen 7B & T & 0.26 & 0.17 & 3.34 & 0.81 & Reversed \\
Original single-turn bargaining & Qwen 7B & F & 0.26 & 0.26 & 0.56 & 0.99 & Retained \\
Controlled stock & Llama 3B & T & 41.89 & 43.18 & 0.30 & 0.00 & Reversed \\
Controlled stock & Llama 3B & F & 41.89 & 13.23 & 0.00 & 0.00 & Retained \\
Controlled stock & Llama 8B & T & 96.12 & 95.62 & 90.75 & 93.53 & Retained \\
Controlled stock & Llama 8B & F & 96.12 & 59.60 & 91.24 & 50.15 & Retained \\
Controlled stock & Qwen 3B & T & 0.00 & 0.00 & 0.00 & 0.00 & Retained \\
Controlled stock & Qwen 3B & F & 0.00 & 0.00 & 0.00 & 0.00 & Retained \\
Controlled stock & Qwen 7B & T & 42.49 & 44.08 & 52.94 & 51.24 & Retained \\
Controlled stock & Qwen 7B & F & 42.49 & 41.99 & 48.06 & 46.97 & Reversed \\
Controlled bargaining & Llama 3B & T & 0.00 & 0.00 & 0.22 & 0.00 & Retained \\
Controlled bargaining & Llama 3B & F & 0.00 & 0.11 & 0.22 & 27.56 & Retained \\
Controlled bargaining & Llama 8B & T & 0.00 & 0.00 & 0.00 & 0.00 & Retained \\
Controlled bargaining & Llama 8B & F & 0.00 & 0.00 & 0.00 & 0.00 & Retained \\
Controlled bargaining & Qwen 3B & T & 8.89 & 0.89 & 0.00 & 0.00 & Retained \\
Controlled bargaining & Qwen 3B & F & 8.89 & 28.00 & 0.00 & 2.33 & Reversed \\
Controlled bargaining & Qwen 7B & T & 0.00 & 0.00 & 4.44 & 0.56 & Retained \\
Controlled bargaining & Qwen 7B & F & 0.00 & 0.00 & 0.00 & 3.11 & Retained \\
\bottomrule
\end{tabular*}
\caption{Decision: complete opposing prompt comparisons (\%).}
\label{tab:opposed-decision-0}
\label{tab:opposed-decision-20}
\label{tab:opposed-decision-40}
\label{tab:opposed-decision-60}
\end{table*}
\begin{table*}[!tbp]
\centering\fontsize{9}{10}\selectfont\setlength{\tabcolsep}{2pt}\renewcommand{\arraystretch}{0.96}
\begin{tabular*}{\textwidth}{@{\extracolsep{\fill}}llcrrrrl@{}}
\toprule
Task & Model & Variant & Base & Trained & Prompt & Opposed & Outcome \\
\midrule
INSPIRED & Llama 3B & T & 21.01 & 5.88 & 5.88 & 0.84 & Retained \\
INSPIRED & Llama 3B & F & 21.01 & 74.79 & 10.08 & 56.30 & Retained \\
INSPIRED & Llama 8B & T & 36.13 & 20.17 & 78.15 & 63.03 & Reversed \\
INSPIRED & Llama 8B & F & 36.13 & 84.03 & 64.71 & 91.60 & Retained \\
INSPIRED & Qwen 3B & T & 86.55 & 81.51 & 90.76 & 87.39 & Reversed \\
INSPIRED & Qwen 3B & F & 86.55 & 91.60 & 73.95 & 87.39 & Retained \\
INSPIRED & Qwen 7B & T & 64.71 & 64.71 & 52.94 & 52.94 & Retained \\
INSPIRED & Qwen 7B & F & 64.71 & 68.07 & 11.76 & 14.29 & Reversed \\
Frames & Llama 3B & T & 29.43 & 39.01 & 34.40 & 56.74 & Retained \\
Frames & Llama 3B & F & 29.43 & 18.79 & 20.57 & 15.60 & Retained \\
Frames & Llama 8B & T & 2.48 & 3.19 & 0.71 & 1.06 & Reversed \\
Frames & Llama 8B & F & 2.48 & 0.35 & 2.84 & 0.35 & Retained \\
Frames & Qwen 3B & T & 1.42 & 2.84 & 6.74 & 6.03 & Retained \\
Frames & Qwen 3B & F & 1.42 & 1.42 & 10.28 & 8.87 & Retained \\
Frames & Qwen 7B & T & 27.30 & 28.37 & 29.43 & 28.37 & Retained \\
Frames & Qwen 7B & F & 27.30 & 26.24 & 45.39 & 42.20 & Reversed \\
Real bargaining history & Llama 3B & T & 0.00 & 0.00 & 0.00 & 0.00 & Retained \\
Real bargaining history & Llama 3B & F & 0.00 & 0.00 & 0.00 & 1.14 & Retained \\
Real bargaining history & Llama 8B & T & 0.00 & 0.00 & 0.00 & 0.00 & Retained \\
Real bargaining history & Llama 8B & F & 0.00 & 0.00 & 0.00 & 0.00 & Retained \\
Real bargaining history & Qwen 3B & T & 85.19 & 27.35 & 11.11 & 0.00 & Retained \\
Real bargaining history & Qwen 3B & F & 85.19 & 99.72 & 4.27 & 84.05 & Reversed \\
Real bargaining history & Qwen 7B & T & 0.00 & 0.00 & 0.00 & 0.00 & Retained \\
Real bargaining history & Qwen 7B & F & 0.00 & 0.00 & 0.00 & 0.00 & Retained \\
Program bargaining: FIRST action & Llama 3B & T & 0.18 & 0.00 & 1.06 & 0.00 & Retained \\
Program bargaining: FIRST action & Llama 3B & F & 0.18 & 1.23 & 1.67 & 24.47 & Retained \\
Program bargaining: FIRST action & Llama 8B & T & 0.04 & 0.00 & 0.09 & 0.00 & Retained \\
Program bargaining: FIRST action & Llama 8B & F & 0.04 & 0.04 & 0.00 & 0.04 & Retained \\
Program bargaining: FIRST action & Qwen 3B & T & 1.54 & 0.04 & 0.62 & 0.09 & Retained \\
Program bargaining: FIRST action & Qwen 3B & F & 1.54 & 14.96 & 0.70 & 10.39 & Retained \\
Program bargaining: FIRST action & Qwen 7B & T & 0.26 & 0.18 & 3.43 & 0.84 & Reversed \\
Program bargaining: FIRST action & Qwen 7B & F & 0.26 & 0.26 & 0.57 & 1.01 & Retained \\
\bottomrule
\end{tabular*}
\caption{Interaction: complete opposing prompt comparisons (\%).}
\label{tab:opposed-interaction-0}
\label{tab:opposed-interaction-20}
\end{table*}

\begin{table}[!htbp]
\begin{minipage}{\linewidth}
\normalsize
\subsection{Supplementary Situational Results}\label{app:situational_full}
The additional tasks include clear preference differences and unchanged choices. In transportation, F favors the slower, cheaper option more than T in three models, while Llama 8B always chooses the faster option. Keeping such unchanged cases shows the full task coverage rather than only the largest contrasts.

We report Base/T/F label distributions for the full task versions, including tasks with small or zero changes. The risk table retains all 1,766 A/B choices; the variance comparison in the main figure uses 1,764 comparable choices. Interpretation performance is in Appendix~\ref{app:interpretation}. Choice labels and program outcomes are behavioral measures, not objective correctness scores.

The supplementary results reveal task- and model-dependent behavioral differences across decision-making settings. While some tasks exhibit substantial shifts between T and F variants, others show limited or no changes.

\end{minipage}
\par\vspace{3pt}
\centering\fontsize{9}{10}\selectfont\setlength{\tabcolsep}{3pt}
\begin{tabular*}{\columnwidth}{@{\extracolsep{\fill}}llrrr@{}}
\toprule
Model & Label & Base & T & F \\
\midrule
Llama 3B & A & 41.56 & 61.66 & 8.55 \\
Llama 3B & B & 58.44 & 38.34 & 91.45 \\
Qwen 3B & B & 73.95 & 66.59 & 80.41 \\
Qwen 3B & A & 26.05 & 33.41 & 19.59 \\
Qwen 7B & B & 86.35 & 82.90 & 88.90 \\
Qwen 7B & A & 13.65 & 17.10 & 11.10 \\
Llama 8B & A & 44.28 & 51.47 & 15.29 \\
Llama 8B & B & 55.72 & 48.53 & 84.71 \\
\bottomrule
\end{tabular*}
\caption{Individual decision making: Risk choices (A/B): complete label distributions (\%), $N=1,766$.}
\label{tab:full-risk}
\end{table}
\begin{table}[!htbp]
\centering\fontsize{9}{10}\selectfont\setlength{\tabcolsep}{3pt}
\begin{tabular*}{\columnwidth}{@{\extracolsep{\fill}}llrrr@{}}
\toprule
Model & Label & Base & T & F \\
\midrule
Llama 3B & Faster, costlier & 31.90 & 32.38 & 0.48 \\
Llama 3B & Slower, cheaper & 68.10 & 67.62 & 99.52 \\
Qwen 3B & Faster, costlier & 95.71 & 100.00 & 71.43 \\
Qwen 3B & Slower, cheaper & 4.29 & 0.00 & 28.57 \\
Qwen 7B & Faster, costlier & 27.14 & 68.57 & 10.48 \\
Qwen 7B & Slower, cheaper & 72.86 & 31.43 & 89.52 \\
Llama 8B & Faster, costlier & 100.00 & 100.00 & 100.00 \\
\bottomrule
\end{tabular*}
\caption{Individual decision making: Transportation: complete label distributions (\%), $N=210$.}
\label{tab:full-travel}
\end{table}
\begin{table}[!htbp]
\centering\fontsize{9}{10}\selectfont\setlength{\tabcolsep}{3pt}
\begin{tabular*}{\columnwidth}{@{\extracolsep{\fill}}llrrr@{}}
\toprule
Model & Label & Base & T & F \\
\midrule
Llama 3B & Fewer observations & 74.33 & 84.18 & 31.30 \\
Llama 3B & More observations & 25.67 & 15.82 & 68.70 \\
Qwen 3B & More observations & 50.96 & 42.50 & 60.00 \\
Qwen 3B & Fewer observations & 49.04 & 57.50 & 40.00 \\
Qwen 7B & More observations & 65.96 & 62.55 & 69.42 \\
Qwen 7B & Fewer observations & 34.04 & 37.45 & 30.58 \\
Llama 8B & More observations & 79.09 & 62.16 & 96.63 \\
Llama 8B & Fewer observations & 20.91 & 37.84 & 3.37 \\
\bottomrule
\end{tabular*}
\caption{Individual decision making: Machine choice: complete label distributions (\%), $N=4,160$.}
\label{tab:full-exploration}
\end{table}
\begin{table}[!htbp]
\centering\fontsize{9}{10}\selectfont\setlength{\tabcolsep}{3pt}
\begin{tabular*}{\columnwidth}{@{\extracolsep{\fill}}llrrr@{}}
\toprule
Model & Label & Base & T & F \\
\midrule
Llama 3B & SELL & 38.81 & 26.76 & 8.49 \\
Llama 3B & HOLD & 47.06 & 36.26 & 91.51 \\
Llama 3B & BUY & 14.13 & 36.98 & 0.00 \\
Qwen 3B & SELL & 99.90 & 99.28 & 100.00 \\
Qwen 3B & BUY & 0.10 & 0.72 & 0.00 \\
Qwen 7B & SELL & 19.77 & 19.38 & 19.96 \\
Qwen 7B & BUY & 28.35 & 27.68 & 29.94 \\
Qwen 7B & HOLD & 51.88 & 52.94 & 50.10 \\
Llama 8B & SELL & 12.58 & 9.11 & 6.51 \\
Llama 8B & BUY & 84.67 & 89.30 & 63.89 \\
Llama 8B & HOLD & 2.75 & 1.59 & 29.60 \\
\bottomrule
\end{tabular*}
\caption{Individual decision making: Original stock: complete label distributions (\%), $N=2,074$.}
\label{tab:full-investor}
\end{table}
\begin{table}[!htbp]
\centering\fontsize{9}{10}\selectfont\setlength{\tabcolsep}{3pt}
\begin{tabular*}{\columnwidth}{@{\extracolsep{\fill}}llrrr@{}}
\toprule
Model & Label & Base & T & F \\
\midrule
Llama 3B & COUNTER & 98.11 & 98.54 & 96.32 \\
Llama 3B & ACCEPT & 1.71 & 1.46 & 2.53 \\
Llama 3B & EXIT & 0.17 & 0.00 & 1.16 \\
Qwen 3B & COUNTER & 97.09 & 98.59 & 83.93 \\
Qwen 3B & ACCEPT & 0.94 & 0.94 & 0.34 \\
Qwen 3B & EXIT & 1.97 & 0.47 & 15.72 \\
Qwen 7B & COUNTER & 98.33 & 98.46 & 98.33 \\
Qwen 7B & ACCEPT & 1.41 & 1.37 & 1.41 \\
Qwen 7B & EXIT & 0.26 & 0.17 & 0.26 \\
Llama 8B & COUNTER & 97.56 & 98.03 & 97.73 \\
Llama 8B & ACCEPT & 1.93 & 1.93 & 1.67 \\
Llama 8B & EXIT & 0.51 & 0.04 & 0.60 \\
\bottomrule
\end{tabular*}
\caption{Individual decision making: Original bargaining: complete label distributions (\%), $N=2,334$.}
\label{tab:full-negotiation}
\end{table}
\begin{table}[!htbp]
\centering\fontsize{9}{10}\selectfont\setlength{\tabcolsep}{3pt}
\begin{tabular*}{\columnwidth}{@{\extracolsep{\fill}}llrrr@{}}
\toprule
Model & Label & Base & T & F \\
\midrule
Llama 3B & SOONER & 100.00 & 100.00 & 100.00 \\
Qwen 3B & SOONER & 44.10 & 58.33 & 44.44 \\
Qwen 3B & LATER & 55.90 & 41.67 & 55.56 \\
Qwen 7B & LATER & 99.65 & 95.14 & 100.00 \\
Qwen 7B & SOONER & 0.35 & 4.86 & 0.00 \\
Llama 8B & SOONER & 100.00 & 100.00 & 100.00 \\
\bottomrule
\end{tabular*}
\caption{Individual decision making: Delayed rewards: complete label distributions (\%), $N=288$.}
\label{tab:full-time}
\end{table}
\begin{table}[!htbp]
\centering\fontsize{9}{10}\selectfont\setlength{\tabcolsep}{3pt}
\begin{tabular*}{\columnwidth}{@{\extracolsep{\fill}}llrrr@{}}
\toprule
Model & Label & Base & T & F \\
\midrule
Llama 3B & COUNTER & 100.00 & 100.00 & 99.89 \\
Llama 3B & EXIT & 0.00 & 0.00 & 0.11 \\
Qwen 3B & EXIT & 8.89 & 0.89 & 28.00 \\
Qwen 3B & COUNTER & 91.11 & 99.11 & 72.00 \\
Qwen 7B & COUNTER & 100.00 & 100.00 & 100.00 \\
Llama 8B & COUNTER & 100.00 & 100.00 & 100.00 \\
\bottomrule
\end{tabular*}
\caption{Individual decision making: Bargaining controlled: complete label distributions (\%), $N=900$.}
\label{tab:full-bargaining_controlled_full}
\end{table}
\begin{table}[!htbp]
\centering\fontsize{9}{10}\selectfont\setlength{\tabcolsep}{3pt}
\begin{tabular*}{\columnwidth}{@{\extracolsep{\fill}}llrrr@{}}
\toprule
Model & Label & Base & T & F \\
\midrule
Llama 3B & COUNTER & 100.00 & 100.00 & 100.00 \\
Qwen 3B & EXIT & 85.19 & 27.35 & 99.72 \\
Qwen 3B & COUNTER & 14.81 & 72.65 & 0.28 \\
Qwen 7B & COUNTER & 100.00 & 100.00 & 100.00 \\
Llama 8B & COUNTER & 100.00 & 100.00 & 100.00 \\
\bottomrule
\end{tabular*}
\caption{Interactive decision making: Bargaining real history: complete label distributions (\%), $N=351$.}
\label{tab:full-bargaining_real_history_full}
\end{table}
\begin{table}[!htbp]
\centering\fontsize{9}{10}\selectfont\setlength{\tabcolsep}{3pt}
\begin{tabular*}{\columnwidth}{@{\extracolsep{\fill}}llrrr@{}}
\toprule
Model & Label & Base & T & F \\
\midrule
Llama 3B & SEARCH\_\allowbreak{}OTHER & 45.04 & 36.52 & 46.81 \\
Llama 3B & ACCEPT & 29.43 & 39.01 & 18.79 \\
Llama 3B & ASK\_\allowbreak{}DETAILS & 25.53 & 24.47 & 34.40 \\
Qwen 3B & SEARCH\_\allowbreak{}OTHER & 90.43 & 80.50 & 91.49 \\
Qwen 3B & ASK\_\allowbreak{}DETAILS & 8.16 & 16.67 & 7.09 \\
Qwen 3B & ACCEPT & 1.42 & 2.84 & 1.42 \\
Qwen 7B & ASK\_\allowbreak{}DETAILS & 35.46 & 29.08 & 38.65 \\
Qwen 7B & SEARCH\_\allowbreak{}OTHER & 37.23 & 42.55 & 35.11 \\
Qwen 7B & ACCEPT & 27.30 & 28.37 & 26.24 \\
Llama 8B & ASK\_\allowbreak{}DETAILS & 72.34 & 86.17 & 60.64 \\
Llama 8B & SEARCH\_\allowbreak{}OTHER & 25.18 & 10.64 & 39.01 \\
Llama 8B & ACCEPT & 2.48 & 3.19 & 0.35 \\
\bottomrule
\end{tabular*}
\caption{Frames: complete label distributions (\%), $N=282$.}
\label{tab:full-frames_feedback_full}
\end{table}
\begin{table}[!htbp]
\centering\fontsize{9}{10}\selectfont\setlength{\tabcolsep}{3pt}
\begin{tabular*}{\columnwidth}{@{\extracolsep{\fill}}llrrr@{}}
\toprule
Model & Label & Base & T & F \\
\midrule
Llama 3B & ACCEPT & 78.15 & 93.28 & 25.21 \\
Llama 3B & ASK\_\allowbreak{}MORE & 21.01 & 5.88 & 74.79 \\
Llama 3B & REJECT & 0.84 & 0.84 & 0.00 \\
Qwen 3B & ASK\_\allowbreak{}MORE & 86.55 & 81.51 & 91.60 \\
Qwen 3B & ACCEPT & 12.61 & 14.29 & 8.40 \\
Qwen 3B & REJECT & 0.84 & 4.20 & 0.00 \\
Qwen 7B & ASK\_\allowbreak{}MORE & 64.71 & 64.71 & 68.07 \\
Qwen 7B & ACCEPT & 25.21 & 23.53 & 25.21 \\
Qwen 7B & REJECT & 10.08 & 11.76 & 6.72 \\
Llama 8B & ACCEPT & 55.46 & 71.43 & 12.61 \\
Llama 8B & ASK\_\allowbreak{}MORE & 36.13 & 20.17 & 84.03 \\
Llama 8B & REJECT & 8.40 & 8.40 & 3.36 \\
\bottomrule
\end{tabular*}
\caption{INSPIRED: complete label distributions (\%), $N=119$.}
\label{tab:full-inspired_feedback_full}
\end{table}
\begin{table}[!htbp]
\centering\fontsize{9}{10}\selectfont\setlength{\tabcolsep}{3pt}
\begin{tabular*}{\columnwidth}{@{\extracolsep{\fill}}llrrr@{}}
\toprule
Model & Label & Base & T & F \\
\midrule
Llama 3B & SELL & 37.21 & 29.75 & 28.16 \\
Llama 3B & HOLD & 20.90 & 27.06 & 58.61 \\
Llama 3B & BUY & 41.89 & 43.18 & 13.23 \\
Qwen 3B & SELL & 99.00 & 94.43 & 99.80 \\
Qwen 3B & HOLD & 1.00 & 5.57 & 0.20 \\
Qwen 7B & SELL & 39.90 & 39.60 & 41.29 \\
Qwen 7B & BUY & 42.49 & 44.08 & 41.99 \\
Qwen 7B & HOLD & 17.61 & 16.32 & 16.72 \\
Llama 8B & BUY & 96.12 & 95.62 & 59.60 \\
Llama 8B & HOLD & 3.88 & 4.38 & 40.40 \\
\bottomrule
\end{tabular*}
\caption{Individual decision making: Stock controlled: complete label distributions (\%), $N=1,005$.}
\label{tab:full-stock_controlled_full}
\end{table}
\begin{table}[!htbp]
\centering\fontsize{9}{10}\selectfont\setlength{\tabcolsep}{3pt}
\begin{tabular*}{\columnwidth}{@{\extracolsep{\fill}}llrrr@{}}
\toprule
Model & Label & Base & T & F \\
\midrule
Llama 3B & COUNTER & 99.82 & 100.00 & 98.55 \\
Llama 3B & EXIT & 0.18 & 0.00 & 1.23 \\
Llama 3B & ACCEPT & 0.00 & 0.00 & 0.22 \\
Qwen 3B & COUNTER & 98.46 & 99.96 & 85.04 \\
Qwen 3B & EXIT & 1.54 & 0.04 & 14.96 \\
Qwen 7B & COUNTER & 99.74 & 99.82 & 99.74 \\
Qwen 7B & EXIT & 0.26 & 0.18 & 0.26 \\
Llama 8B & COUNTER & 99.96 & 100.00 & 99.96 \\
Llama 8B & EXIT & 0.04 & 0.00 & 0.04 \\
\bottomrule
\end{tabular*}
\caption{Interactive decision making: Craigslist interaction below asking: complete label distributions (\%), $N=2,272$.}
\label{tab:full-craigslist_interaction_below_asking}
\end{table}

\begin{table}[!htbp]
\begin{minipage}{\linewidth}
\normalsize
\subsection{Programmed Bargaining Outcomes}\label{app:program_outcomes}
The first action distribution above and final disengagement rates measure different parts of the same trajectories. Table~\ref{tab:program-final-exit} reports EXIT at any point within the program's maximum of three choices, retaining all 2,272 histories with initial offers below the asking price. For Qwen 3B, final EXIT rates are 48.24\% for F and 0.26\% for T, compared with 7.35\% at Base. The final outcome therefore captures divergence that is not fully represented by the first action alone.
\end{minipage}
\par\vspace{3pt}
\centering\small\setlength{\tabcolsep}{4pt}
\begin{tabular*}{\columnwidth}{@{\extracolsep{\fill}}lrrr@{}}
\toprule
Model & Base & T & F \\
\midrule
Llama 3B & 0.18 & 0.00 & 1.23 \\
Llama 8B & 0.04 & 0.00 & 0.04 \\
Qwen 3B & 7.35 & 0.26 & 48.24 \\
Qwen 7B & 0.26 & 0.18 & 0.26 \\
\bottomrule
\end{tabular*}
\caption{Interactive decision making: Final disengagement rates (\%) in programmed bargaining.}
\label{tab:program-final-exit}
\end{table}

\clearpage

%% file: sections/appendix_statistics.tex
\begin{table*}[!tbp]
\begin{minipage}[t]{\columnwidth}
\normalsize
\section{Statistical Analysis}\label{app:statistics}
\subsection{Design, Tests, and Uncertainty}\label{app:stat_methods}

We retrospectively analyzed the frozen per input outputs. All comparisons preserve input pairing. Behavioral and interpretation outcomes use paired permutation tests that swap conditions within whole source clusters. We enumerate all swaps with at most twenty nonzero cluster differences; otherwise we use 99,999 random swaps and the add one correction. The smallest Monte Carlo p value (0.00001) is a resolution limit, not an exact tail estimate. The tests assume condition exchangeability within the retained blocks under the null.

We report effect sizes and marginal 95\% bootstrap percentile intervals with paired cluster resampling from 19,999 resamples. Both conditions are resampled together. Macro F1 is recomputed from sufficient counts over fixed task classes. Intervals are marginal and are not adjusted for multiple comparisons; accuracy and label rate changes are percentage points, while macro F1 changes use 100 times the raw F1 difference.

Circa groups all replies to the same question within each context (3,375 groups), with an additional ten-context sensitivity analysis. CommitmentBank groups source passages: label rates use 1,009 inputs in 1,006 groups, while performance uses 858 adapted-majority-reference inputs in 856 groups. FOMC groups 2,421 statements into 27 years; dependencies across adjacent years are not estimated. Original risk choices group equivalent option descriptions, machine choices retain all games for each of 26 participants, and INSPIRED/Frames group the 119/282 source dialogues.

Controlled risk summaries retain both presentation frames and both option orders within each cell defined by probability and payoff gap: six cells in discovery and twenty in replication. These are fixed design grids; resampling does not establish generalization to arbitrary economic situations.
\end{minipage}\hfill
\begin{minipage}[t]{\columnwidth}
\normalsize
Controlled bargaining retains all offers and option orders within fifteen source scenarios. Offer response is the EXIT change from 40\% to 90\% of asking price, and differences between these responses are paired across model conditions. EXIT is not a transaction-acceptance threshold.

Controlled movie/travel feedback comparisons use only common records where Base, T and F all reach turn two. We preserve costs and option orders within each authored scenario version. Only two versions are available, so we report descriptive effects and discrete tests but withhold nominal bootstrap confidence intervals. These selected cohort estimates are not unconditional causal effects.

All capability tables test accuracy differences. Capability outcomes use exact two sided McNemar tests, paired bootstrap intervals, and separate one sided degradation tests. GSM8K scoring follows the rules described in Appendix~\ref{app:capability_scope}. BoolQ additionally has passage block sensitivity analyses.

The frozen ARC parser does not recognize 87 reference labels among 2,000 (86 numerical option identifiers and one E). We preserve the recorded scoring with the full denominator and disclose this limitation rather than silently remapping labels. Removing these pairs scored as incorrect in both conditions leaves exact McNemar p values unchanged but changes the accuracy denominator.
\end{minipage}
\end{table*}
\begin{table*}[!tbp]
\begin{minipage}[t]{\columnwidth}
\normalsize
Holm correction is applied at $\alpha=.05$ within the listed analysis families. The trained variant family has 24 tests, Prompt has 18, and combined guidance has 36. One-sided capability tests use separate corrections over the same respective families. Sensitivity checks have separate families and cannot replace a failed primary result. A result that is not significant does not establish equivalence or noninferiority. These analyses do not estimate variability across training seeds or effects across the population of language models.

Test choice follows the paired outcome and dependence structure \citep{dror-etal-2018-hitchhikers}. These are retrospective exploratory analyses of frozen outputs. No significance test is applied to the retention rate. The statistical families below concern the specified T/F comparisons and capability conditions.
\end{minipage}\hfill
\begin{minipage}[t]{\columnwidth}
\normalsize
\subsection{Complete Results and Correction Families}\label{app:stat_results}
Each named test group below is one Holm correction family; its size is stated explicitly. Delta is the guided minus reference difference in percentage points, or 100 times the macro F1 difference. N/G gives paired input count and the number of blocks. CI is the marginal 95\% paired bootstrap interval. Both raw and adjusted two sided p values are reported; capability tables additionally report separately corrected one sided degradation p values. Bold adjusted p values meet .05. NA intervals are withheld for the two version controlled feedback probes. Model abbreviations follow the other appendix tables; CB denotes CommitmentBank. In capability tables, A:B denotes condition A minus reference B; each test group specifies whether T/F denotes DPO or prompting. Combined conditions retain the trained pole before the prompt pole.
\end{minipage}
\par\vspace{8pt}
\begin{minipage}{\linewidth}
\normalsize
\subsection{Interpretation Tests}
After Holm correction, 34 of 36 label proportion tests and 56 of 72 performance tests are significant. The Circa T macro F1 improvements are supported in all four models; the tables also distinguish label changes from changes in task accuracy.

\subsubsection{Interpretation: Label Proportions}\label{app:stat-interpretation-label}
Holm correction covers 36 comparisons.
\end{minipage}
\par\vspace{3pt}

\centering\fontsize{9}{10}\selectfont\setlength{\tabcolsep}{2pt}\renewcommand{\arraystretch}{0.96}
\begin{tabular*}{\textwidth}{@{\extracolsep{\fill}}l>{\raggedright\arraybackslash}p{.16\textwidth}>{\raggedright\arraybackslash}p{.10\textwidth}rrlrr@{}}
\toprule
Model & Task / outcome & Contrast & N/G & $\Delta$ & 95\% CI & $p$ & $p_{\mathrm{Holm}}$ \\
\midrule
Llama 3B & Circa / UNCLEAR & T-BASE & 32602/3375 & -21.24 & [-21.84, -20.63] & 1.00e-05 & \textbf{3.50e-04} \\
Llama 3B & Circa / UNCLEAR & F-BASE & 32602/3375 & +48.50 & [+47.74, +49.24] & 1.00e-05 & \textbf{3.50e-04} \\
Llama 3B & Circa / UNCLEAR & F-T & 32602/3375 & +69.74 & [+69.01, +70.45] & 1.00e-05 & \textbf{3.50e-04} \\
Llama 8B & Circa / UNCLEAR & T-BASE & 32602/3375 & -17.13 & [-17.64, -16.62] & 1.00e-05 & \textbf{3.50e-04} \\
Llama 8B & Circa / UNCLEAR & F-BASE & 32602/3375 & +40.50 & [+39.80, +41.21] & 1.00e-05 & \textbf{3.50e-04} \\
Llama 8B & Circa / UNCLEAR & F-T & 32602/3375 & +57.63 & [+56.88, +58.37] & 1.00e-05 & \textbf{3.50e-04} \\
Qwen 3B & Circa / UNCLEAR & T-BASE & 32602/3375 & -1.44 & [-1.60, -1.28] & 1.00e-05 & \textbf{3.50e-04} \\
Qwen 3B & Circa / UNCLEAR & F-BASE & 32602/3375 & +3.85 & [+3.61, +4.10] & 1.00e-05 & \textbf{3.50e-04} \\
Qwen 3B & Circa / UNCLEAR & F-T & 32602/3375 & +5.29 & [+5.00, +5.58] & 1.00e-05 & \textbf{3.50e-04} \\
Qwen 7B & Circa / UNCLEAR & T-BASE & 32602/3375 & -0.83 & [-1.04, -0.61] & 1.00e-05 & \textbf{3.50e-04} \\
Qwen 7B & Circa / UNCLEAR & F-BASE & 32602/3375 & +2.71 & [+2.47, +2.96] & 1.00e-05 & \textbf{3.50e-04} \\
Qwen 7B & Circa / UNCLEAR & F-T & 32602/3375 & +3.54 & [+3.20, +3.87] & 1.00e-05 & \textbf{3.50e-04} \\
\bottomrule
\end{tabular*}
\caption{Interpretation: Label Proportions: paired tests, part 1 (Holm family: 36).}
\label{tab:stat-interpretation_label-0}
\end{table*}
\begin{table*}[!tbp]
\centering\fontsize{9}{10}\selectfont\setlength{\tabcolsep}{2pt}\renewcommand{\arraystretch}{0.96}
\begin{tabular*}{\textwidth}{@{\extracolsep{\fill}}l>{\raggedright\arraybackslash}p{.16\textwidth}>{\raggedright\arraybackslash}p{.10\textwidth}rrlrr@{}}
\toprule
Model & Task / outcome & Contrast & N/G & $\Delta$ & 95\% CI & $p$ & $p_{\mathrm{Holm}}$ \\
\midrule
Llama 3B & CB / FALSE & T-BASE & 1009/1006 & +8.62 & [+6.94, +10.41] & 1.00e-05 & \textbf{3.50e-04} \\
Llama 3B & CB / FALSE & F-BASE & 1009/1006 & -28.34 & [-31.09, -25.54] & 1.00e-05 & \textbf{3.50e-04} \\
Llama 3B & CB / FALSE & F-T & 1009/1006 & -36.97 & [-39.96, -33.99] & 1.00e-05 & \textbf{3.50e-04} \\
Llama 8B & CB / FALSE & T-BASE & 1009/1006 & +18.43 & [+16.09, +20.85] & 1.00e-05 & \textbf{3.50e-04} \\
Llama 8B & CB / FALSE & F-BASE & 1009/1006 & -11.20 & [-13.18, -9.24] & 1.00e-05 & \textbf{3.50e-04} \\
Llama 8B & CB / FALSE & F-T & 1009/1006 & -29.63 & [-32.44, -26.81] & 1.00e-05 & \textbf{3.50e-04} \\
Qwen 3B & CB / FALSE & T-BASE & 1009/1006 & +6.24 & [+4.76, +7.82] & 1.00e-05 & \textbf{3.50e-04} \\
Qwen 3B & CB / FALSE & F-BASE & 1009/1006 & -11.60 & [-13.58, -9.68] & 1.00e-05 & \textbf{3.50e-04} \\
Qwen 3B & CB / FALSE & F-T & 1009/1006 & -17.84 & [-20.24, -15.56] & 1.00e-05 & \textbf{3.50e-04} \\
Qwen 7B & CB / FALSE & T-BASE & 1009/1006 & +1.19 & [+0.50, +1.98] & 0.0018 & \textbf{0.0110} \\
Qwen 7B & CB / FALSE & F-BASE & 1009/1006 & -5.45 & [-6.93, -4.06] & 1.00e-05 & \textbf{3.50e-04} \\
Qwen 7B & CB / FALSE & F-T & 1009/1006 & -6.64 & [-8.22, -5.15] & 1.00e-05 & \textbf{3.50e-04} \\
Llama 3B & fomc / HAWKISH & T-BASE & 2421/27 & +0.50 & [+0.27, +0.72] & 0.0020 & \textbf{0.0110} \\
Llama 3B & fomc / HAWKISH & F-BASE & 2421/27 & -0.33 & [-0.61, -0.09] & 0.0312 & 0.0625 \\
Llama 3B & fomc / HAWKISH & F-T & 2421/27 & -0.83 & [-1.27, -0.43] & 4.88e-04 & \textbf{0.0034} \\
Llama 8B & fomc / HAWKISH & T-BASE & 2421/27 & +3.68 & [+2.97, +4.42] & 1.00e-05 & \textbf{3.50e-04} \\
Llama 8B & fomc / HAWKISH & F-BASE & 2421/27 & -1.90 & [-2.74, -1.21] & 2.00e-05 & \textbf{3.50e-04} \\
Llama 8B & fomc / HAWKISH & F-T & 2421/27 & -5.58 & [-6.61, -4.70] & 1.00e-05 & \textbf{3.50e-04} \\
Qwen 3B & fomc / HAWKISH & T-BASE & 2421/27 & +1.28 & [+0.81, +1.82] & 7.63e-06 & \textbf{2.75e-04} \\
Qwen 3B & fomc / HAWKISH & F-BASE & 2421/27 & -1.07 & [-1.52, -0.68] & 3.05e-05 & \textbf{3.50e-04} \\
Qwen 3B & fomc / HAWKISH & F-T & 2421/27 & -2.35 & [-3.03, -1.74] & 1.00e-05 & \textbf{3.50e-04} \\
Qwen 7B & fomc / HAWKISH & T-BASE & 2421/27 & +0.12 & [-0.17, +0.40] & 0.5938 & 0.5938 \\
Qwen 7B & fomc / HAWKISH & F-BASE & 2421/27 & -0.45 & [-0.77, -0.21] & 0.0039 & \textbf{0.0156} \\
Qwen 7B & fomc / HAWKISH & F-T & 2421/27 & -0.58 & [-0.93, -0.24] & 0.0073 & \textbf{0.0220} \\
\bottomrule
\end{tabular*}
\caption{Interpretation: Label Proportions: paired tests, part 2 (Holm family: 36).}
\label{tab:stat-interpretation_label-18}
\end{table*}
\begin{table*}[!tbp]
\begin{minipage}{\linewidth}
\normalsize
\subsubsection{Interpretation: Performance}\label{app:stat-interpretation-performance}
Holm correction covers 72 comparisons.
\end{minipage}
\par\vspace{3pt}

\centering\fontsize{9}{10}\selectfont\setlength{\tabcolsep}{2pt}\renewcommand{\arraystretch}{0.96}
\begin{tabular*}{\textwidth}{@{\extracolsep{\fill}}l>{\raggedright\arraybackslash}p{.16\textwidth}>{\raggedright\arraybackslash}p{.10\textwidth}rrlrr@{}}
\toprule
Model & Task / outcome & Contrast & N/G & $\Delta$ & 95\% CI & $p$ & $p_{\mathrm{Holm}}$ \\
\midrule
Llama 3B & Circa / accuracy & T-BASE & 32602/3375 & +11.98 & [+11.36, +12.62] & 1.00e-05 & \textbf{7.20e-04} \\
Llama 3B & Circa / macro F1 & T-BASE & 32602/3375 & +7.03 & [+6.26, +7.80] & 1.00e-05 & \textbf{7.20e-04} \\
Llama 3B & Circa / accuracy & F-BASE & 32602/3375 & -34.94 & [-35.63, -34.24] & 1.00e-05 & \textbf{7.20e-04} \\
Llama 3B & Circa / macro F1 & F-BASE & 32602/3375 & -22.84 & [-23.44, -22.24] & 1.00e-05 & \textbf{7.20e-04} \\
Llama 3B & Circa / accuracy & F-T & 32602/3375 & -46.93 & [-47.70, -46.15] & 1.00e-05 & \textbf{7.20e-04} \\
Llama 3B & Circa / macro F1 & F-T & 32602/3375 & -29.87 & [-30.78, -28.96] & 1.00e-05 & \textbf{7.20e-04} \\
Llama 8B & Circa / accuracy & T-BASE & 32602/3375 & +14.23 & [+13.76, +14.69] & 1.00e-05 & \textbf{7.20e-04} \\
Llama 8B & Circa / macro F1 & T-BASE & 32602/3375 & +11.46 & [+11.01, +11.90] & 1.00e-05 & \textbf{7.20e-04} \\
Llama 8B & Circa / accuracy & F-BASE & 32602/3375 & -14.41 & [-14.93, -13.88] & 1.00e-05 & \textbf{7.20e-04} \\
Llama 8B & Circa / macro F1 & F-BASE & 32602/3375 & -10.83 & [-11.31, -10.35] & 1.00e-05 & \textbf{7.20e-04} \\
Llama 8B & Circa / accuracy & F-T & 32602/3375 & -28.63 & [-29.33, -27.94] & 1.00e-05 & \textbf{7.20e-04} \\
Llama 8B & Circa / macro F1 & F-T & 32602/3375 & -22.29 & [-22.89, -21.69] & 1.00e-05 & \textbf{7.20e-04} \\
Qwen 3B & Circa / accuracy & T-BASE & 32602/3375 & +1.07 & [+0.93, +1.21] & 1.00e-05 & \textbf{7.20e-04} \\
Qwen 3B & Circa / macro F1 & T-BASE & 32602/3375 & +1.09 & [+0.84, +1.35] & 1.00e-05 & \textbf{7.20e-04} \\
Qwen 3B & Circa / accuracy & F-BASE & 32602/3375 & -3.48 & [-3.72, -3.26] & 1.00e-05 & \textbf{7.20e-04} \\
Qwen 3B & Circa / macro F1 & F-BASE & 32602/3375 & -5.19 & [-5.60, -4.79] & 1.00e-05 & \textbf{7.20e-04} \\
Qwen 3B & Circa / accuracy & F-T & 32602/3375 & -4.55 & [-4.82, -4.29] & 1.00e-05 & \textbf{7.20e-04} \\
Qwen 3B & Circa / macro F1 & F-T & 32602/3375 & -6.28 & [-6.73, -5.83] & 1.00e-05 & \textbf{7.20e-04} \\
\bottomrule
\end{tabular*}
\caption{Interpretation: Performance: paired tests, part 1 (Holm family: 72).}
\label{tab:stat-interpretation_performance-0}
\end{table*}
\begin{table*}[!tbp]
\centering\fontsize{9}{10}\selectfont\setlength{\tabcolsep}{2pt}\renewcommand{\arraystretch}{0.96}
\begin{tabular*}{\textwidth}{@{\extracolsep{\fill}}l>{\raggedright\arraybackslash}p{.16\textwidth}>{\raggedright\arraybackslash}p{.10\textwidth}rrlrr@{}}
\toprule
Model & Task / outcome & Contrast & N/G & $\Delta$ & 95\% CI & $p$ & $p_{\mathrm{Holm}}$ \\
\midrule
Qwen 7B & Circa / accuracy & T-BASE & 32602/3375 & +2.03 & [+1.86, +2.21] & 1.00e-05 & \textbf{7.20e-04} \\
Qwen 7B & Circa / macro F1 & T-BASE & 32602/3375 & +2.27 & [+2.10, +2.45] & 1.00e-05 & \textbf{7.20e-04} \\
Qwen 7B & Circa / accuracy & F-BASE & 32602/3375 & -3.32 & [-3.54, -3.11] & 1.00e-05 & \textbf{7.20e-04} \\
Qwen 7B & Circa / macro F1 & F-BASE & 32602/3375 & -3.17 & [-3.37, -2.96] & 1.00e-05 & \textbf{7.20e-04} \\
Qwen 7B & Circa / accuracy & F-T & 32602/3375 & -5.35 & [-5.64, -5.07] & 1.00e-05 & \textbf{7.20e-04} \\
Qwen 7B & Circa / macro F1 & F-T & 32602/3375 & -5.44 & [-5.71, -5.18] & 1.00e-05 & \textbf{7.20e-04} \\
Llama 3B & CB / accuracy & T-BASE & 858/856 & -2.45 & [-4.43, -0.47] & 0.0194 & 0.2327 \\
Llama 3B & CB / macro F1 & T-BASE & 858/856 & -4.76 & [-6.53, -3.06] & 1.00e-05 & \textbf{7.20e-04} \\
Llama 3B & CB / accuracy & F-BASE & 858/856 & +6.99 & [+3.61, +10.37] & 1.00e-04 & \textbf{0.0022} \\
Llama 3B & CB / macro F1 & F-BASE & 858/856 & +9.03 & [+6.18, +11.95] & 1.00e-05 & \textbf{7.20e-04} \\
Llama 3B & CB / accuracy & F-T & 858/856 & +9.44 & [+5.59, +13.30] & 1.00e-05 & \textbf{7.20e-04} \\
Llama 3B & CB / macro F1 & F-T & 858/856 & +13.79 & [+10.72, +16.92] & 1.00e-05 & \textbf{7.20e-04} \\
Llama 8B & CB / accuracy & T-BASE & 858/856 & +22.26 & [+19.16, +25.32] & 1.00e-05 & \textbf{7.20e-04} \\
Llama 8B & CB / macro F1 & T-BASE & 858/856 & +19.34 & [+16.50, +22.18] & 1.00e-05 & \textbf{7.20e-04} \\
Llama 8B & CB / accuracy & F-BASE & 858/856 & -16.43 & [-19.07, -13.82] & 1.00e-05 & \textbf{7.20e-04} \\
Llama 8B & CB / macro F1 & F-BASE & 858/856 & -20.52 & [-23.19, -17.76] & 1.00e-05 & \textbf{7.20e-04} \\
Llama 8B & CB / accuracy & F-T & 858/856 & -38.69 & [-42.31, -35.08] & 1.00e-05 & \textbf{7.20e-04} \\
Llama 8B & CB / macro F1 & F-T & 858/856 & -39.86 & [-42.93, -36.77] & 1.00e-05 & \textbf{7.20e-04} \\
\bottomrule
\end{tabular*}
\caption{Interpretation: Performance: paired tests, part 2 (Holm family: 72).}
\label{tab:stat-interpretation_performance-18}
\end{table*}
\begin{table*}[!tbp]
\centering\fontsize{9}{10}\selectfont\setlength{\tabcolsep}{2pt}\renewcommand{\arraystretch}{0.96}
\begin{tabular*}{\textwidth}{@{\extracolsep{\fill}}l>{\raggedright\arraybackslash}p{.16\textwidth}>{\raggedright\arraybackslash}p{.10\textwidth}rrlrr@{}}
\toprule
Model & Task / outcome & Contrast & N/G & $\Delta$ & 95\% CI & $p$ & $p_{\mathrm{Holm}}$ \\
\midrule
Qwen 3B & CB / accuracy & T-BASE & 858/856 & +4.78 & [+3.38, +6.29] & 1.00e-05 & \textbf{7.20e-04} \\
Qwen 3B & CB / macro F1 & T-BASE & 858/856 & +2.95 & [+1.95, +4.01] & 1.00e-05 & \textbf{7.20e-04} \\
Qwen 3B & CB / accuracy & F-BASE & 858/856 & -10.02 & [-12.22, -7.92] & 1.00e-05 & \textbf{7.20e-04} \\
Qwen 3B & CB / macro F1 & F-BASE & 858/856 & -8.20 & [-10.06, -6.39] & 1.00e-05 & \textbf{7.20e-04} \\
Qwen 3B & CB / accuracy & F-T & 858/856 & -14.80 & [-17.37, -12.33] & 1.00e-05 & \textbf{7.20e-04} \\
Qwen 3B & CB / macro F1 & F-T & 858/856 & -11.16 & [-13.22, -9.16] & 1.00e-05 & \textbf{7.20e-04} \\
Qwen 7B & CB / accuracy & T-BASE & 858/856 & +0.93 & [-0.12, +1.98] & 0.1142 & 0.8690 \\
Qwen 7B & CB / macro F1 & T-BASE & 858/856 & +0.71 & [-0.20, +1.61] & 0.1086 & 0.8690 \\
Qwen 7B & CB / accuracy & F-BASE & 858/856 & -5.59 & [-7.34, -3.95] & 1.00e-05 & \textbf{7.20e-04} \\
Qwen 7B & CB / macro F1 & F-BASE & 858/856 & -4.70 & [-6.22, -3.23] & 1.00e-05 & \textbf{7.20e-04} \\
Qwen 7B & CB / accuracy & F-T & 858/856 & -6.53 & [-8.51, -4.55] & 1.00e-05 & \textbf{7.20e-04} \\
Qwen 7B & CB / macro F1 & F-T & 858/856 & -5.41 & [-7.11, -3.74] & 1.00e-05 & \textbf{7.20e-04} \\
Llama 3B & fomc / accuracy & T-BASE & 2421/27 & -0.74 & [-2.15, +0.60] & 0.3300 & 0.9900 \\
Llama 3B & fomc / macro F1 & T-BASE & 2421/27 & +1.05 & [-0.19, +2.21] & 0.1123 & 0.8690 \\
Llama 3B & fomc / accuracy & F-BASE & 2421/27 & +0.95 & [-0.40, +2.20] & 0.1956 & 0.8690 \\
Llama 3B & fomc / macro F1 & F-BASE & 2421/27 & -1.21 & [-2.48, -0.05] & 0.0755 & 0.7549 \\
Llama 3B & fomc / accuracy & F-T & 2421/27 & +1.69 & [-0.09, +3.43] & 0.0871 & 0.7840 \\
Llama 3B & fomc / macro F1 & F-T & 2421/27 & -2.26 & [-3.98, -0.56] & 0.0209 & 0.2327 \\
\bottomrule
\end{tabular*}
\caption{Interpretation: Performance: paired tests, part 3 (Holm family: 72).}
\label{tab:stat-interpretation_performance-36}
\end{table*}
\begin{table*}[!tbp]
\centering\fontsize{9}{10}\selectfont\setlength{\tabcolsep}{2pt}\renewcommand{\arraystretch}{0.96}
\begin{tabular*}{\textwidth}{@{\extracolsep{\fill}}l>{\raggedright\arraybackslash}p{.16\textwidth}>{\raggedright\arraybackslash}p{.10\textwidth}rrlrr@{}}
\toprule
Model & Task / outcome & Contrast & N/G & $\Delta$ & 95\% CI & $p$ & $p_{\mathrm{Holm}}$ \\
\midrule
Llama 8B & fomc / accuracy & T-BASE & 2421/27 & +1.98 & [+0.82, +3.11] & 0.0054 & 0.0763 \\
Llama 8B & fomc / macro F1 & T-BASE & 2421/27 & +5.27 & [+3.78, +6.69] & 1.00e-05 & \textbf{7.20e-04} \\
Llama 8B & fomc / accuracy & F-BASE & 2421/27 & -2.85 & [-3.85, -1.95] & 1.00e-05 & \textbf{7.20e-04} \\
Llama 8B & fomc / macro F1 & F-BASE & 2421/27 & -7.28 & [-8.95, -5.69] & 1.00e-05 & \textbf{7.20e-04} \\
Llama 8B & fomc / accuracy & F-T & 2421/27 & -4.83 & [-6.26, -3.40] & 1.00e-05 & \textbf{7.20e-04} \\
Llama 8B & fomc / macro F1 & F-T & 2421/27 & -12.54 & [-14.40, -10.66] & 1.00e-05 & \textbf{7.20e-04} \\
Qwen 3B & fomc / accuracy & T-BASE & 2421/27 & +0.70 & [+0.22, +1.17] & 0.0165 & 0.2140 \\
Qwen 3B & fomc / macro F1 & T-BASE & 2421/27 & +1.42 & [+0.62, +2.21] & 0.0033 & 0.0523 \\
Qwen 3B & fomc / accuracy & F-BASE & 2421/27 & -0.50 & [-1.02, +0.04] & 0.1278 & 0.8690 \\
Qwen 3B & fomc / macro F1 & F-BASE & 2421/27 & -1.54 & [-2.33, -0.77] & 0.0012 & \textbf{0.0248} \\
Qwen 3B & fomc / accuracy & F-T & 2421/27 & -1.20 & [-1.81, -0.56] & 0.0039 & 0.0583 \\
Qwen 3B & fomc / macro F1 & F-T & 2421/27 & -2.96 & [-4.05, -1.84] & 7.00e-05 & \textbf{0.0016} \\
Qwen 7B & fomc / accuracy & T-BASE & 2421/27 & +0.54 & [-0.41, +1.48] & 0.3483 & 0.9900 \\
Qwen 7B & fomc / macro F1 & T-BASE & 2421/27 & +0.35 & [-0.59, +1.32] & 0.5324 & 0.9900 \\
Qwen 7B & fomc / accuracy & F-BASE & 2421/27 & -1.45 & [-2.12, -0.71] & 0.0014 & \textbf{0.0264} \\
Qwen 7B & fomc / macro F1 & F-BASE & 2421/27 & -1.44 & [-2.10, -0.77] & 3.60e-04 & \textbf{0.0076} \\
Qwen 7B & fomc / accuracy & F-T & 2421/27 & -1.98 & [-2.96, -0.97] & 0.0023 & \textbf{0.0398} \\
Qwen 7B & fomc / macro F1 & F-T & 2421/27 & -1.79 & [-2.68, -0.89] & 0.0018 & \textbf{0.0319} \\
\bottomrule
\end{tabular*}
\caption{Interpretation: Performance: paired tests, part 4 (Holm family: 72).}
\label{tab:stat-interpretation_performance-54}
\end{table*}
\begin{table*}[!tbp]
\begin{minipage}{\linewidth}
\normalsize
\subsubsection{Circa Context Sensitivity: Labels}\label{app:stat-sensitivity-circa-topic-label}
Holm correction covers 12 comparisons.

The context-level sensitivity analysis confirms significant differences in both prediction distributions and task performance after Holm correction, supporting the robustness of the observed personality effects on Circa.

\end{minipage}
\par\vspace{3pt}

\centering\fontsize{9}{10}\selectfont\setlength{\tabcolsep}{2pt}\renewcommand{\arraystretch}{0.96}
\begin{tabular*}{\textwidth}{@{\extracolsep{\fill}}l>{\raggedright\arraybackslash}p{.16\textwidth}>{\raggedright\arraybackslash}p{.10\textwidth}rrlrr@{}}
\toprule
Model & Task / outcome & Contrast & N/G & $\Delta$ & 95\% CI & $p$ & $p_{\mathrm{Holm}}$ \\
\midrule
Llama 3B & Circa / UNCLEAR & T-BASE & 32602/10 & -21.24 & [-25.12, -18.10] & 0.0020 & \textbf{0.0234} \\
Llama 3B & Circa / UNCLEAR & F-BASE & 32602/10 & +48.50 & [+44.10, +53.39] & 0.0020 & \textbf{0.0234} \\
Llama 3B & Circa / UNCLEAR & F-T & 32602/10 & +69.74 & [+64.66, +74.73] & 0.0020 & \textbf{0.0234} \\
Llama 8B & Circa / UNCLEAR & T-BASE & 32602/10 & -17.13 & [-18.62, -15.37] & 0.0020 & \textbf{0.0234} \\
Llama 8B & Circa / UNCLEAR & F-BASE & 32602/10 & +40.50 & [+35.99, +44.72] & 0.0020 & \textbf{0.0234} \\
Llama 8B & Circa / UNCLEAR & F-T & 32602/10 & +57.63 & [+52.91, +62.16] & 0.0020 & \textbf{0.0234} \\
Qwen 3B & Circa / UNCLEAR & T-BASE & 32602/10 & -1.44 & [-1.78, -1.11] & 0.0020 & \textbf{0.0234} \\
Qwen 3B & Circa / UNCLEAR & F-BASE & 32602/10 & +3.85 & [+2.88, +4.93] & 0.0020 & \textbf{0.0234} \\
Qwen 3B & Circa / UNCLEAR & F-T & 32602/10 & +5.29 & [+4.31, +6.30] & 0.0020 & \textbf{0.0234} \\
Qwen 7B & Circa / UNCLEAR & T-BASE & 32602/10 & -0.83 & [-1.30, -0.35] & 0.0156 & \textbf{0.0234} \\
Qwen 7B & Circa / UNCLEAR & F-BASE & 32602/10 & +2.71 & [+2.13, +3.33] & 0.0020 & \textbf{0.0234} \\
Qwen 7B & Circa / UNCLEAR & F-T & 32602/10 & +3.54 & [+2.52, +4.57] & 0.0020 & \textbf{0.0234} \\
\bottomrule
\end{tabular*}
\caption{Circa Context Sensitivity: Labels: paired tests, part 1 (Holm family: 12).}
\label{tab:stat-sensitivity_circa_topic_label-0}
\end{table*}
\begin{table*}[!tbp]
\begin{minipage}{\linewidth}
\normalsize
\subsubsection{Circa Context Sensitivity: Performance}\label{app:stat-sensitivity-circa-topic-performance}
Holm correction covers 24 comparisons.
\end{minipage}
\par\vspace{3pt}

\centering\fontsize{9}{10}\selectfont\setlength{\tabcolsep}{2pt}\renewcommand{\arraystretch}{0.96}
\begin{tabular*}{\textwidth}{@{\extracolsep{\fill}}l>{\raggedright\arraybackslash}p{.16\textwidth}>{\raggedright\arraybackslash}p{.10\textwidth}rrlrr@{}}
\toprule
Model & Task / outcome & Contrast & N/G & $\Delta$ & 95\% CI & $p$ & $p_{\mathrm{Holm}}$ \\
\midrule
Llama 3B & Circa / accuracy & T-BASE & 32602/10 & +11.98 & [+9.70, +14.99] & 0.0020 & \textbf{0.0469} \\
Llama 3B & Circa / macro F1 & T-BASE & 32602/10 & +7.03 & [+5.26, +9.11] & 0.0020 & \textbf{0.0469} \\
Llama 3B & Circa / accuracy & F-BASE & 32602/10 & -34.94 & [-39.17, -31.20] & 0.0020 & \textbf{0.0469} \\
Llama 3B & Circa / macro F1 & F-BASE & 32602/10 & -22.84 & [-25.59, -19.97] & 0.0020 & \textbf{0.0469} \\
Llama 3B & Circa / accuracy & F-T & 32602/10 & -46.93 & [-50.76, -43.00] & 0.0020 & \textbf{0.0469} \\
Llama 3B & Circa / macro F1 & F-T & 32602/10 & -29.87 & [-31.97, -27.64] & 0.0020 & \textbf{0.0469} \\
Llama 8B & Circa / accuracy & T-BASE & 32602/10 & +14.23 & [+12.51, +16.02] & 0.0020 & \textbf{0.0469} \\
Llama 8B & Circa / macro F1 & T-BASE & 32602/10 & +11.46 & [+10.37, +12.48] & 0.0020 & \textbf{0.0469} \\
Llama 8B & Circa / accuracy & F-BASE & 32602/10 & -14.41 & [-16.70, -12.25] & 0.0020 & \textbf{0.0469} \\
Llama 8B & Circa / macro F1 & F-BASE & 32602/10 & -10.83 & [-12.47, -9.14] & 0.0020 & \textbf{0.0469} \\
Llama 8B & Circa / accuracy & F-T & 32602/10 & -28.63 & [-31.95, -25.11] & 0.0020 & \textbf{0.0469} \\
Llama 8B & Circa / macro F1 & F-T & 32602/10 & -22.29 & [-24.26, -20.28] & 0.0020 & \textbf{0.0469} \\
Qwen 3B & Circa / accuracy & T-BASE & 32602/10 & +1.07 & [+0.71, +1.45] & 0.0020 & \textbf{0.0469} \\
Qwen 3B & Circa / macro F1 & T-BASE & 32602/10 & +1.09 & [+0.60, +1.71] & 0.0059 & \textbf{0.0469} \\
Qwen 3B & Circa / accuracy & F-BASE & 32602/10 & -3.48 & [-4.45, -2.56] & 0.0020 & \textbf{0.0469} \\
Qwen 3B & Circa / macro F1 & F-BASE & 32602/10 & -5.19 & [-6.10, -4.10] & 0.0020 & \textbf{0.0469} \\
Qwen 3B & Circa / accuracy & F-T & 32602/10 & -4.55 & [-5.47, -3.61] & 0.0020 & \textbf{0.0469} \\
Qwen 3B & Circa / macro F1 & F-T & 32602/10 & -6.28 & [-7.13, -5.22] & 0.0020 & \textbf{0.0469} \\
\bottomrule
\end{tabular*}
\caption{Circa Context Sensitivity: Performance: paired tests, part 1 (Holm family: 24).}
\label{tab:stat-sensitivity_circa_topic_performance-0}
\end{table*}
\begin{table*}[!tbp]
\centering\fontsize{9}{10}\selectfont\setlength{\tabcolsep}{2pt}\renewcommand{\arraystretch}{0.96}
\begin{tabular*}{\textwidth}{@{\extracolsep{\fill}}l>{\raggedright\arraybackslash}p{.16\textwidth}>{\raggedright\arraybackslash}p{.10\textwidth}rrlrr@{}}
\toprule
Model & Task / outcome & Contrast & N/G & $\Delta$ & 95\% CI & $p$ & $p_{\mathrm{Holm}}$ \\
\midrule
Qwen 7B & Circa / accuracy & T-BASE & 32602/10 & +2.03 & [+1.67, +2.46] & 0.0020 & \textbf{0.0469} \\
Qwen 7B & Circa / macro F1 & T-BASE & 32602/10 & +2.27 & [+1.96, +2.62] & 0.0020 & \textbf{0.0469} \\
Qwen 7B & Circa / accuracy & F-BASE & 32602/10 & -3.32 & [-3.79, -2.84] & 0.0020 & \textbf{0.0469} \\
Qwen 7B & Circa / macro F1 & F-BASE & 32602/10 & -3.17 & [-3.49, -2.83] & 0.0020 & \textbf{0.0469} \\
Qwen 7B & Circa / accuracy & F-T & 32602/10 & -5.35 & [-6.23, -4.54] & 0.0020 & \textbf{0.0469} \\
Qwen 7B & Circa / macro F1 & F-T & 32602/10 & -5.44 & [-6.08, -4.81] & 0.0020 & \textbf{0.0469} \\
\bottomrule
\end{tabular*}
\caption{Circa Context Sensitivity: Performance: paired tests, part 2 (Holm family: 24).}
\label{tab:stat-sensitivity_circa_topic_performance-18}
\end{table*}
\begin{table*}[!tbp]
\begin{minipage}{\linewidth}
\normalsize
\subsection{Decision and Interaction Tests}
T/F machine choice differences pass correction in all four models. The larger risk summary replication supports five of twelve comparisons, while the small discovery probe supports none. The complete results separate these findings from decision and interaction comparisons that do not pass correction.

\subsubsection{Main Decision and Interaction Outcomes}\label{app:stat-main-behavior}
Holm correction covers 48 comparisons.
\end{minipage}
\par\vspace{3pt}

\centering\fontsize{9}{10}\selectfont\setlength{\tabcolsep}{2pt}\renewcommand{\arraystretch}{0.96}
\begin{tabular*}{\textwidth}{@{\extracolsep{\fill}}l>{\raggedright\arraybackslash}p{.16\textwidth}>{\raggedright\arraybackslash}p{.10\textwidth}rrlrr@{}}
\toprule
Model & Task / outcome & Contrast & N/G & $\Delta$ & 95\% CI & $p$ & $p_{\mathrm{Holm}}$ \\
\midrule
Llama 3B & risk / high variance & T-BASE & 1764/1764 & -3.80 & [-5.90, -1.70] & 4.60e-04 & \textbf{0.0092} \\
Llama 3B & risk / high variance & F-BASE & 1764/1764 & +15.25 & [+12.64, +17.91] & 1.00e-05 & \textbf{4.60e-04} \\
Llama 3B & risk / high variance & F-T & 1764/1764 & +19.05 & [+15.82, +22.34] & 1.00e-05 & \textbf{4.60e-04} \\
Llama 3B & Machine / fewer observations & T-BASE & 4160/26 & +9.86 & [+8.85, +11.01] & 1.00e-05 & \textbf{4.60e-04} \\
Llama 3B & Machine / fewer observations & F-BASE & 4160/26 & -43.03 & [-44.52, -41.44] & 1.00e-05 & \textbf{4.60e-04} \\
Llama 3B & Machine / fewer observations & F-T & 4160/26 & -52.88 & [-54.81, -50.91] & 1.00e-05 & \textbf{4.60e-04} \\
Llama 3B & INSPIRED / ASK & T-BASE & 119/119 & -15.13 & [-21.85, -9.24] & 7.63e-06 & \textbf{3.59e-04} \\
Llama 3B & INSPIRED / ASK & F-BASE & 119/119 & +53.78 & [+44.54, +63.03] & 1.00e-05 & \textbf{4.60e-04} \\
Llama 3B & INSPIRED / ASK & F-T & 119/119 & +68.91 & [+60.50, +77.31] & 1.00e-05 & \textbf{4.60e-04} \\
Llama 3B & Frames / ACCEPT & T-BASE & 282/282 & +9.57 & [+6.38, +13.12] & 1.00e-05 & \textbf{4.60e-04} \\
Llama 3B & Frames / ACCEPT & F-BASE & 282/282 & -10.64 & [-14.54, -7.09] & 1.00e-05 & \textbf{4.60e-04} \\
Llama 3B & Frames / ACCEPT & F-T & 282/282 & -20.21 & [-24.82, -15.60] & 1.00e-05 & \textbf{4.60e-04} \\
Llama 8B & risk / high variance & T-BASE & 1764/1764 & -1.30 & [-2.55, -0.06] & 0.0527 & 0.7375 \\
Llama 8B & risk / high variance & F-BASE & 1764/1764 & +21.26 & [+18.99, +23.58] & 1.00e-05 & \textbf{4.60e-04} \\
Llama 8B & risk / high variance & F-T & 1764/1764 & +22.56 & [+19.95, +25.17] & 1.00e-05 & \textbf{4.60e-04} \\
Llama 8B & Machine / fewer observations & T-BASE & 4160/26 & +16.92 & [+15.43, +18.51] & 1.00e-05 & \textbf{4.60e-04} \\
Llama 8B & Machine / fewer observations & F-BASE & 4160/26 & -17.55 & [-18.89, -16.15] & 1.00e-05 & \textbf{4.60e-04} \\
Llama 8B & Machine / fewer observations & F-T & 4160/26 & -34.47 & [-35.87, -33.12] & 1.00e-05 & \textbf{4.60e-04} \\
\bottomrule
\end{tabular*}
\caption{Main Decision and Interaction Outcomes: paired tests, part 1 (Holm family: 48).}
\label{tab:stat-main_behavior-0}
\end{table*}
\begin{table*}[!tbp]
\centering\fontsize{9}{10}\selectfont\setlength{\tabcolsep}{2pt}\renewcommand{\arraystretch}{0.96}
\begin{tabular*}{\textwidth}{@{\extracolsep{\fill}}l>{\raggedright\arraybackslash}p{.16\textwidth}>{\raggedright\arraybackslash}p{.10\textwidth}rrlrr@{}}
\toprule
Model & Task / outcome & Contrast & N/G & $\Delta$ & 95\% CI & $p$ & $p_{\mathrm{Holm}}$ \\
\midrule
Llama 8B & INSPIRED / ASK & T-BASE & 119/119 & -15.97 & [-22.69, -10.08] & 3.81e-06 & \textbf{1.83e-04} \\
Llama 8B & INSPIRED / ASK & F-BASE & 119/119 & +47.90 & [+38.66, +57.14] & 1.00e-05 & \textbf{4.60e-04} \\
Llama 8B & INSPIRED / ASK & F-T & 119/119 & +63.87 & [+55.46, +72.27] & 1.00e-05 & \textbf{4.60e-04} \\
Llama 8B & Frames / ACCEPT & T-BASE & 282/282 & +0.71 & [+0.00, +1.77] & 0.5000 & 1.000 \\
Llama 8B & Frames / ACCEPT & F-BASE & 282/282 & -2.13 & [-3.90, -0.71] & 0.0312 & 0.5000 \\
Llama 8B & Frames / ACCEPT & F-T & 282/282 & -2.84 & [-4.96, -1.06] & 0.0078 & 0.1406 \\
Qwen 3B & risk / high variance & T-BASE & 1764/1764 & -3.40 & [-4.71, -2.15] & 1.00e-05 & \textbf{4.60e-04} \\
Qwen 3B & risk / high variance & F-BASE & 1764/1764 & +4.31 & [+3.06, +5.56] & 1.00e-05 & \textbf{4.60e-04} \\
Qwen 3B & risk / high variance & F-T & 1764/1764 & +7.71 & [+6.01, +9.47] & 1.00e-05 & \textbf{4.60e-04} \\
Qwen 3B & Machine / fewer observations & T-BASE & 4160/26 & +8.46 & [+7.40, +9.62] & 1.00e-05 & \textbf{4.60e-04} \\
Qwen 3B & Machine / fewer observations & F-BASE & 4160/26 & -9.04 & [-10.43, -7.64] & 1.00e-05 & \textbf{4.60e-04} \\
Qwen 3B & Machine / fewer observations & F-T & 4160/26 & -17.50 & [-18.70, -16.25] & 1.00e-05 & \textbf{4.60e-04} \\
Qwen 3B & INSPIRED / ASK & T-BASE & 119/119 & -5.04 & [-10.08, -0.84] & 0.0703 & 0.9141 \\
Qwen 3B & INSPIRED / ASK & F-BASE & 119/119 & +5.04 & [+1.68, +9.24] & 0.0312 & 0.5000 \\
Qwen 3B & INSPIRED / ASK & F-T & 119/119 & +10.08 & [+4.20, +15.97] & 0.0018 & \textbf{0.0348} \\
Qwen 3B & Frames / ACCEPT & T-BASE & 282/282 & +1.42 & [+0.35, +2.84] & 0.1250 & 1.000 \\
Qwen 3B & Frames / ACCEPT & F-BASE & 282/282 & +0.00 & [-1.06, +1.06] & 1.000 & 1.000 \\
Qwen 3B & Frames / ACCEPT & F-T & 282/282 & -1.42 & [-2.84, -0.35] & 0.1250 & 1.000 \\
\bottomrule
\end{tabular*}
\caption{Main Decision and Interaction Outcomes: paired tests, part 2 (Holm family: 48).}
\label{tab:stat-main_behavior-18}
\end{table*}
\begin{table*}[!tbp]
\centering\fontsize{9}{10}\selectfont\setlength{\tabcolsep}{2pt}\renewcommand{\arraystretch}{0.96}
\begin{tabular*}{\textwidth}{@{\extracolsep{\fill}}l>{\raggedright\arraybackslash}p{.16\textwidth}>{\raggedright\arraybackslash}p{.10\textwidth}rrlrr@{}}
\toprule
Model & Task / outcome & Contrast & N/G & $\Delta$ & 95\% CI & $p$ & $p_{\mathrm{Holm}}$ \\
\midrule
Qwen 7B & risk / high variance & T-BASE & 1764/1764 & +1.19 & [+0.34, +2.04] & 0.0091 & 0.1542 \\
Qwen 7B & risk / high variance & F-BASE & 1764/1764 & +0.28 & [-0.45, +1.02] & 0.5537 & 1.000 \\
Qwen 7B & risk / high variance & F-T & 1764/1764 & -0.91 & [-2.04, +0.23] & 0.1446 & 1.000 \\
Qwen 7B & Machine / fewer observations & T-BASE & 4160/26 & +3.41 & [+2.69, +4.13] & 1.00e-05 & \textbf{4.60e-04} \\
Qwen 7B & Machine / fewer observations & F-BASE & 4160/26 & -3.46 & [-4.04, -2.88] & 1.00e-05 & \textbf{4.60e-04} \\
Qwen 7B & Machine / fewer observations & F-T & 4160/26 & -6.88 & [-7.79, -5.87] & 1.00e-05 & \textbf{4.60e-04} \\
Qwen 7B & INSPIRED / ASK & T-BASE & 119/119 & +0.00 & [-3.36, +3.36] & 1.000 & 1.000 \\
Qwen 7B & INSPIRED / ASK & F-BASE & 119/119 & +3.36 & [+0.00, +7.56] & 0.2188 & 1.000 \\
Qwen 7B & INSPIRED / ASK & F-T & 119/119 & +3.36 & [-1.68, +8.40] & 0.3438 & 1.000 \\
Qwen 7B & Frames / ACCEPT & T-BASE & 282/282 & +1.06 & [-0.71, +2.84] & 0.4531 & 1.000 \\
Qwen 7B & Frames / ACCEPT & F-BASE & 282/282 & -1.06 & [-2.84, +0.71] & 0.4531 & 1.000 \\
Qwen 7B & Frames / ACCEPT & F-T & 282/282 & -2.13 & [-4.61, +0.35] & 0.1796 & 1.000 \\
\bottomrule
\end{tabular*}
\caption{Main Decision and Interaction Outcomes: paired tests, part 3 (Holm family: 48).}
\label{tab:stat-main_behavior-36}
\end{table*}
\begin{table*}[!tbp]
\begin{minipage}{\linewidth}
\normalsize
\subsubsection{Expected Payoff Probe: Discovery}\label{app:stat-risk-summary-discovery}
Holm correction covers 12 comparisons.
\end{minipage}
\par\vspace{3pt}

\centering\fontsize{9}{10}\selectfont\setlength{\tabcolsep}{2pt}\renewcommand{\arraystretch}{0.96}
\begin{tabular*}{\textwidth}{@{\extracolsep{\fill}}l>{\raggedright\arraybackslash}p{.16\textwidth}>{\raggedright\arraybackslash}p{.10\textwidth}rrlrr@{}}
\toprule
Model & Task / outcome & Contrast & N/G & $\Delta$ & 95\% CI & $p$ & $p_{\mathrm{Holm}}$ \\
\midrule
Llama 3B & Risk summary / switch rate & T-BASE & 24/6 & +25.00 & [-4.17, +58.33] & 0.3750 & 1.000 \\
Llama 3B & Risk summary / switch rate & F-BASE & 24/6 & -16.67 & [-25.00, -8.33] & 0.1250 & 1.000 \\
Llama 3B & Risk summary / switch rate & F-T & 24/6 & -41.67 & [-66.67, -20.83] & 0.0625 & 0.6250 \\
Llama 8B & Risk summary / switch rate & T-BASE & 24/6 & -8.33 & [-16.67, +0.00] & 0.5000 & 1.000 \\
Llama 8B & Risk summary / switch rate & F-BASE & 24/6 & -45.83 & [-50.00, -37.50] & 0.0312 & 0.3750 \\
Llama 8B & Risk summary / switch rate & F-T & 24/6 & -37.50 & [-45.83, -29.17] & 0.0312 & 0.3750 \\
Qwen 3B & Risk summary / switch rate & T-BASE & 24/6 & -4.17 & [-12.50, +0.00] & 1.000 & 1.000 \\
Qwen 3B & Risk summary / switch rate & F-BASE & 24/6 & -16.67 & [-25.00, -8.33] & 0.1250 & 1.000 \\
Qwen 3B & Risk summary / switch rate & F-T & 24/6 & -12.50 & [-20.83, -4.17] & 0.2500 & 1.000 \\
Qwen 7B & Risk summary / switch rate & T-BASE & 24/6 & +12.50 & [+4.17, +20.83] & 0.2500 & 1.000 \\
Qwen 7B & Risk summary / switch rate & F-BASE & 24/6 & +0.00 & [+0.00, +0.00] & 1.000 & 1.000 \\
Qwen 7B & Risk summary / switch rate & F-T & 24/6 & -12.50 & [-20.83, -4.17] & 0.2500 & 1.000 \\
\bottomrule
\end{tabular*}
\caption{Expected Payoff Probe: Discovery: paired tests, part 1 (Holm family: 12).}
\label{tab:stat-risk_summary_discovery-0}
\end{table*}
\begin{table*}[!tbp]
\begin{minipage}{\linewidth}
\normalsize
\subsubsection{Expected Payoff Probe: Replication}\label{app:stat-risk-summary-replication}
Holm correction covers 12 comparisons.
\end{minipage}
\par\vspace{3pt}

\centering\fontsize{9}{10}\selectfont\setlength{\tabcolsep}{2pt}\renewcommand{\arraystretch}{0.96}
\begin{tabular*}{\textwidth}{@{\extracolsep{\fill}}l>{\raggedright\arraybackslash}p{.16\textwidth}>{\raggedright\arraybackslash}p{.10\textwidth}rrlrr@{}}
\toprule
Model & Task / outcome & Contrast & N/G & $\Delta$ & 95\% CI & $p$ & $p_{\mathrm{Holm}}$ \\
\midrule
Llama 3B & Risk summary / switch rate & T-BASE & 80/20 & +0.00 & [-10.00, +10.00] & 1.000 & 1.000 \\
Llama 3B & Risk summary / switch rate & F-BASE & 80/20 & -13.75 & [-20.00, -6.25] & 0.0074 & 0.0517 \\
Llama 3B & Risk summary / switch rate & F-T & 80/20 & -13.75 & [-21.25, -5.00] & 0.0114 & 0.0681 \\
Llama 8B & Risk summary / switch rate & T-BASE & 80/20 & -17.50 & [-22.50, -12.50] & 1.22e-04 & \textbf{0.0012} \\
Llama 8B & Risk summary / switch rate & F-BASE & 80/20 & -50.00 & [-50.00, -50.00] & 1.91e-06 & \textbf{2.29e-05} \\
Llama 8B & Risk summary / switch rate & F-T & 80/20 & -32.50 & [-37.50, -27.50] & 1.91e-06 & \textbf{2.29e-05} \\
Qwen 3B & Risk summary / switch rate & T-BASE & 80/20 & +2.50 & [-7.50, +13.75] & 0.8203 & 1.000 \\
Qwen 3B & Risk summary / switch rate & F-BASE & 80/20 & -22.50 & [-32.50, -12.50] & 0.0011 & \textbf{0.0099} \\
Qwen 3B & Risk summary / switch rate & F-T & 80/20 & -25.00 & [-37.50, -12.50] & 0.0034 & \textbf{0.0273} \\
Qwen 7B & Risk summary / switch rate & T-BASE & 80/20 & +0.00 & [-8.75, +8.75] & 1.000 & 1.000 \\
Qwen 7B & Risk summary / switch rate & F-BASE & 80/20 & -2.50 & [-10.00, +5.00] & 0.7539 & 1.000 \\
Qwen 7B & Risk summary / switch rate & F-T & 80/20 & -2.50 & [-10.00, +5.00] & 0.7656 & 1.000 \\
\bottomrule
\end{tabular*}
\caption{Expected Payoff Probe: Replication: paired tests, part 1 (Holm family: 12).}
\label{tab:stat-risk_summary_replication-0}
\end{table*}
\begin{table*}[!tbp]
\begin{minipage}{\linewidth}
\normalsize
\subsubsection{Controlled Bargaining: Exit Rates}\label{app:stat-controlled-bargaining-rate}
Holm correction covers 12 comparisons.

Significant differences in bargaining exit rates are observed for Llama 8B and Qwen 3B, whereas Llama 3B and Qwen 7B show no exit-rate differences in these controlled probes.
\end{minipage}
\par\vspace{3pt}

\centering\fontsize{9}{10}\selectfont\setlength{\tabcolsep}{2pt}\renewcommand{\arraystretch}{0.96}
\begin{tabular*}{\textwidth}{@{\extracolsep{\fill}}l>{\raggedright\arraybackslash}p{.16\textwidth}>{\raggedright\arraybackslash}p{.10\textwidth}rrlrr@{}}
\toprule
Model & Task / outcome & Contrast & N/G & $\Delta$ & 95\% CI & $p$ & $p_{\mathrm{Holm}}$ \\
\midrule
Llama 3B & bargaining / EXIT & T-BASE & 315/15 & +0.00 & [+0.00, +0.00] & 1.000 & 1.000 \\
Llama 3B & bargaining / EXIT & F-BASE & 315/15 & +0.00 & [+0.00, +0.00] & 1.000 & 1.000 \\
Llama 3B & bargaining / EXIT & F-T & 315/15 & +0.00 & [+0.00, +0.00] & 1.000 & 1.000 \\
Llama 8B & bargaining / EXIT & T-BASE & 315/15 & -24.44 & [-31.11, -17.78] & 1.22e-04 & \textbf{0.0011} \\
Llama 8B & bargaining / EXIT & F-BASE & 315/15 & +1.90 & [-1.27, +4.76] & 0.3350 & 1.000 \\
Llama 8B & bargaining / EXIT & F-T & 315/15 & +26.35 & [+19.68, +32.70] & 6.10e-05 & \textbf{7.32e-04} \\
Qwen 3B & bargaining / EXIT & T-BASE & 315/15 & -4.13 & [-5.40, -2.86] & 4.88e-04 & \textbf{0.0039} \\
Qwen 3B & bargaining / EXIT & F-BASE & 315/15 & +14.60 & [+11.75, +17.46] & 6.10e-05 & \textbf{7.32e-04} \\
Qwen 3B & bargaining / EXIT & F-T & 315/15 & +18.73 & [+15.87, +21.59] & 6.10e-05 & \textbf{7.32e-04} \\
Qwen 7B & bargaining / EXIT & T-BASE & 315/15 & +0.00 & [+0.00, +0.00] & 1.000 & 1.000 \\
Qwen 7B & bargaining / EXIT & F-BASE & 315/15 & +0.00 & [+0.00, +0.00] & 1.000 & 1.000 \\
Qwen 7B & bargaining / EXIT & F-T & 315/15 & +0.00 & [+0.00, +0.00] & 1.000 & 1.000 \\
\bottomrule
\end{tabular*}
\caption{Controlled Bargaining: Exit Rates: paired tests, part 1 (Holm family: 12).}
\label{tab:stat-controlled_bargaining_rate-0}
\end{table*}
\begin{table*}[!tbp]
\begin{minipage}{\linewidth}
\normalsize
\subsubsection{Controlled Bargaining: Offer Response}\label{app:stat-bargaining-offer-response}
Holm correction covers 24 comparisons.
\end{minipage}
\par\vspace{3pt}

\centering\fontsize{9}{10}\selectfont\setlength{\tabcolsep}{2pt}\renewcommand{\arraystretch}{0.96}
\begin{tabular*}{\textwidth}{@{\extracolsep{\fill}}l>{\raggedright\arraybackslash}p{.16\textwidth}>{\raggedright\arraybackslash}p{.10\textwidth}rrlrr@{}}
\toprule
Model & Task / outcome & Contrast & N/G & $\Delta$ & 95\% CI & $p$ & $p_{\mathrm{Holm}}$ \\
\midrule
Llama 3B & bargaining / EXIT response & BASE:high-low & 90/15 & +0.00 & [+0.00, +0.00] & 1.000 & 1.000 \\
Llama 3B & bargaining / EXIT response & T:high-low & 90/15 & +0.00 & [+0.00, +0.00] & 1.000 & 1.000 \\
Llama 3B & bargaining / EXIT response & F:high-low & 90/15 & +0.00 & [+0.00, +0.00] & 1.000 & 1.000 \\
Llama 3B & bargaining / EXIT response difference & T-BASE & 90/15 & +0.00 & [+0.00, +0.00] & 1.000 & 1.000 \\
Llama 3B & bargaining / EXIT response difference & F-BASE & 90/15 & +0.00 & [+0.00, +0.00] & 1.000 & 1.000 \\
Llama 3B & bargaining / EXIT response difference & F-T & 90/15 & +0.00 & [+0.00, +0.00] & 1.000 & 1.000 \\
Llama 8B & bargaining / EXIT response & BASE:high-low & 90/15 & -80.00 & [-93.33, -64.44] & 1.22e-04 & \textbf{0.0027} \\
Llama 8B & bargaining / EXIT response & T:high-low & 90/15 & -22.22 & [-28.89, -13.33] & 0.0020 & \textbf{0.0352} \\
Llama 8B & bargaining / EXIT response & F:high-low & 90/15 & -75.56 & [-88.89, -60.00] & 6.10e-05 & \textbf{0.0015} \\
Llama 8B & bargaining / EXIT response difference & T-BASE & 90/15 & +57.78 & [+46.67, +68.89] & 1.22e-04 & \textbf{0.0027} \\
Llama 8B & bargaining / EXIT response difference & F-BASE & 90/15 & +4.44 & [-4.44, +15.56] & 0.7500 & 1.000 \\
Llama 8B & bargaining / EXIT response difference & F-T & 90/15 & -53.33 & [-62.22, -44.44] & 6.10e-05 & \textbf{0.0015} \\
Qwen 3B & bargaining / EXIT response & BASE:high-low & 90/15 & -22.22 & [-28.89, -13.33] & 0.0020 & \textbf{0.0352} \\
Qwen 3B & bargaining / EXIT response & T:high-low & 90/15 & -2.22 & [-6.67, +0.00] & 1.000 & 1.000 \\
Qwen 3B & bargaining / EXIT response & F:high-low & 90/15 & -33.33 & [-40.00, -26.67] & 1.22e-04 & \textbf{0.0027} \\
Qwen 3B & bargaining / EXIT response difference & T-BASE & 90/15 & +20.00 & [+11.11, +28.89] & 0.0039 & 0.0625 \\
Qwen 3B & bargaining / EXIT response difference & F-BASE & 90/15 & -11.11 & [-20.00, +0.00] & 0.1250 & 1.000 \\
Qwen 3B & bargaining / EXIT response difference & F-T & 90/15 & -31.11 & [-33.33, -26.67] & 1.22e-04 & \textbf{0.0027} \\
\bottomrule
\end{tabular*}
\caption{Controlled Bargaining: Offer Response: paired tests, part 1 (Holm family: 24).}
\label{tab:stat-bargaining_offer_response-0}
\end{table*}
\begin{table*}[!tbp]
\centering\fontsize{9}{10}\selectfont\setlength{\tabcolsep}{2pt}\renewcommand{\arraystretch}{0.96}
\begin{tabular*}{\textwidth}{@{\extracolsep{\fill}}l>{\raggedright\arraybackslash}p{.16\textwidth}>{\raggedright\arraybackslash}p{.10\textwidth}rrlrr@{}}
\toprule
Model & Task / outcome & Contrast & N/G & $\Delta$ & 95\% CI & $p$ & $p_{\mathrm{Holm}}$ \\
\midrule
Qwen 7B & bargaining / EXIT response & BASE:high-low & 90/15 & +0.00 & [+0.00, +0.00] & 1.000 & 1.000 \\
Qwen 7B & bargaining / EXIT response & T:high-low & 90/15 & +0.00 & [+0.00, +0.00] & 1.000 & 1.000 \\
Qwen 7B & bargaining / EXIT response & F:high-low & 90/15 & +0.00 & [+0.00, +0.00] & 1.000 & 1.000 \\
Qwen 7B & bargaining / EXIT response difference & T-BASE & 90/15 & +0.00 & [+0.00, +0.00] & 1.000 & 1.000 \\
Qwen 7B & bargaining / EXIT response difference & F-BASE & 90/15 & +0.00 & [+0.00, +0.00] & 1.000 & 1.000 \\
Qwen 7B & bargaining / EXIT response difference & F-T & 90/15 & +0.00 & [+0.00, +0.00] & 1.000 & 1.000 \\
\bottomrule
\end{tabular*}
\caption{Controlled Bargaining: Offer Response: paired tests, part 2 (Holm family: 24).}
\label{tab:stat-bargaining_offer_response-18}
\end{table*}
\begin{table*}[!tbp]
\begin{minipage}{\linewidth}
\normalsize
\subsection{Controlled Feedback Tests}
None of the 72 feedback comparisons passes Holm correction. These probes contain only two authored scenario versions and a selected second turn cohort, so their observed differences are reported descriptively rather than treated as established effects.

\subsubsection{Controlled Feedback Probes}\label{app:stat-controlled-feedback}
Holm correction covers 72 comparisons.
\end{minipage}
\par\vspace{3pt}

\centering\fontsize{9}{10}\selectfont\setlength{\tabcolsep}{2pt}\renewcommand{\arraystretch}{0.96}
\begin{tabular*}{\textwidth}{@{\extracolsep{\fill}}l>{\raggedright\arraybackslash}p{.30\textwidth}>{\raggedright\arraybackslash}p{.08\textwidth}rrlrr@{}}
\toprule
Model & Task / outcome & Contrast & N/G & $\Delta$ & 95\% CI & $p$ & $p_{\mathrm{Holm}}$ \\
\midrule
Llama 3B & Movie / fits:ACCEPT & T-BASE & 12/2 & +0.00 & NA & 1.000 & 1.000 \\
Llama 3B & Movie / fits:ACCEPT & F-BASE & 12/2 & +0.00 & NA & 1.000 & 1.000 \\
Llama 3B & Movie / fits:ACCEPT & F-T & 12/2 & +0.00 & NA & 1.000 & 1.000 \\
Llama 3B & Movie / mismatch:OTHER & T-BASE & 12/2 & +0.00 & NA & 1.000 & 1.000 \\
Llama 3B & Movie / mismatch:OTHER & F-BASE & 12/2 & +0.00 & NA & 1.000 & 1.000 \\
Llama 3B & Movie / mismatch:OTHER & F-T & 12/2 & +0.00 & NA & 1.000 & 1.000 \\
Llama 3B & Movie / unresolved:ASK & T-BASE & 12/2 & -8.33 & NA & 1.000 & 1.000 \\
Llama 3B & Movie / unresolved:ASK & F-BASE & 12/2 & +0.00 & NA & 1.000 & 1.000 \\
Llama 3B & Movie / unresolved:ASK & F-T & 12/2 & +8.33 & NA & 1.000 & 1.000 \\
Llama 3B & Travel / fits:ACCEPT & T-BASE & 12/2 & +8.33 & NA & 1.000 & 1.000 \\
Llama 3B & Travel / fits:ACCEPT & F-BASE & 12/2 & +0.00 & NA & 1.000 & 1.000 \\
Llama 3B & Travel / fits:ACCEPT & F-T & 12/2 & -8.33 & NA & 1.000 & 1.000 \\
Llama 3B & Travel / mismatch:OTHER & T-BASE & 12/2 & +0.00 & NA & 1.000 & 1.000 \\
Llama 3B & Travel / mismatch:OTHER & F-BASE & 12/2 & +0.00 & NA & 1.000 & 1.000 \\
Llama 3B & Travel / mismatch:OTHER & F-T & 12/2 & +0.00 & NA & 1.000 & 1.000 \\
Llama 3B & Travel / unresolved:ASK & T-BASE & 12/2 & +0.00 & NA & 1.000 & 1.000 \\
Llama 3B & Travel / unresolved:ASK & F-BASE & 12/2 & +0.00 & NA & 1.000 & 1.000 \\
Llama 3B & Travel / unresolved:ASK & F-T & 12/2 & +0.00 & NA & 1.000 & 1.000 \\
\bottomrule
\end{tabular*}
\caption{Controlled Feedback Probes: paired tests, part 1 (Holm family: 72).}
\label{tab:stat-controlled_feedback-0}
\end{table*}
\begin{table*}[!tbp]
\centering\fontsize{9}{10}\selectfont\setlength{\tabcolsep}{2pt}\renewcommand{\arraystretch}{0.96}
\begin{tabular*}{\textwidth}{@{\extracolsep{\fill}}l>{\raggedright\arraybackslash}p{.30\textwidth}>{\raggedright\arraybackslash}p{.08\textwidth}rrlrr@{}}
\toprule
Model & Task / outcome & Contrast & N/G & $\Delta$ & 95\% CI & $p$ & $p_{\mathrm{Holm}}$ \\
\midrule
Llama 8B & Movie / fits:ACCEPT & T-BASE & 10/2 & +60.00 & NA & 0.5000 & 1.000 \\
Llama 8B & Movie / fits:ACCEPT & F-BASE & 10/2 & -20.00 & NA & 1.000 & 1.000 \\
Llama 8B & Movie / fits:ACCEPT & F-T & 10/2 & -80.00 & NA & 0.5000 & 1.000 \\
Llama 8B & Movie / mismatch:OTHER & T-BASE & 10/2 & -50.00 & NA & 0.5000 & 1.000 \\
Llama 8B & Movie / mismatch:OTHER & F-BASE & 10/2 & +20.00 & NA & 0.5000 & 1.000 \\
Llama 8B & Movie / mismatch:OTHER & F-T & 10/2 & +70.00 & NA & 0.5000 & 1.000 \\
Llama 8B & Movie / unresolved:ASK & T-BASE & 10/2 & +0.00 & NA & 1.000 & 1.000 \\
Llama 8B & Movie / unresolved:ASK & F-BASE & 10/2 & -20.00 & NA & 1.000 & 1.000 \\
Llama 8B & Movie / unresolved:ASK & F-T & 10/2 & -20.00 & NA & 1.000 & 1.000 \\
Llama 8B & Travel / fits:ACCEPT & T-BASE & 12/2 & +25.00 & NA & 0.5000 & 1.000 \\
Llama 8B & Travel / fits:ACCEPT & F-BASE & 12/2 & -25.00 & NA & 0.5000 & 1.000 \\
Llama 8B & Travel / fits:ACCEPT & F-T & 12/2 & -50.00 & NA & 0.5000 & 1.000 \\
Llama 8B & Travel / mismatch:OTHER & T-BASE & 12/2 & +0.00 & NA & 1.000 & 1.000 \\
Llama 8B & Travel / mismatch:OTHER & F-BASE & 12/2 & +41.67 & NA & 0.5000 & 1.000 \\
Llama 8B & Travel / mismatch:OTHER & F-T & 12/2 & +41.67 & NA & 0.5000 & 1.000 \\
Llama 8B & Travel / unresolved:ASK & T-BASE & 12/2 & +0.00 & NA & 1.000 & 1.000 \\
Llama 8B & Travel / unresolved:ASK & F-BASE & 12/2 & +0.00 & NA & 1.000 & 1.000 \\
Llama 8B & Travel / unresolved:ASK & F-T & 12/2 & +0.00 & NA & 1.000 & 1.000 \\
\bottomrule
\end{tabular*}
\caption{Controlled Feedback Probes: paired tests, part 2 (Holm family: 72).}
\label{tab:stat-controlled_feedback-18}
\end{table*}
\begin{table*}[!tbp]
\centering\fontsize{9}{10}\selectfont\setlength{\tabcolsep}{2pt}\renewcommand{\arraystretch}{0.96}
\begin{tabular*}{\textwidth}{@{\extracolsep{\fill}}l>{\raggedright\arraybackslash}p{.30\textwidth}>{\raggedright\arraybackslash}p{.08\textwidth}rrlrr@{}}
\toprule
Model & Task / outcome & Contrast & N/G & $\Delta$ & 95\% CI & $p$ & $p_{\mathrm{Holm}}$ \\
\midrule
Qwen 3B & Movie / fits:ACCEPT & T-BASE & 12/2 & -16.67 & NA & 0.5000 & 1.000 \\
Qwen 3B & Movie / fits:ACCEPT & F-BASE & 12/2 & -16.67 & NA & 1.000 & 1.000 \\
Qwen 3B & Movie / fits:ACCEPT & F-T & 12/2 & +0.00 & NA & 1.000 & 1.000 \\
Qwen 3B & Movie / mismatch:OTHER & T-BASE & 12/2 & +0.00 & NA & 1.000 & 1.000 \\
Qwen 3B & Movie / mismatch:OTHER & F-BASE & 12/2 & +0.00 & NA & 1.000 & 1.000 \\
Qwen 3B & Movie / mismatch:OTHER & F-T & 12/2 & +0.00 & NA & 1.000 & 1.000 \\
Qwen 3B & Movie / unresolved:ASK & T-BASE & 12/2 & +0.00 & NA & 1.000 & 1.000 \\
Qwen 3B & Movie / unresolved:ASK & F-BASE & 12/2 & +0.00 & NA & 1.000 & 1.000 \\
Qwen 3B & Movie / unresolved:ASK & F-T & 12/2 & +0.00 & NA & 1.000 & 1.000 \\
Qwen 3B & Travel / fits:ACCEPT & T-BASE & 8/2 & +0.00 & NA & 1.000 & 1.000 \\
Qwen 3B & Travel / fits:ACCEPT & F-BASE & 8/2 & +0.00 & NA & 1.000 & 1.000 \\
Qwen 3B & Travel / fits:ACCEPT & F-T & 8/2 & +0.00 & NA & 1.000 & 1.000 \\
Qwen 3B & Travel / mismatch:OTHER & T-BASE & 8/2 & -12.50 & NA & 1.000 & 1.000 \\
Qwen 3B & Travel / mismatch:OTHER & F-BASE & 8/2 & +0.00 & NA & 1.000 & 1.000 \\
Qwen 3B & Travel / mismatch:OTHER & F-T & 8/2 & +12.50 & NA & 1.000 & 1.000 \\
Qwen 3B & Travel / unresolved:ASK & T-BASE & 8/2 & +0.00 & NA & 1.000 & 1.000 \\
Qwen 3B & Travel / unresolved:ASK & F-BASE & 8/2 & +0.00 & NA & 1.000 & 1.000 \\
Qwen 3B & Travel / unresolved:ASK & F-T & 8/2 & +0.00 & NA & 1.000 & 1.000 \\
\bottomrule
\end{tabular*}
\caption{Controlled Feedback Probes: paired tests, part 3 (Holm family: 72).}
\label{tab:stat-controlled_feedback-36}
\end{table*}
\begin{table*}[!tbp]
\centering\fontsize{9}{10}\selectfont\setlength{\tabcolsep}{2pt}\renewcommand{\arraystretch}{0.96}
\begin{tabular*}{\textwidth}{@{\extracolsep{\fill}}l>{\raggedright\arraybackslash}p{.30\textwidth}>{\raggedright\arraybackslash}p{.08\textwidth}rrlrr@{}}
\toprule
Model & Task / outcome & Contrast & N/G & $\Delta$ & 95\% CI & $p$ & $p_{\mathrm{Holm}}$ \\
\midrule
Qwen 7B & Movie / fits:ACCEPT & T-BASE & 12/2 & -16.67 & NA & 1.000 & 1.000 \\
Qwen 7B & Movie / fits:ACCEPT & F-BASE & 12/2 & +0.00 & NA & 1.000 & 1.000 \\
Qwen 7B & Movie / fits:ACCEPT & F-T & 12/2 & +16.67 & NA & 1.000 & 1.000 \\
Qwen 7B & Movie / mismatch:OTHER & T-BASE & 12/2 & +0.00 & NA & 1.000 & 1.000 \\
Qwen 7B & Movie / mismatch:OTHER & F-BASE & 12/2 & +0.00 & NA & 1.000 & 1.000 \\
Qwen 7B & Movie / mismatch:OTHER & F-T & 12/2 & +0.00 & NA & 1.000 & 1.000 \\
Qwen 7B & Movie / unresolved:ASK & T-BASE & 12/2 & +0.00 & NA & 1.000 & 1.000 \\
Qwen 7B & Movie / unresolved:ASK & F-BASE & 12/2 & +0.00 & NA & 1.000 & 1.000 \\
Qwen 7B & Movie / unresolved:ASK & F-T & 12/2 & +0.00 & NA & 1.000 & 1.000 \\
Qwen 7B & Travel / fits:ACCEPT & T-BASE & 12/2 & +0.00 & NA & 1.000 & 1.000 \\
Qwen 7B & Travel / fits:ACCEPT & F-BASE & 12/2 & -8.33 & NA & 1.000 & 1.000 \\
Qwen 7B & Travel / fits:ACCEPT & F-T & 12/2 & -8.33 & NA & 1.000 & 1.000 \\
Qwen 7B & Travel / mismatch:OTHER & T-BASE & 12/2 & +0.00 & NA & 1.000 & 1.000 \\
Qwen 7B & Travel / mismatch:OTHER & F-BASE & 12/2 & -8.33 & NA & 1.000 & 1.000 \\
Qwen 7B & Travel / mismatch:OTHER & F-T & 12/2 & -8.33 & NA & 1.000 & 1.000 \\
Qwen 7B & Travel / unresolved:ASK & T-BASE & 12/2 & +0.00 & NA & 1.000 & 1.000 \\
Qwen 7B & Travel / unresolved:ASK & F-BASE & 12/2 & +0.00 & NA & 1.000 & 1.000 \\
Qwen 7B & Travel / unresolved:ASK & F-T & 12/2 & +0.00 & NA & 1.000 & 1.000 \\
\bottomrule
\end{tabular*}
\caption{Controlled Feedback Probes: paired tests, part 4 (Holm family: 72).}
\label{tab:stat-controlled_feedback-54}
\end{table*}
\begin{table*}[!tbp]
\begin{minipage}{\linewidth}
\normalsize
\subsection{Capability Tests}
\subsubsection{Prompt Capabilities}\label{app:stat-capability-prompt}
Holm correction covers 18 comparisons.

Personality guidance produces task-dependent capability changes. Prompting generally reduces ARC-Easy accuracy, while combined guidance often improves GSM8K performance but can negatively affect BoolQ accuracy.
\end{minipage}
\par\vspace{3pt}

\centering\fontsize{9}{10}\selectfont\setlength{\tabcolsep}{2pt}\renewcommand{\arraystretch}{0.96}
\begin{tabular*}{\textwidth}{@{\extracolsep{\fill}}l>{\raggedright\arraybackslash}p{.14\textwidth}>{\raggedright\arraybackslash}p{.09\textwidth}rrlrrr@{}}
\toprule
Model & Task / outcome & Contrast & N/G & $\Delta$ & 95\% CI & $p$ & $p_{\mathrm{Holm}}$ & $p_{\mathrm{degr.}}$ \\
\midrule
Llama 3B & ARC-Easy & T:Base P & 2000/2000 & -3.70 & [-4.90, -2.50] & 2.56e-09 & \textbf{3.84e-08} & \textbf{1.92e-08} \\
Llama 3B & BoolQ & T:Base P & 2000/2000 & -0.80 & [-1.95, +0.35] & 0.1915 & 1.000 & 0.7660 \\
Llama 3B & ARC-Easy & F:Base P & 2000/2000 & -5.65 & [-6.95, -4.35] & 6.02e-18 & \textbf{1.08e-16} & \textbf{5.42e-17} \\
Llama 3B & BoolQ & F:Base P & 2000/2000 & +0.20 & [-0.95, +1.35] & 0.7956 & 1.000 & 1.000 \\
Qwen 3B & GSM8K & T:Base P & 1300/1300 & -0.31 & [-1.85, +1.23] & 0.7750 & 1.000 & 1.000 \\
Qwen 3B & ARC-Easy & T:Base P & 2000/2000 & -1.95 & [-2.80, -1.05] & 2.18e-05 & \textbf{3.05e-04} & \textbf{1.53e-04} \\
Qwen 3B & BoolQ & T:Base P & 2000/2000 & -0.35 & [-1.15, +0.45] & 0.4704 & 1.000 & 1.000 \\
Qwen 3B & GSM8K & F:Base P & 1300/1300 & -0.23 & [-1.77, +1.31] & 0.8496 & 1.000 & 1.000 \\
Qwen 3B & ARC-Easy & F:Base P & 2000/2000 & -2.00 & [-2.95, -1.10] & 2.97e-05 & \textbf{3.86e-04} & \textbf{1.93e-04} \\
Qwen 3B & BoolQ & F:Base P & 2000/2000 & -0.25 & [-1.10, +0.60] & 0.6488 & 1.000 & 1.000 \\
Qwen 7B & GSM8K & T:Base P & 1300/1300 & -0.23 & [-1.69, +1.23] & 0.8392 & 1.000 & 1.000 \\
Qwen 7B & ARC-Easy & T:Base P & 2000/2000 & -0.85 & [-1.55, -0.20] & 0.0186 & 0.2049 & 0.1024 \\
Qwen 7B & BoolQ & T:Base P & 2000/2000 & -4.05 & [-5.25, -2.85] & 4.69e-11 & \textbf{7.97e-10} & \textbf{3.99e-10} \\
Qwen 7B & GSM8K & F:Base P & 1300/1300 & -0.23 & [-1.77, +1.31] & 0.8454 & 1.000 & 1.000 \\
Qwen 7B & ARC-Easy & F:Base P & 2000/2000 & -1.35 & [-2.10, -0.65] & 3.55e-04 & \textbf{0.0043} & \textbf{0.0021} \\
Qwen 7B & BoolQ & F:Base P & 2000/2000 & -3.85 & [-5.00, -2.70] & 9.89e-11 & \textbf{1.58e-09} & \textbf{7.91e-10} \\
Llama 3B & GSM8K & Prompt-T-BASE & 1300/1300 & -1.54 & [-3.62, +0.54] & 0.1657 & 1.000 & 0.7455 \\
Llama 3B & GSM8K & Prompt-F-BASE & 1300/1300 & -1.85 & [-3.92, +0.23] & 0.0932 & 0.9319 & 0.4660 \\
\bottomrule
\end{tabular*}
\caption{Prompt Capabilities: paired tests, part 1 (Holm family: 18).}
\label{tab:stat-capability_prompt-0}
\end{table*}
\begin{table*}[!tbp]
\begin{minipage}{\linewidth}
\normalsize
\subsubsection{Prompt BoolQ Passage Sensitivity}\label{app:stat-capability-prompt-passage-sensitivity}
Holm correction covers 6 comparisons.
\end{minipage}
\par\vspace{3pt}

\centering\fontsize{9}{10}\selectfont\setlength{\tabcolsep}{2pt}\renewcommand{\arraystretch}{0.96}
\begin{tabular*}{\textwidth}{@{\extracolsep{\fill}}l>{\raggedright\arraybackslash}p{.16\textwidth}>{\raggedright\arraybackslash}p{.10\textwidth}rrlrr@{}}
\toprule
Model & Task / outcome & Contrast & N/G & $\Delta$ & 95\% CI & $p$ & $p_{\mathrm{Holm}}$ \\
\midrule
Llama 3B & BoolQ & T:Base P & 2000/1897 & -0.80 & [-1.95, +0.35] & 0.1975 & 0.7901 \\
Llama 3B & BoolQ & F:Base P & 2000/1897 & +0.20 & [-0.95, +1.36] & 0.7993 & 1.000 \\
Qwen 3B & BoolQ & T:Base P & 2000/1897 & -0.35 & [-1.16, +0.45] & 0.4759 & 1.000 \\
Qwen 3B & BoolQ & F:Base P & 2000/1897 & -0.25 & [-1.11, +0.60] & 0.6483 & 1.000 \\
Qwen 7B & BoolQ & T:Base P & 2000/1897 & -4.05 & [-5.27, -2.83] & 1.00e-05 & \textbf{6.00e-05} \\
Qwen 7B & BoolQ & F:Base P & 2000/1897 & -3.85 & [-5.06, -2.67] & 1.00e-05 & \textbf{6.00e-05} \\
\bottomrule
\end{tabular*}
\caption{Prompt BoolQ Passage Sensitivity: paired tests, part 1 (Holm family: 6).}
\label{tab:stat-capability_prompt_passage_sensitivity-0}
\end{table*}
\begin{table*}[!tbp]
\begin{minipage}{\linewidth}
\normalsize
\subsubsection{Combined Capabilities}\label{app:stat-capability-combined}
Holm correction covers 36 comparisons.
\end{minipage}
\par\vspace{3pt}

\centering\fontsize{9}{10}\selectfont\setlength{\tabcolsep}{2pt}\renewcommand{\arraystretch}{0.96}
\begin{tabular*}{\textwidth}{@{\extracolsep{\fill}}l>{\raggedright\arraybackslash}p{.14\textwidth}>{\raggedright\arraybackslash}p{.09\textwidth}rrlrrr@{}}
\toprule
Model & Task / outcome & Contrast & N/G & $\Delta$ & 95\% CI & $p$ & $p_{\mathrm{Holm}}$ & $p_{\mathrm{degr.}}$ \\
\midrule
Llama 3B & GSM8K & T+T:T & 1300/1300 & +19.92 & [+16.92, +22.92] & 3.01e-37 & \textbf{1.02e-35} & 1.000 \\
Llama 3B & ARC-Easy & T+T:T & 2000/2000 & -3.85 & [-5.05, -2.65] & 5.95e-10 & \textbf{1.55e-08} & \textbf{8.92e-09} \\
Llama 3B & BoolQ & T+T:T & 2000/2000 & -1.40 & [-2.55, -0.25] & 0.0212 & 0.2332 & 0.2438 \\
Llama 3B & GSM8K & T+F:T & 1300/1300 & +19.31 & [+16.31, +22.38] & 1.34e-33 & \textbf{4.41e-32} & 1.000 \\
Llama 3B & ARC-Easy & T+F:T & 2000/2000 & -4.45 & [-5.75, -3.20] & 1.37e-11 & \textbf{3.83e-10} & \textbf{2.19e-10} \\
Llama 3B & BoolQ & T+F:T & 2000/2000 & +0.20 & [-0.95, +1.35] & 0.7985 & 1.000 & 1.000 \\
Llama 3B & GSM8K & F+T:F & 1300/1300 & +42.54 & [+39.46, +45.62] & 2.40e-124 & \textbf{8.63e-123} & 1.000 \\
Llama 3B & ARC-Easy & F+T:F & 2000/2000 & -1.60 & [-2.75, -0.40] & 0.0090 & 0.1356 & 0.1221 \\
Llama 3B & BoolQ & F+T:F & 2000/2000 & -0.85 & [-2.05, +0.30] & 0.1868 & 0.9647 & 1.000 \\
Llama 3B & GSM8K & F+F:F & 1300/1300 & +39.69 & [+36.62, +42.77] & 3.57e-110 & \textbf{1.25e-108} & 1.000 \\
Llama 3B & ARC-Easy & F+F:F & 2000/2000 & -5.85 & [-7.20, -4.50] & 3.22e-17 & \textbf{1.03e-15} & \textbf{5.79e-16} \\
Llama 3B & BoolQ & F+F:F & 2000/2000 & -0.15 & [-1.30, +1.00] & 0.8654 & 1.000 & 1.000 \\
Qwen 3B & GSM8K & T+T:T & 1300/1300 & +3.92 & [+2.15, +5.77] & 4.04e-05 & \textbf{8.07e-04} & 1.000 \\
Qwen 3B & ARC-Easy & T+T:T & 2000/2000 & -1.20 & [-2.15, -0.30] & 0.0149 & 0.1820 & 0.1784 \\
Qwen 3B & BoolQ & T+T:T & 2000/2000 & +0.20 & [-0.70, +1.10] & 0.7520 & 1.000 & 1.000 \\
Qwen 3B & GSM8K & T+F:T & 1300/1300 & +4.69 & [+2.85, +6.54] & 7.49e-07 & \textbf{1.57e-05} & 1.000 \\
Qwen 3B & ARC-Easy & T+F:T & 2000/2000 & -1.35 & [-2.30, -0.35] & 0.0093 & 0.1356 & 0.1221 \\
Qwen 3B & BoolQ & T+F:T & 2000/2000 & -0.05 & [-1.00, +0.90] & 1.000 & 1.000 & 1.000 \\
\bottomrule
\end{tabular*}
\caption{Combined Capabilities: paired tests, part 1 (Holm family: 36).}
\label{tab:stat-capability_combined-0}
\end{table*}
\begin{table*}[!tbp]
\centering\fontsize{9}{10}\selectfont\setlength{\tabcolsep}{2pt}\renewcommand{\arraystretch}{0.96}
\begin{tabular*}{\textwidth}{@{\extracolsep{\fill}}l>{\raggedright\arraybackslash}p{.14\textwidth}>{\raggedright\arraybackslash}p{.09\textwidth}rrlrrr@{}}
\toprule
Model & Task / outcome & Contrast & N/G & $\Delta$ & 95\% CI & $p$ & $p_{\mathrm{Holm}}$ & $p_{\mathrm{degr.}}$ \\
\midrule
Qwen 3B & GSM8K & F+T:F & 1300/1300 & +2.62 & [+0.92, +4.38] & 0.0042 & 0.0670 & 1.000 \\
Qwen 3B & ARC-Easy & F+T:F & 2000/2000 & -0.90 & [-1.90, +0.05] & 0.0854 & 0.5979 & 0.8114 \\
Qwen 3B & BoolQ & F+T:F & 2000/2000 & -1.15 & [-2.05, -0.30] & 0.0140 & 0.1820 & 0.1750 \\
Qwen 3B & GSM8K & F+F:F & 1300/1300 & +3.00 & [+1.23, +4.77] & 0.0011 & \textbf{0.0185} & 1.000 \\
Qwen 3B & ARC-Easy & F+F:F & 2000/2000 & -1.00 & [-2.00, +0.00] & 0.0619 & 0.5226 & 0.6193 \\
Qwen 3B & BoolQ & F+F:F & 2000/2000 & -1.15 & [-2.10, -0.20] & 0.0235 & 0.2349 & 0.2584 \\
Qwen 7B & GSM8K & T+T:T & 1300/1300 & +5.85 & [+4.00, +7.77] & 8.90e-10 & \textbf{2.23e-08} & 1.000 \\
Qwen 7B & ARC-Easy & T+T:T & 2000/2000 & -0.75 & [-1.50, +0.00] & 0.0581 & 0.5226 & 0.6097 \\
Qwen 7B & BoolQ & T+T:T & 2000/2000 & -4.75 & [-5.95, -3.60] & 1.27e-15 & \textbf{3.93e-14} & \textbf{2.22e-14} \\
Qwen 7B & GSM8K & T+F:T & 1300/1300 & +5.23 & [+3.38, +7.15] & 9.23e-08 & \textbf{2.03e-06} & 1.000 \\
Qwen 7B & ARC-Easy & T+F:T & 2000/2000 & -1.45 & [-2.25, -0.65] & 4.22e-04 & \textbf{0.0080} & \textbf{0.0061} \\
Qwen 7B & BoolQ & T+F:T & 2000/2000 & -4.50 & [-5.65, -3.35] & 5.41e-15 & \textbf{1.62e-13} & \textbf{9.19e-14} \\
Qwen 7B & GSM8K & F+T:F & 1300/1300 & +5.46 & [+3.62, +7.31] & 1.32e-08 & \textbf{3.17e-07} & 1.000 \\
Qwen 7B & ARC-Easy & F+T:F & 2000/2000 & -0.55 & [-1.25, +0.15] & 0.1608 & 0.9647 & 1.000 \\
Qwen 7B & BoolQ & F+T:F & 2000/2000 & -4.55 & [-5.85, -3.25] & 3.25e-12 & \textbf{9.43e-11} & \textbf{5.37e-11} \\
Qwen 7B & GSM8K & F+F:F & 1300/1300 & +5.31 & [+3.46, +7.23] & 4.30e-08 & \textbf{9.90e-07} & 1.000 \\
Qwen 7B & ARC-Easy & F+F:F & 2000/2000 & -1.35 & [-2.15, -0.60] & 7.30e-04 & \textbf{0.0131} & \textbf{0.0102} \\
Qwen 7B & BoolQ & F+F:F & 2000/2000 & -4.15 & [-5.35, -2.95] & 1.44e-11 & \textbf{3.88e-10} & \textbf{2.23e-10} \\
\bottomrule
\end{tabular*}
\caption{Combined Capabilities: paired tests, part 2 (Holm family: 36).}
\label{tab:stat-capability_combined-18}
\end{table*}
\begin{table*}[!tbp]
\begin{minipage}{\linewidth}
\normalsize
\subsubsection{Combined BoolQ Passage Sensitivity}\label{app:stat-capability-combined-passage-sensitivity}
Holm correction covers 12 comparisons.
\end{minipage}
\par\vspace{3pt}

\centering\fontsize{9}{10}\selectfont\setlength{\tabcolsep}{2pt}\renewcommand{\arraystretch}{0.96}
\begin{tabular*}{\textwidth}{@{\extracolsep{\fill}}l>{\raggedright\arraybackslash}p{.16\textwidth}>{\raggedright\arraybackslash}p{.10\textwidth}rrlrr@{}}
\toprule
Model & Task / outcome & Contrast & N/G & $\Delta$ & 95\% CI & $p$ & $p_{\mathrm{Holm}}$ \\
\midrule
Llama 3B & BoolQ & T+T:T & 2000/1897 & -1.40 & [-2.57, -0.20] & 0.0253 & 0.1703 \\
Llama 3B & BoolQ & T+F:T & 2000/1897 & +0.20 & [-0.99, +1.36] & 0.8033 & 1.000 \\
Llama 3B & BoolQ & F+T:F & 2000/1897 & -0.85 & [-2.06, +0.35] & 0.1893 & 0.9466 \\
Llama 3B & BoolQ & F+F:F & 2000/1897 & -0.15 & [-1.31, +1.00] & 0.8655 & 1.000 \\
Qwen 3B & BoolQ & T+T:T & 2000/1897 & +0.20 & [-0.75, +1.15] & 0.7564 & 1.000 \\
Qwen 3B & BoolQ & T+F:T & 2000/1897 & -0.05 & [-1.01, +0.91] & 1.000 & 1.000 \\
Qwen 3B & BoolQ & F+T:F & 2000/1897 & -1.15 & [-2.04, -0.25] & 0.0154 & 0.1229 \\
Qwen 3B & BoolQ & F+F:F & 2000/1897 & -1.15 & [-2.11, -0.20] & 0.0243 & 0.1703 \\
Qwen 7B & BoolQ & T+T:T & 2000/1897 & -4.75 & [-5.98, -3.57] & 1.00e-05 & \textbf{1.20e-04} \\
Qwen 7B & BoolQ & T+F:T & 2000/1897 & -4.50 & [-5.69, -3.35] & 1.00e-05 & \textbf{1.20e-04} \\
Qwen 7B & BoolQ & F+T:F & 2000/1897 & -4.55 & [-5.87, -3.29] & 1.00e-05 & \textbf{1.20e-04} \\
Qwen 7B & BoolQ & F+F:F & 2000/1897 & -4.15 & [-5.37, -2.95] & 1.00e-05 & \textbf{1.20e-04} \\
\bottomrule
\end{tabular*}
\caption{Combined BoolQ Passage Sensitivity: paired tests, part 1 (Holm family: 12).}
\label{tab:stat-capability_combined_passage_sensitivity-0}
\end{table*}
\begin{table*}[!tbp]
\begin{minipage}{\linewidth}
\normalsize
\subsubsection{Trained Variant Capabilities}\label{app:stat-capability-current-dpo}
Only the Llama 3B T BoolQ decline passes the primary two sided correction among 24 comparisons. The separately corrected degradation tests additionally identify Qwen 3B F BoolQ and Llama 8B F ARC-Easy declines. Small score changes should therefore not be equated with proven capability equivalence.

Holm correction covers 24 comparisons.
\end{minipage}
\par\vspace{3pt}

\centering\fontsize{9}{10}\selectfont\setlength{\tabcolsep}{2pt}\renewcommand{\arraystretch}{0.96}
\begin{tabular*}{\textwidth}{@{\extracolsep{\fill}}l>{\raggedright\arraybackslash}p{.14\textwidth}>{\raggedright\arraybackslash}p{.09\textwidth}rrlrrr@{}}
\toprule
Model & Task / outcome & Contrast & N/G & $\Delta$ & 95\% CI & $p$ & $p_{\mathrm{Holm}}$ & $p_{\mathrm{degr.}}$ \\
\midrule
Llama 3B & ARC-Easy & F:Base & 2000/2000 & -0.15 & [-0.85, +0.55] & 0.7877 & 1.000 & 1.000 \\
Llama 3B & BoolQ & F:Base & 2000/2000 & +1.65 & [+0.55, +2.75] & 0.0050 & 0.1046 & 1.000 \\
Llama 3B & ARC-Easy & T:Base & 2000/2000 & -0.55 & [-1.15, +0.05] & 0.0989 & 1.000 & 1.000 \\
Llama 3B & BoolQ & T:Base & 2000/2000 & -2.25 & [-3.05, -1.50] & 1.18e-08 & \textbf{2.82e-07} & \textbf{1.41e-07} \\
Qwen 3B & GSM8K & F:Base & 1300/1300 & +0.77 & [-0.31, +1.85] & 0.2026 & 1.000 & 1.000 \\
Qwen 3B & ARC-Easy & F:Base & 2000/2000 & -0.10 & [-0.60, +0.40] & 0.8388 & 1.000 & 1.000 \\
Qwen 3B & BoolQ & F:Base & 2000/2000 & -0.85 & [-1.40, -0.35] & 0.0023 & 0.0533 & \textbf{0.0266} \\
Qwen 3B & GSM8K & T:Base & 1300/1300 & -0.54 & [-1.38, +0.31] & 0.2810 & 1.000 & 1.000 \\
Qwen 3B & ARC-Easy & T:Base & 2000/2000 & -0.20 & [-0.70, +0.30] & 0.5572 & 1.000 & 1.000 \\
Qwen 3B & BoolQ & T:Base & 2000/2000 & +0.30 & [-0.15, +0.75] & 0.2632 & 1.000 & 1.000 \\
Qwen 7B & GSM8K & T:Base & 1300/1300 & +0.46 & [-0.38, +1.31] & 0.3616 & 1.000 & 1.000 \\
Qwen 7B & ARC-Easy & T:Base & 2000/2000 & -0.05 & [-0.30, +0.20] & 1.000 & 1.000 & 1.000 \\
Qwen 7B & BoolQ & T:Base & 2000/2000 & +0.20 & [-0.20, +0.60] & 0.4545 & 1.000 & 1.000 \\
Qwen 7B & ARC-Easy & F:Base & 2000/2000 & -0.15 & [-0.40, +0.10] & 0.4531 & 1.000 & 1.000 \\
Qwen 7B & BoolQ & F:Base & 2000/2000 & -0.20 & [-0.65, +0.25] & 0.5034 & 1.000 & 1.000 \\
Qwen 7B & GSM8K & F:Base & 1300/1300 & -0.62 & [-1.46, +0.23] & 0.2005 & 1.000 & 1.000 \\
Llama 3B & GSM8K & T:Base & 1300/1300 & -0.69 & [-2.38, +1.00] & 0.4744 & 1.000 & 1.000 \\
Llama 3B & GSM8K & F:Base & 1300/1300 & -0.85 & [-2.77, +1.08] & 0.4307 & 1.000 & 1.000 \\
\bottomrule
\end{tabular*}
\caption{Trained Variant Capabilities: paired tests, part 1 (Holm family: 24).}
\label{tab:stat-capability_current_dpo-0}
\end{table*}
\begin{table*}[!tbp]
\centering\fontsize{9}{10}\selectfont\setlength{\tabcolsep}{2pt}\renewcommand{\arraystretch}{0.96}
\begin{tabular*}{\textwidth}{@{\extracolsep{\fill}}l>{\raggedright\arraybackslash}p{.14\textwidth}>{\raggedright\arraybackslash}p{.09\textwidth}rrlrrr@{}}
\toprule
Model & Task / outcome & Contrast & N/G & $\Delta$ & 95\% CI & $p$ & $p_{\mathrm{Holm}}$ & $p_{\mathrm{degr.}}$ \\
\midrule
Llama 8B & GSM8K & T:Base & 1300/1300 & -0.08 & [-1.62, +1.46] & 1.000 & 1.000 & 1.000 \\
Llama 8B & GSM8K & F:Base & 1300/1300 & -1.15 & [-2.69, +0.38] & 0.1548 & 1.000 & 1.000 \\
Llama 8B & ARC-Easy & T:Base & 2000/2000 & -0.05 & [-0.45, +0.35] & 1.000 & 1.000 & 1.000 \\
Llama 8B & ARC-Easy & F:Base & 2000/2000 & -0.85 & [-1.40, -0.30] & 0.0033 & 0.0732 & \textbf{0.0366} \\
Llama 8B & BoolQ & T:Base & 2000/2000 & -0.05 & [-0.55, +0.45] & 1.000 & 1.000 & 1.000 \\
Llama 8B & BoolQ & F:Base & 2000/2000 & -0.40 & [-0.95, +0.15] & 0.2153 & 1.000 & 1.000 \\
\bottomrule
\end{tabular*}
\caption{Trained Variant Capabilities: paired tests, part 2 (Holm family: 24).}
\label{tab:stat-capability_current_dpo-18}
\end{table*}
\begin{table*}[!tbp]
\begin{minipage}{\linewidth}
\normalsize
\subsubsection{Trained Variant BoolQ Passage Sensitivity}\label{app:stat-capability-current-dpo-current-passage-sensitivity}
Holm correction covers 8 comparisons.

The passage-level sensitivity analysis identifies significant BoolQ performance differences for both Llama 3B variants and the Qwen 3B F variant after Holm correction.
\end{minipage}
\par\vspace{3pt}

\centering\fontsize{9}{10}\selectfont\setlength{\tabcolsep}{2pt}\renewcommand{\arraystretch}{0.96}
\begin{tabular*}{\textwidth}{@{\extracolsep{\fill}}l>{\raggedright\arraybackslash}p{.16\textwidth}>{\raggedright\arraybackslash}p{.10\textwidth}rrlrr@{}}
\toprule
Model & Task / outcome & Contrast & N/G & $\Delta$ & 95\% CI & $p$ & $p_{\mathrm{Holm}}$ \\
\midrule
Llama 3B & BoolQ & F:Base & 2000/1897 & +1.65 & [+0.55, +2.79] & 0.0051 & \textbf{0.0305} \\
Llama 3B & BoolQ & T:Base & 2000/1897 & -2.25 & [-3.05, -1.45] & 1.00e-05 & \textbf{8.00e-05} \\
Qwen 3B & BoolQ & F:Base & 2000/1897 & -0.85 & [-1.40, -0.30] & 0.0030 & \textbf{0.0211} \\
Qwen 3B & BoolQ & T:Base & 2000/1897 & +0.30 & [-0.15, +0.75] & 0.2632 & 1.000 \\
Qwen 7B & BoolQ & T:Base & 2000/1897 & +0.20 & [-0.20, +0.60] & 0.4545 & 1.000 \\
Qwen 7B & BoolQ & F:Base & 2000/1897 & -0.20 & [-0.65, +0.25] & 0.5034 & 1.000 \\
Llama 8B & BoolQ & T:Base & 2000/1897 & -0.05 & [-0.55, +0.45] & 1.000 & 1.000 \\
Llama 8B & BoolQ & F:Base & 2000/1897 & -0.40 & [-0.96, +0.15] & 0.2135 & 1.000 \\
\bottomrule
\end{tabular*}
\caption{Trained Variant BoolQ Passage Sensitivity: paired tests, part 1 (Holm family: 8).}
\label{tab:stat-capability_current_dpo_current_passage_sensitivity-0}
\end{table*}

\clearpage

%% file: sections/appendix_other_dimensions.tex
\clearpage
\begin{table*}[p]
\begin{minipage}{\textwidth}
\normalsize
\section{Complementary MBTI Dimensions}\label{app:otherdimensions}
We extend the situational evaluation to E/I, S/N, and J/P on Llama 3.2 3B, Qwen2.5 3B, and Qwen2.5 7B, using the corresponding early-stopped DPO variants and fixed axis-specific prompts. We select two tasks per behavioral level and retain all eight guidance conditions per model and dimension. Inputs, output requirements, and greedy decoding are held fixed across conditions; Llama 3.1 8B is outside this supplement.

A and B denote the first and second poles of each axis (E/I, S/N, J/P). A guided and B guided report DPO / Prompt, with Prompt applied to Base; A+B denotes DPO A with Prompt B. All values are percentages of all task inputs, including unparseable responses in the denominator. These are descriptive judgment and choice rates, rather than accuracy scores.

\end{minipage}
\par\vspace{7pt}
\centering\fontsize{9}{10}\selectfont\setlength{\tabcolsep}{2pt}
\renewcommand{\arraystretch}{1.04}
\begin{tabular*}{\textwidth}{@{\extracolsep{\fill}}lllccrrrr@{}}
\toprule
Axis & Model & Measure & A guided & B guided & A+A & A+B & B+A & B+B \\
\midrule
E/I & Llama 3B & Circa unclear & 5.6 / 0.6 & 78.5 / 3.4 & 0.6 & 1.9 & 10.7 & 44.5 \\
E/I & Llama 3B & Circa unparseable & 0.5 / 0.0 & 0.0 / 0.0 & 0.0 & 0.0 & 0.0 & 0.0 \\
E/I & Llama 3B & CB false & 2.1 / 75.5 & 60.6 / 66.9 & 55.2 & 37.2 & 97.7 & 91.0 \\
E/I & Llama 3B & CB unparseable & 94.3 / 0.1 & 38.9 / 3.6 & 4.3 & 22.9 & 0.0 & 3.9 \\
E/I & Qwen 3B & Circa unclear & 75.9 / 78.9 & 94.3 / 83.4 & 70.0 & 75.2 & 87.0 & 90.5 \\
E/I & Qwen 3B & Circa unparseable & 4.9 / 0.0 & 0.9 / 0.0 & 0.2 & 0.2 & 0.0 & 0.0 \\
E/I & Qwen 3B & CB false & 29.0 / 29.1 & 18.1 / 25.4 & 33.8 & 30.2 & 19.5 & 16.2 \\
E/I & Qwen 3B & CB unparseable & 0.0 / 0.0 & 0.0 / 0.0 & 0.0 & 0.0 & 0.0 & 0.0 \\
E/I & Qwen 7B & Circa unclear & 59.5 / 33.5 & 65.6 / 45.9 & 27.9 & 40.1 & 41.3 & 52.2 \\
E/I & Qwen 7B & Circa unparseable & 1.6 / 0.2 & 0.3 / 0.1 & 0.3 & 0.2 & 0.2 & 0.1 \\
E/I & Qwen 7B & CB false & 26.4 / 25.5 & 24.3 / 23.4 & 24.9 & 23.0 & 27.4 & 24.4 \\
E/I & Qwen 7B & CB unparseable & 0.0 / 0.0 & 0.1 / 0.0 & 0.0 & 0.0 & 0.0 & 0.0 \\
\midrule
S/N & Llama 3B & Circa unclear & 2.3 / 1.8 & 53.5 / 2.4 & 0.2 & 0.5 & 20.5 & 23.6 \\
S/N & Llama 3B & Circa unparseable & 0.0 / 0.0 & 0.0 / 0.0 & 0.0 & 0.0 & 0.0 & 0.1 \\
S/N & Llama 3B & CB false & 11.4 / 71.1 & 16.0 / 67.7 & 89.3 & 90.3 & 12.6 & 6.8 \\
S/N & Llama 3B & CB unparseable & 88.6 / 1.6 & 9.2 / 5.6 & 6.6 & 7.7 & 0.1 & 0.3 \\
S/N & Qwen 3B & Circa unclear & 80.4 / 75.5 & 89.4 / 82.8 & 58.8 & 68.0 & 87.0 & 91.5 \\
S/N & Qwen 3B & Circa unparseable & 1.0 / 0.0 & 3.1 / 0.1 & 0.1 & 0.1 & 0.1 & 0.2 \\
S/N & Qwen 3B & CB false & 38.9 / 28.5 & 17.9 / 27.8 & 40.1 & 38.9 & 17.3 & 17.2 \\
S/N & Qwen 3B & CB unparseable & 0.0 / 0.0 & 0.0 / 0.0 & 0.0 & 0.0 & 0.0 & 0.0 \\
S/N & Qwen 7B & Circa unclear & 42.8 / 34.0 & 66.3 / 48.1 & 25.5 & 38.9 & 42.6 & 55.3 \\
S/N & Qwen 7B & Circa unparseable & 18.9 / 0.1 & 0.4 / 0.1 & 0.1 & 0.1 & 0.1 & 0.1 \\
S/N & Qwen 7B & CB false & 27.3 / 26.0 & 23.5 / 19.3 & 28.6 & 21.6 & 22.4 & 15.9 \\
S/N & Qwen 7B & CB unparseable & 0.8 / 0.0 & 0.0 / 0.0 & 0.0 & 0.0 & 0.0 & 0.0 \\
\midrule
J/P & Llama 3B & Circa unclear & 21.9 / 4.8 & 41.2 / 5.1 & 8.2 & 7.7 & 14.5 & 19.3 \\
J/P & Llama 3B & Circa unparseable & 0.8 / 0.0 & 0.0 / 0.0 & 0.0 & 0.0 & 0.0 & 0.0 \\
J/P & Llama 3B & CB false & 24.1 / 73.2 & 39.1 / 79.8 & 76.0 & 80.5 & 94.9 & 96.8 \\
J/P & Llama 3B & CB unparseable & 75.3 / 0.0 & 60.0 / 0.0 & 0.3 & 0.0 & 0.0 & 0.2 \\
J/P & Qwen 3B & Circa unclear & 47.1 / 81.5 & 94.8 / 85.9 & 68.8 & 74.0 & 91.6 & 94.1 \\
J/P & Qwen 3B & Circa unparseable & 33.9 / 0.0 & 0.0 / 0.2 & 0.1 & 0.4 & 0.0 & 0.1 \\
J/P & Qwen 3B & CB false & 33.1 / 36.5 & 19.3 / 20.0 & 42.8 & 27.0 & 25.3 & 13.8 \\
J/P & Qwen 3B & CB unparseable & 0.1 / 0.0 & 0.0 / 0.0 & 0.0 & 0.0 & 0.0 & 0.0 \\
J/P & Qwen 7B & Circa unclear & 54.2 / 36.0 & 72.7 / 56.6 & 27.8 & 45.4 & 47.5 & 71.3 \\
J/P & Qwen 7B & Circa unparseable & 1.0 / 0.1 & 0.1 / 0.2 & 0.1 & 0.2 & 0.2 & 0.3 \\
J/P & Qwen 7B & CB false & 29.2 / 23.3 & 21.6 / 19.3 & 26.0 & 22.7 & 19.5 & 13.6 \\
J/P & Qwen 7B & CB unparseable & 0.0 / 0.0 & 0.0 / 0.0 & 0.0 & 0.0 & 0.0 & 0.0 \\
\bottomrule
\end{tabular*}
\caption{Selected situational interpretation results (\%). Circa reports UNCLEAR judgments and unparseable outputs over 32,602 inputs; CommitmentBank (CB) reports FALSE judgments and unparseable outputs over 1,009 inputs. Guided cells show DPO / Prompt; combined columns show the trained pole plus the prompted pole.}
\label{tab:other-interpretation}
\end{table*}

\begin{table*}[p]
\begin{minipage}{\textwidth}
\normalsize
\subsection{Individual and Interactive Decisions}
Stock and bargaining report SELL and EXIT choices in the controlled scenarios (1,005 and 900 inputs). INSPIRED reports ASK\_MORE after recommendation feedback (119 inputs), and Frames reports SEARCH\_OTHER after travel feedback (282 inputs). All responses in these four selected tasks are parseable across the reported conditions.

\paragraph{Brief analysis.}
Without additional prompts, I, N, and P yield more UNCLEAR judgments on Circa and more ASK\_MORE choices on INSPIRED than their opposite poles in all three models. The size of these differences and their response to aligned or opposing prompts vary across tasks and models. Controlled bargaining shows no DPO EXIT choices in Llama 3B or Qwen 7B. High unparseable rates in some interpretation conditions, especially Llama 3B on CommitmentBank, limit behavioral interpretation of those differences.

\end{minipage}
\par\vspace{7pt}
\centering\fontsize{9}{10}\selectfont\setlength{\tabcolsep}{2pt}
\renewcommand{\arraystretch}{1.04}
\begin{tabular*}{\textwidth}{@{\extracolsep{\fill}}lllccrrrr@{}}
\toprule
Axis & Model & Measure & A guided & B guided & A+A & A+B & B+A & B+B \\
\midrule
E/I & Llama 3B & Stock sell & 48.5 / 29.4 & 9.2 / 0.0 & 29.9 & 0.7 & 12.9 & 0.0 \\
E/I & Llama 3B & Bargaining exit & 0.0 / 0.0 & 0.0 / 84.0 & 0.0 & 11.7 & 1.2 & 100.0 \\
E/I & Llama 3B & INSPIRED ask & 17.6 / 0.0 & 89.9 / 10.9 & 0.0 & 3.4 & 32.8 & 86.6 \\
E/I & Llama 3B & Frames search & 28.7 / 51.1 & 49.3 / 69.1 & 18.4 & 37.9 & 84.4 & 70.6 \\
E/I & Qwen 3B & Stock sell & 99.7 / 81.9 & 92.9 / 0.1 & 98.4 & 78.3 & 0.2 & 0.0 \\
E/I & Qwen 3B & Bargaining exit & 9.1 / 0.0 & 10.7 / 1.8 & 0.0 & 0.3 & 0.4 & 20.6 \\
E/I & Qwen 3B & INSPIRED ask & 72.3 / 88.2 & 92.4 / 94.1 & 73.1 & 79.0 & 91.6 & 95.0 \\
E/I & Qwen 3B & Frames search & 86.2 / 80.1 & 94.7 / 69.5 & 79.4 & 69.5 & 84.0 & 71.6 \\
E/I & Qwen 7B & Stock sell & 40.4 / 15.4 & 39.0 / 14.6 & 12.5 & 17.9 & 16.3 & 11.7 \\
E/I & Qwen 7B & Bargaining exit & 0.0 / 0.0 & 0.0 / 7.4 & 0.0 & 7.1 & 0.0 & 5.9 \\
E/I & Qwen 7B & INSPIRED ask & 63.0 / 40.3 & 66.4 / 34.5 & 37.8 & 32.8 & 37.8 & 31.9 \\
E/I & Qwen 7B & Frames search & 34.4 / 39.4 & 41.5 / 52.1 & 35.1 & 46.8 & 44.7 & 59.2 \\
\midrule
S/N & Llama 3B & Stock sell & 65.6 / 8.3 & 27.3 / 0.4 & 40.8 & 10.3 & 0.0 & 0.0 \\
S/N & Llama 3B & Bargaining exit & 0.0 / 48.2 & 0.0 / 0.2 & 52.3 & 0.4 & 0.0 & 0.0 \\
S/N & Llama 3B & INSPIRED ask & 1.7 / 1.7 & 95.0 / 18.5 & 0.0 & 1.7 & 81.5 & 97.5 \\
S/N & Llama 3B & Frames search & 18.1 / 43.3 & 66.0 / 83.0 & 8.2 & 24.8 & 98.6 & 99.6 \\
S/N & Qwen 3B & Stock sell & 97.5 / 86.8 & 99.6 / 100.0 & 63.6 & 98.4 & 88.2 & 97.6 \\
S/N & Qwen 3B & Bargaining exit & 5.0 / 2.7 & 19.2 / 0.0 & 1.6 & 0.0 & 9.4 & 0.8 \\
S/N & Qwen 3B & INSPIRED ask & 79.8 / 90.8 & 87.4 / 89.1 & 88.2 & 87.4 & 95.0 & 91.6 \\
S/N & Qwen 3B & Frames search & 86.5 / 80.1 & 92.2 / 70.6 & 75.5 & 59.2 & 85.1 & 79.1 \\
S/N & Qwen 7B & Stock sell & 37.5 / 33.5 & 41.4 / 6.8 & 29.3 & 8.0 & 34.2 & 6.7 \\
S/N & Qwen 7B & Bargaining exit & 0.0 / 45.4 & 0.0 / 0.0 & 50.8 & 0.0 & 32.8 & 0.0 \\
S/N & Qwen 7B & INSPIRED ask & 59.7 / 5.0 & 72.3 / 60.5 & 0.8 & 46.2 & 21.8 & 73.9 \\
S/N & Qwen 7B & Frames search & 44.0 / 49.6 & 34.4 / 55.0 & 53.5 & 58.9 & 46.5 & 52.5 \\
\midrule
J/P & Llama 3B & Stock sell & 74.6 / 53.7 & 31.2 / 0.0 & 75.4 & 16.5 & 4.5 & 0.0 \\
J/P & Llama 3B & Bargaining exit & 0.0 / 30.1 & 0.0 / 0.0 & 22.9 & 0.0 & 47.4 & 0.0 \\
J/P & Llama 3B & INSPIRED ask & 5.0 / 1.7 & 100.0 / 16.0 & 0.8 & 4.2 & 84.0 & 97.5 \\
J/P & Llama 3B & Frames search & 22.3 / 37.9 & 68.4 / 80.5 & 8.5 & 41.1 & 96.5 & 98.6 \\
J/P & Qwen 3B & Stock sell & 97.7 / 92.1 & 99.6 / 58.2 & 91.7 & 45.5 & 93.1 & 62.3 \\
J/P & Qwen 3B & Bargaining exit & 6.6 / 0.0 & 6.7 / 0.0 & 1.2 & 0.0 & 0.0 & 0.0 \\
J/P & Qwen 3B & INSPIRED ask & 62.2 / 79.0 & 94.1 / 96.6 & 46.2 & 77.3 & 95.8 & 99.2 \\
J/P & Qwen 3B & Frames search & 75.9 / 64.2 & 96.5 / 83.7 & 48.9 & 72.0 & 76.2 & 91.1 \\
J/P & Qwen 7B & Stock sell & 43.2 / 42.5 & 36.8 / 0.7 & 43.5 & 2.0 & 36.5 & 0.3 \\
J/P & Qwen 7B & Bargaining exit & 0.0 / 24.4 & 0.0 / 0.0 & 20.4 & 0.0 & 24.4 & 0.0 \\
J/P & Qwen 7B & INSPIRED ask & 61.3 / 16.0 & 68.9 / 57.1 & 12.6 & 51.3 & 16.8 & 70.6 \\
J/P & Qwen 7B & Frames search & 36.5 / 35.8 & 39.7 / 56.0 & 34.8 & 52.1 & 38.7 & 59.9 \\
\bottomrule
\end{tabular*}
\caption{Selected individual and interactive decision results (\%), with the same guidance notation as Table~\ref{tab:other-interpretation}. Bargaining exit rates aggregate the fixed buyer-offer scenarios. INSPIRED ask and Frames search denote ASK\_MORE and SEARCH\_OTHER, respectively.}
\label{tab:other-decisions}
\end{table*}
\clearpage